\documentclass{article} 
\usepackage{iclr2019_conference,times}

\usepackage{amsmath,amsfonts,bm}

\def\eqref#1{equation~\ref{#1}}

\def\1{\bm{1}}

\DeclareMathAlphabet{\mathsfit}{\encodingdefault}{\sfdefault}{m}{sl}
\SetMathAlphabet{\mathsfit}{bold}{\encodingdefault}{\sfdefault}{bx}{n}

\usepackage{hyperref}
\usepackage{url}
\usepackage[utf8]{inputenc} 
\usepackage[T1]{fontenc}    
\usepackage{hyperref}       
\usepackage{url}            
\usepackage{booktabs}       
\usepackage{amsfonts}       
\usepackage{nicefrac}       
\usepackage{microtype}      
\usepackage{graphicx}
\usepackage{amsmath}
\usepackage[noend]{algpseudocode}
\usepackage{algorithmicx,algorithm}
\usepackage{caption,subcaption}
\usepackage{multirow}
\usepackage{wrapfig}
\usepackage{color}

\newcommand{\red}[1]{\textcolor{red}{#1}}
\newcommand{\blue}[1]{\textcolor{blue}{#1}}

\title{Dynamically Unfolding Recurrent Restorer:\\
A Moving Endpoint Control Method for Image Restoration}

\author{
	Xiaoshuai Zhang\thanks{Equal contribution.}\\
	Institute of Computer Science and Technology,\\
	Peking University\\
	\texttt{jet@pku.edu.cn} \\
	\And
	Yiping Lu$^*$ \\
	School Of Mathmatical Science,\\
	Peking university \\
	\texttt{luyiping9712@pku.edu.cn} \\
	\And
	Jiaying Liu \\
	Institute of Computer Science and Technology,\\
	Peking University\\
	\texttt{liujiaying@pku.edu.cn}\\
	\And
	Bin Dong \\
	Beijing International Center for Mathematical Research, Peking University\\
	Center for Data Science, Peking University\\
	Beijing Institute of Big Data Research,\\
	Beijing, China\\
	\texttt{dongbin@math.pku.edu.cn}\\
}

\iclrfinalcopy 
\begin{document}

\maketitle

\begin{abstract}
In this paper, we propose a new control framework called
the moving endpoint control to restore images corrupted by
different degradation levels using a single model. The proposed control problem
contains an image restoration dynamic which is modeled by a convolutional RNN.
The moving endpoint, which is essentially the terminal time of
the associated dynamic, is determined by a policy network. We call
the proposed model the dynamically unfolding recurrent restorer (DURR).
Numerical experiments show that DURR is able to achieve
state-of-the-art performances on blind image denoising and
JPEG image deblocking. Furthermore, DURR can well generalize to images with
higher degradation levels that are not included in the training stage.
\end{abstract}
\section{Introduction}

Image restoration, including image denoising, deblurring, inpainting,
\textit{etc.}, is one of the most important areas in imaging science.
Its major purpose is to obtain high quality reconstructions of images
corrupted in various ways during imaging, acquisiting, and storing,
and enable us to see crucial but subtle objects that reside in the images. Image restoration has been an active research area. Numerous models
and algorithms have been developed for the past few decades. Before
the uprise of deep learning methods, there were two classes of
image restoration approaches that were widely adopted in the field: transformation based approach and PDE approach. The transformation based
approach includes wavelet and wavelet frame based methods
\citep{elad2005simultaneous, starck2005image,
	daubechies2007iteratively,cai2009split},
dictionary learning based methods \citep{aharon2006rm}, similarity based methods
\citep{buades2005non,dabov2007image}, low-rank models
\citep{ji2010robust,gu2014weighted}, \textit{etc.}
The PDE approach includes variational models \citep{mumford1989optimal,
	rudin1992nonlinear,bredies2010total}), nonlinear diffusions
\citep{perona1990scale,catte1992image,weickert1998anisotropic},
nonlinear hyperbolic equations \citep{Osher1990}, \textit{etc.}
More recently, deep connections between wavelet frame based methods
and PDE approach were established \citep{cai2012image,
	cai2016image,dong2017image}.

One of the greatest challenge for image restoration is to properly
handle image degradations of different levels. In the existing
transformation based or PDE based methods, there is always at least one
tuning parameter (\textit{e.g.} the regularization parameter for variational models
and terminal time for nonlinear diffusions) that needs to be
manually selected. The choice of the parameter heavily relies on
the degradation level.

Recent years, deep learning models for image restoration tasks
have significantly advanced the state-of-the-art of the field.
\cite{jain2009natural} proposed a convolutional neural network (CNN)
for image denoising which has better expressive power than the MRF models
by \cite{lan2006efficient}. Inspired by nonlinear diffusions,
\cite{Chen2017Trainable} designed a deep neural network
for image denoising and \cite{zhang2017beyond} improves the capacity
by introducing a deeper neural network with residual connections.
\cite{tai2017memnet} introduced a deep network with long term memory
which was inspired by neural science. However, these models
cannot gracefully handle images with varied degradation levels.
Although one may train different models for images with different levels,
this may limit the application of these models in practice due to
lack of flexibility.

Taking blind image denoising for example. \cite{zhang2017beyond}
designed a 20-layer neural network for the task, called DnCNN-B,
which had a huge number of parameters.
To reduce number of parameters, \cite{lefkimmiatis2017universal}
proposed the UNLNet$_5$, by unrolling a projection gradient algorithm
for a constrained optimization model. However,
\cite{lefkimmiatis2017universal} also observed a drop in PSNR comparing
to DnCNN. Therefore, the design of a light-weighted and yet effective model
for blind image denoising remains a challenge. Moreover,
deep learning based models trained on simulated gaussian noise
images usually fail to handle real world noise,
as will be illustrated in later sections.

Another example is JPEG image deblocking. JPEG is
the most commonly used lossy image compression method. However,
this method tend to introduce undesired artifacts as
the compression rate increases. JPEG image deblocking aims to eliminate
the artifacts and improve the image quality. Recently,
deep learning based methods were proposed for JPEG deblocking
\citep{dong2015compression,zhang2017beyond,zhang2018dmcnn}.
However, most of their models are trained and evaluated on a given
quality factor. Thus it would be hard for these methods to apply to
Internet images, where the quality factors are usually unknown.

In this paper, we propose a single image restoration model that can robustly restore images with varied degradation levels even when the degradation level is well outside of that of the training set. Our proposed model for image restoration is inspired by the recent development on the relation between deep learning and optimal control. The relation between
supervised deep learning methods and optimal control
has been discovered and exploited by \cite{Weinan2017A, Lu2018Beyond,
	Chang2017Reversible,fang2017feature}. The key idea is to consider
the residual block $x_{n+1}=x_n+f(x_n)$ as an approximation to
the continuous dynamics $\dot X = f(X)$. In particular,
\cite{Lu2018Beyond,fang2017feature} demonstrated that the training process
of a class of deep models (\textit{e.g.} ResNet by \cite{he2016deep},
PolyNet by \cite{zhang2017polynet}, \textit{etc.})
can be understood as solving the following control problem:

\begin{align}
\min_w&\ \left(L(X(T),y)+ \int_{0}^{\tau} R(w(t),t)dt\right) \nonumber\\
s.t.&\  \dot{X}=f(X(t),w(t)),t\in(0,\tau) \\
& X(0)=x_0. \nonumber
\end{align}

Here $x_0$ is the input, $y$ is the regression target or label,
$\dot{X}=f(X,w)$ is the deep neural network with parameter $w(t)$, $R$ is
the regularization term and $L$ can be any loss function to measure
the difference between the reconstructed images and the ground truths.

\begin{figure}[ht]
	\centering
	\includegraphics[width=5.3in]{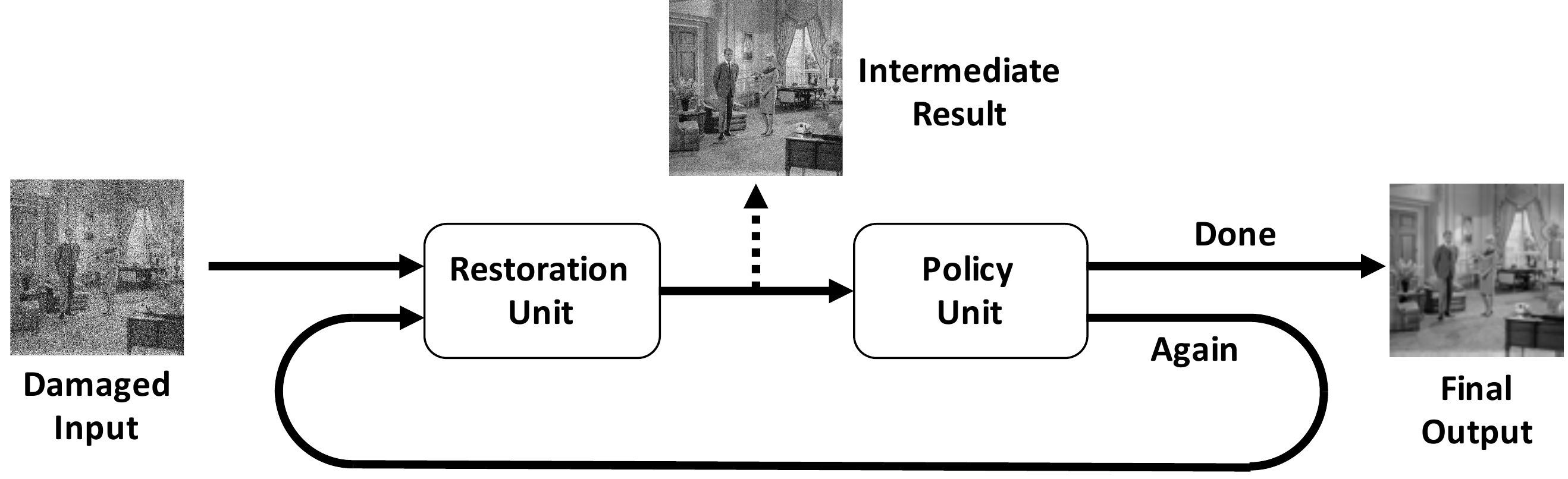}
	\caption{Pipeline of the dynamically unfolding recurrent restorer (DURR).}
	\label{rlrnn}
\end{figure}

In the context of image restoration, the control dynamic $\dot X = f(X(t),\omega(t)), t\in(0,\tau)$
can be, for example, a diffusion process learned using a deep neural network.
The terminal time $\tau$ of the diffusion corresponds to the depth of the neural network.
Previous works simply fixed the depth of the network, \textit{i.e.} the terminal time,
as a fixed hyper-parameter. However \cite{mrazek2003selection} showed that
the optimal terminal time of diffusion differs from image to image.
Furthermore, when an image is corrupted by higher noise levels,
the optimal terminal time for a typical noise removal diffusion should be
greater than when a less noisy image is being processed. This is
the main reason why current deep models are not robust enough to handle images
with varied noise levels. In this paper, we no longer treat the terminal time
as a hyper-parameter. Instead, we design a new architecture (see Fig. \ref{rlrnn})
that contains both a deep diffusion-like network and another network that determines
the optimal terminal time for each input image. We propose a novel moving endpoint control model
to train the aforementioned architecture. We call
the proposed architecture the dynamically unfolding recurrent restorer (DURR).

\begin{figure}[ht]
	\centering
	\includegraphics[width=4.0in]{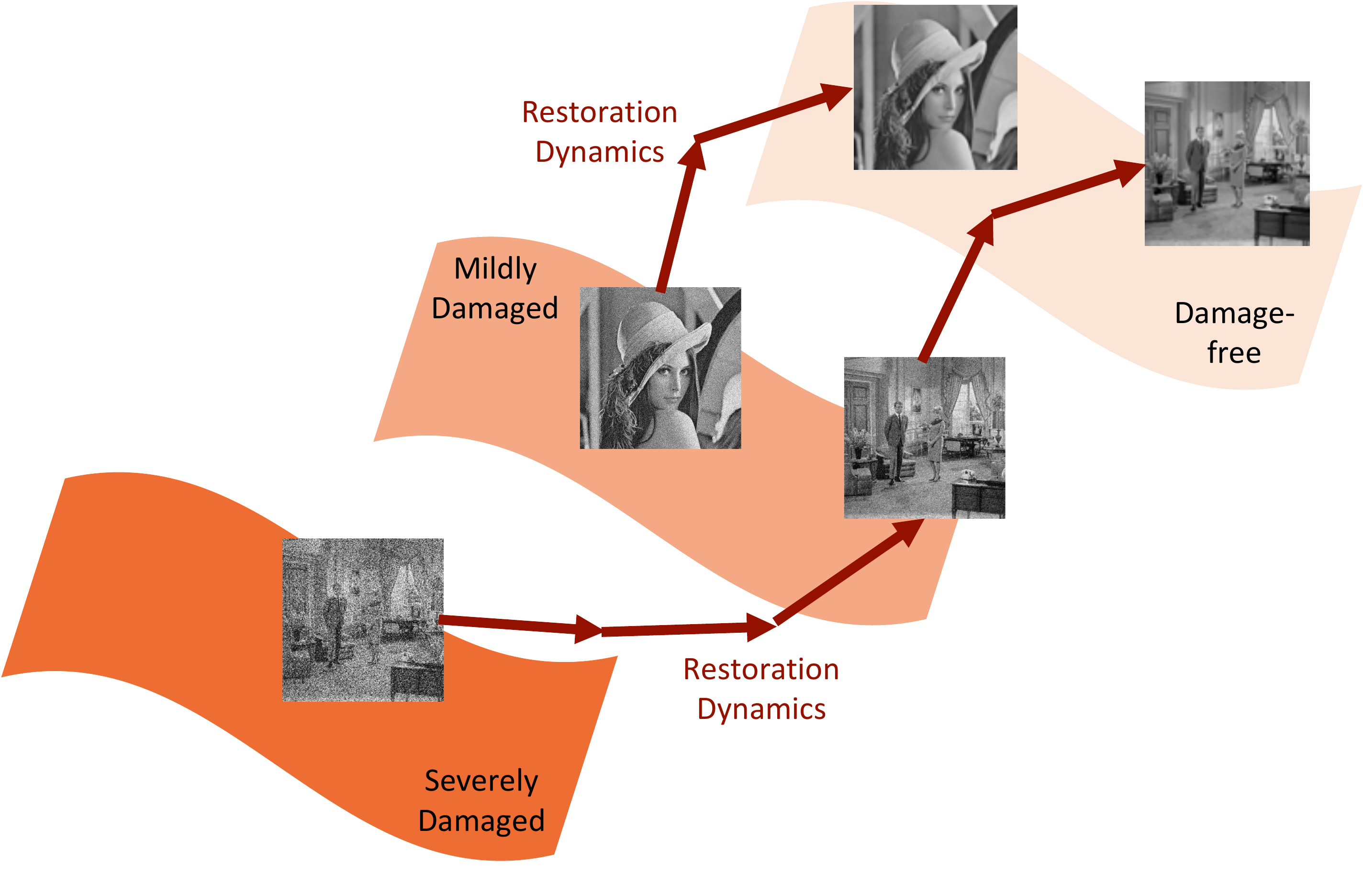}
	\caption{The proposed moving endpoint control model: evolving a learned reconstruction dynamics and ending at high-quality images.}
	\label{model}
\end{figure}

We first cast the model in the continuum setting. Let $x_0$ be an observed degraded image
and $y$ be its corresponding damage-free counterpart. We want to learn
a time-independent dynamic system $\dot{X}=f(X(t),w)$ with parameters $w$
so that $X(0)=x$ and $X(\tau)\approx y$ for some $\tau>0$. See
Fig. \ref{model} for an illustration of our idea. The reason that we
do not require $X(\tau)=y$ is to avoid over-fitting. For varied
degradation levels and different images, the optimal terminal time $\tau$
of the dynamics may vary. Therefore, we need to include the variable $\tau$
in the learning process as well. The learning of the dynamic system and
the terminal time can be gracefully casted as the following
moving endpoint control problem:

\begin{align}
\min_{w,\tau(x)}&\ L(X(\tau),y)+\int_{0}^{\tau(x)} R(w(t),t)dt \nonumber \\
s.t.&\ \dot{X}=f(X(t),w(t)), t\in(0,\tau(x)) \\
&\ X(0)=x.\nonumber
\end{align}\label{Model:Continuum}

Different from the previous control problem, in our model
the terminal time $\tau$ is also a parameter to be optimized and it depends on the data $x$.
The dynamic system $\dot{X}=f(X(t),w)$ is modeled by a recurrent neural network (RNN)
with a residual connection, which can be understood as a residual network with
shared weights \citep{Liao2016Bridging}. We shall refer to this RNN as
the \textit{restoration unit}. In order to learn the terminal time of
the dynamics, we adopt a \textit{policy network} to adaptively determine
an optimal stopping time. Our learning framework is demonstrated
in Fig. \ref{rlrnn}. We note that the above moving endpoint control problem
can be regarded as the penalized version of the well-known fixed endpoint control
problem in optimal control \citep{Evans2005An}, where instead of penalizing
the difference between $X(\tau)$ and $y$, the constraint $X(\tau)=y$ is
strictly enforced.

\begin{figure}[htp!]
	\centering
	\begin{subfigure}[t]{0.22\textwidth}
		\centering
		\includegraphics[width=1\textwidth]{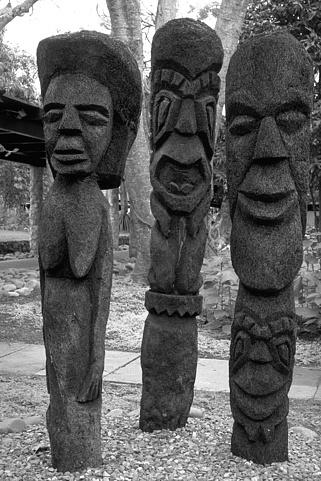}
		\subcaption*{Ground Truth}
	\end{subfigure}
	\quad
	\begin{subfigure}[t]{0.22\textwidth}
		\centering
		\includegraphics[width=1\textwidth]{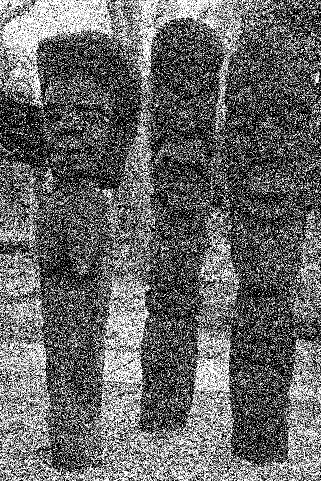}
		\subcaption*{Noisy Input, 10.72dB}
	\end{subfigure}
	\quad
	\begin{subfigure}[t]{0.22\textwidth}
		\centering
		\includegraphics[width=1\textwidth]{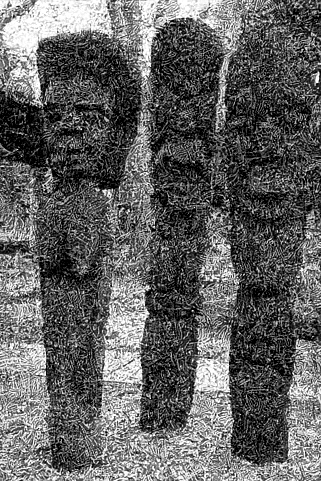}
		\subcaption*{DnCNN, 14.72dB}
	\end{subfigure}
	\quad
	\begin{subfigure}[t]{0.22\textwidth}
		\centering
		\includegraphics[width=1\textwidth]{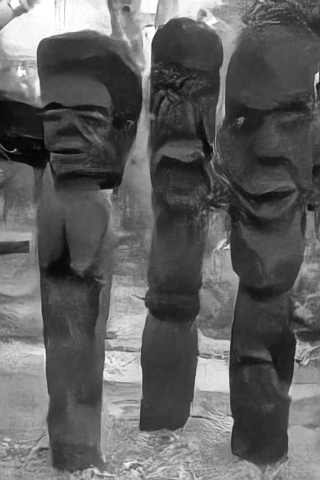}
		\subcaption*{DURR, 21.00dB}
	\end{subfigure}\\
	\centering
	\begin{subfigure}[t]{0.22\textwidth}
		\centering
		\includegraphics[width=1\textwidth]{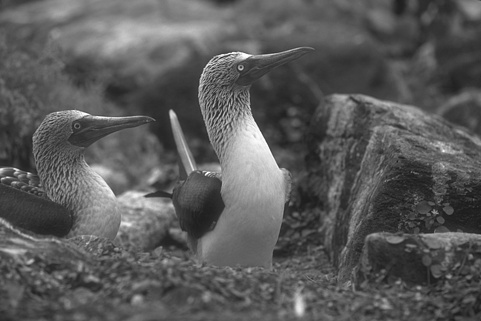}
		\subcaption*{Ground Truth}
	\end{subfigure}
	\quad
	\begin{subfigure}[t]{0.22\textwidth}
		\centering
		\includegraphics[width=1\textwidth]{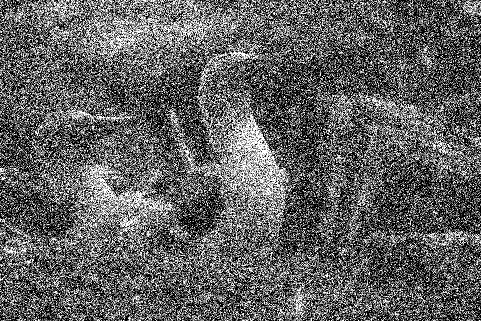}
		\subcaption*{Noisy Input, 10.48dB}
	\end{subfigure}
	\quad
	\begin{subfigure}[t]{0.22\textwidth}
		\centering
		\includegraphics[width=1\textwidth]{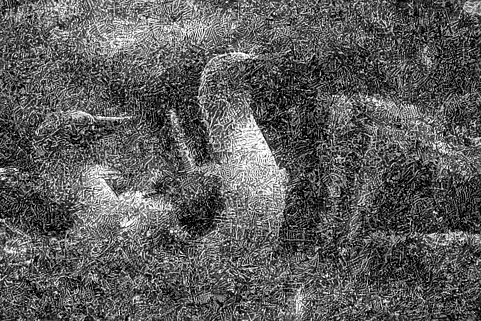}
		\subcaption*{DnCNN, 14.46dB}
	\end{subfigure}
	\quad
	\begin{subfigure}[t]{0.22\textwidth}
		\centering
		\includegraphics[width=1\textwidth]{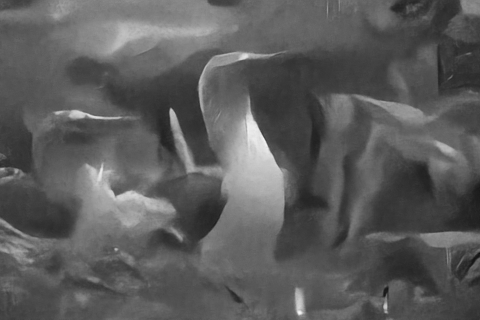}
		\subcaption*{DURR, 24.94dB}
	\end{subfigure}\\
	\caption{Denoising results of images from BSD68 under
		extreme noise conditions not seen in training data ($\sigma=95$).}
	\label{appeal}
\end{figure}

In short, we summarize our contribution as following:
\begin{itemize}
	\item  We are the first to use convolutional RNN for image restoration
	with unknown degradation levels, where the unfolding time of the RNN
	is determined dynamically at run-time by a policy unit (could be either
	handcrafted or RL-based).
	\item The proposed model achieves state-of-the-art performances
	with significantly less parameters and better running efficiencies
	than some of the state-of-the-art models.
	\item  We reveal the relationship between the generalization power
	and unfolding time of the RNN by extensive experiments.
	The proposed model, DURR, has strong generalization to images
	with varied degradation levels and even to the degradation level
	that is unseen by the model during training (Fig. \ref{appeal}).
	\item The DURR is able to well handle real image denoising
	without further modification. Qualitative results have shown that
	our processed images have better visual quality,
	especially sharper details compared to others.
\end{itemize}

\section{Method}

The proposed architecture, \textit{i.e.} DURR, contains an RNN (called the restoration unit) imitating a nonlinear diffusion for image restoration, and a deep policy network (policy unit) to determine the terminal time of the RNN. In this section, we discuss the training of the two components based on our moving endpoint control formulation. As will be elaborated, we first rain the restoration unit to determine $\omega$, and then train the policy unit to estimate $\tau(x)$.

\subsection{Training the Restoration Unit}

If the terminal time $\tau$ for every input $x_i$ is given (\textit{i.e.} given
a certain policy), the restoration unit can be optimized accordingly.
We would like to show in this section that the policy used during training
greatly influence the performance and the generalization ability of
the restoration unit. More specifically, a restoration unit can be better trained
by a good policy.

The simplest policy is to fix the loop time $\tau$ as a constant for
every input. We name such policy as ``naive policy''. A more reasonable policy is
to manually assign an unfolding time for each degradation level
during training. We shall call this policy the ``refined policy''. Since we
have not trained the policy unit yet, to evaluate the performance of the trained
restoration units, we manually pick the output image with the highest PSNR
(\textit{i.e.} the peak PSNR).

\begin{wrapfigure}{r}{3.2in}
	\includegraphics[width=3.2in]{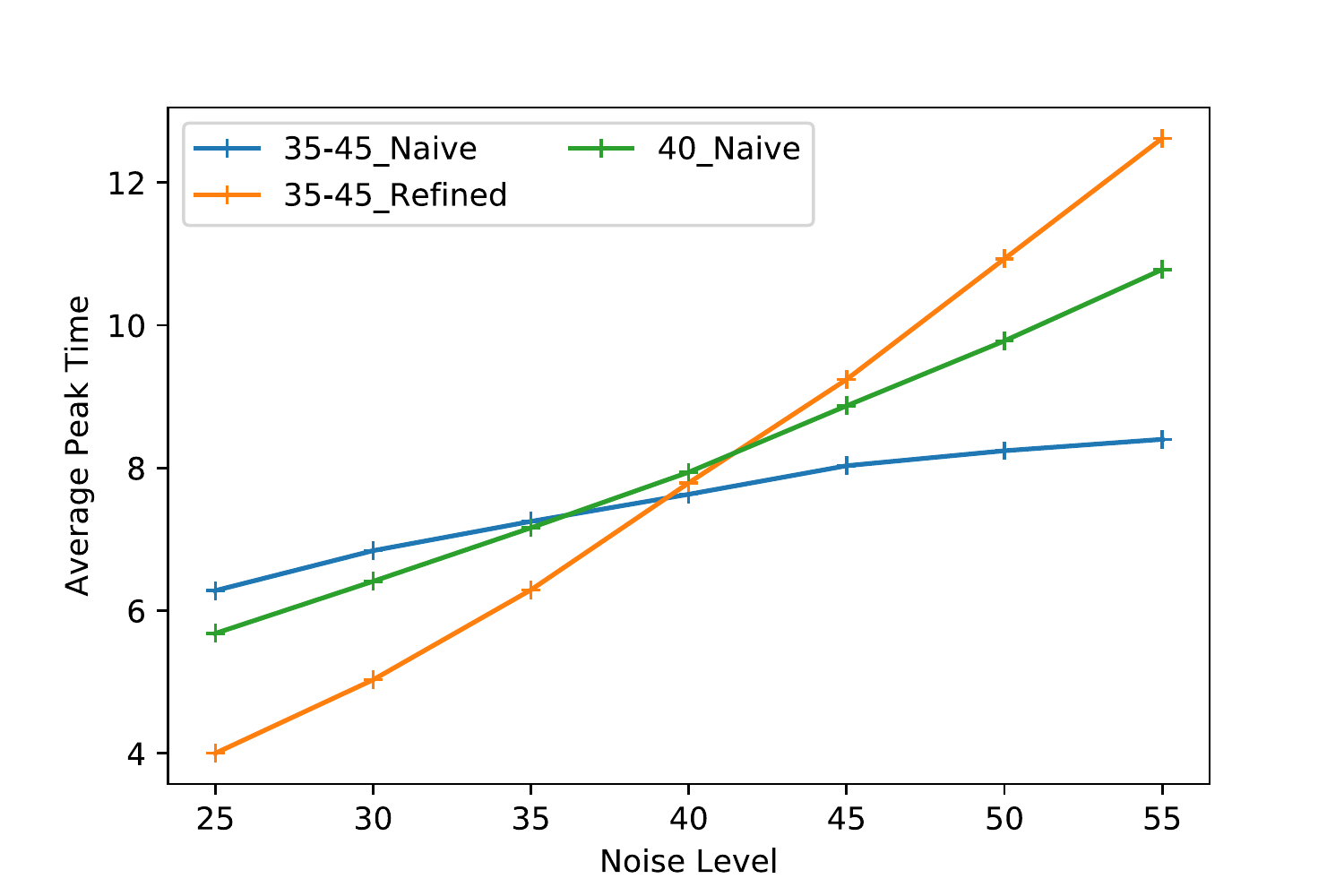}
	\caption{Average peak time on BSD68 with different training strategies.}
	\label{pre2}
\end{wrapfigure}

We take denoising as an example here. The peak PSNRs of
the restoration unit trained with different policies are listed
in Table. \ref{pre1}. Fig. \ref{pre2} illustrates the average loop times
when the peak PSNRs appear. The training is done on both
single noise level ($\sigma=40$) and multiple noise levels ($\sigma=35,45$).
For the refined policy, the noise levels and the associated loop times are
(35, 6), (45, 9). For the naive policy, we always fix the loop times to 8.

\begin{table}[ht]
	\caption{Average peak PSNR on BSD68 with different training strategies.}
	\label{pre1}
	\centering
	\begin{tabular}{ll|lllllllllll}
		\midrule
		\multicolumn{2}{c|}{Strategy}&\multicolumn{7}{c}{Noise Level}\\
		\midrule
		Training Noise&Policy &$25$&$30$&$35$&$40$&$45$&$50$&$55$\\
		\midrule
		40      &Naive   & 28.61         & 28.13        & 27.62        &\textbf{27.19} & 26.57        & 26.17        & 24.00\\
		35, 45  &Naive   & 27.74         & 27.17        & 26.66        & 26.24         & 26.75        & 25.61        & 24.75 \\
		35, 45  &Refined & \textbf{29.14}&\textbf{28.33}&\textbf{27.67}&\textbf{27.19} &\textbf{27.69}&\textbf{26.61}&\textbf{25.88} \\
		\bottomrule
	\end{tabular}
\end{table}

As we can see, the refined policy brings the best performance on all
the noise levels including 40. The restoration unit trained for specific
noise level (\textit{i.e.} $\sigma=40$) is only comparable to the one with
refined policy on noise level 40. The restoration unit trained on multiple noise levels
with naive policy has the worst performance.

These results indicate that the restoration unit has the potential to
generalize on unseen degradation levels when trained with good policies.
According to Fig. \ref{pre2}, the generalization reflects
on the loop times of the restoration unit. It can be observed that the model
with steeper slopes have stronger ability to generalize as well as
better performances.

According to these results, the restoration unit we used in DURR is trained
using the refined policy. More specifically, for image denoising, the noise level
and the associated loop times are set to (25, 4), (35, 6), (45, 9), and (55, 12). For JPEG image deblocking, the quality factor (QF) and
the associated loop times are set to (20, 6) and (30, 4).

\subsection{Training The Policy Unit}

We discuss two approaches that can be used as policy unit:

\textbf{Handcraft policy:}  Previous work \citep{mrazek2003selection}
has proposed a handcraft policy that selects a terminal time
which optimizes the correlation of the signal and noise in
the filtered image. This criterion can be used directly as our
policy unit, but the independency of signal and noise may not hold
for some restoration tasks such as real image denoising, which has
higher noise level in the low-light regions, and JPEG image deblocking,
in which artifacts are highly related to the original image.
Another potential stopping criterion of the diffusion is
no-reference image quality assessment \citep{mittal2012no},
which can provide quality assessment to a processed image
without the ground truth image. However, to the best of our knowledge,
the performaces of these assessments are still far from satisfactory.
Because of the limitations of the handcraft policies,
we will not include them in our experiments.

\textbf{Reinforcement learning based policy:} We start with a discretization of the moving endpoint problem
(\ref{Model:Continuum}) on the dataset $\{(x_i,y_i)|i=1, 2, \cdots, d\}$,
where $\{x_i\}$ are degraded observations of the damage-free images $\{y_i\}$.
The discrete moving endpoint control problem is given as follows:
	
\begin{align}
\min_{w,\{N_i\}_{i=1}^d} &\ R(w) + \lambda \sum_{i=1}^{d} L(X_{N_i}^i,y_i)\nonumber \\
s.t.&\ X_n^i = X_{n-1}^i+\Delta t f(X_{n-1}^i,w), n = 1, 2, \cdots, N_i, (i = 1, 2, \cdots, d)\\
&\ X_0^i = x_i, i = 1, 2, \cdots, d .\nonumber
\end{align}\label{Model:Discrete}
	
Here, $X_n^i = X_{n-1}^i+\Delta t f(X_{n-1}^i,w)$ is the
forward Euler approximation of the dynamics $\dot{X}=f(X(t),w)$. The terminal time $\{N_i\}$ is determined by
a policy network $P(x,\theta)$, where $x$ is the output of
the restoration unit at each iteration and $\theta$ the weight.
In other words, the role of the policy network is to stop the iteration of
the restoration unit when an ideal image restoration result is achieved.
The reward function of the policy unit can be naturally defined by

\begin{equation}
r(\{X_n^i\}) =  \left\{
\begin{array}{lr}
\lambda \left(L(x_{n-1},y_i)-L(x_n,y_i)\right)&\text{If choose to continue}\\
0& \text{Otherwise}
\end{array}
\right.
\label{rewardfunc}
\end{equation}

In order to solve the problem (\ref{Model:Discrete}),
we need to optimize two networks simultaneously, \textit{i.e.} the restoration unit
and the policy unit. The first is an restoration unit which approximates
the controlled dynamics and the other is the policy unit to give
the optimized terminating conditions. The objective function we use
to optimize the policy network can be written as

\begin{align}
J = \mathbb{E}_{X\sim \pi_\theta} \sum_{n}^{N_i}[r(\{X_n^i,w\})],
\end{align}

where $\pi_\theta$ denotes the distribution of
the trajectories $X=\{X^i_n,n=1,\ldots,N_i, i=1,\ldots,d\}$ under
the policy network $P(\cdot,\theta)$. Thus, reinforcement learning
techniques can be used here to learn a neural network to work as
a policy unit. We utilize Deep Q-learning \citep{mnih2015human}
as our learning strategy and denote this approach simply as \textbf{DURR}.

\section{Experiments}

\subsection{Experiment Settings}

In all denoising experiments, we follow the same settings as in
\cite{Chen2017Trainable,zhang2017beyond,lefkimmiatis2017universal}.
All models are evaluated using the mean PSNR as the quantitative metric
on the BSD68 \citep{MartinFTM01}. The training set and test set of
the BSD500 (400 images in total) are used for training.
Both the training and evaluation process are done on gray-scale images.

The restoration unit is a simple U-Net \citep{ronneberger2015u} style
fully convolutional neural network. For the training process of the restoration unit,
the noise levels of 25, 35, 45 and 55 are used. Images are
cut into $64 \times 64$ patches, and the batch-size is set to 24.
The Adam optimizer with the learning rate 1e-3 is adopted and
the learning rate is scaled down by a factor of 10 on training plateaux.

The policy unit is composed of two ResUnit and an LSTM cell.
For the policy unit training, we utilize the reward function
in Eq.\ref{rewardfunc}. For training the policy unit,
an RMSprop optimizer with learning rate 1e-4 is adopted.
We've also tested other network structures, these tests and
the detailed network structures of our model are demonstrated
in the appendix.

In all JPEG deblocking experiments, we follow the settings as in
\cite{zhang2017beyond,zhang2018dmcnn}. All models are evaluated using
the mean PSNR as the quantitative metric on
the LIVE1 dataset \citep{sheikh2005live}. The training set and testing set
of BSD500 are used for training. Both the training and evaluation processes
are done on the Y channel (the luminance channel) of the YCbCr color space.
The images with quality factors 20 and 30 are used during
the training process of the restoration unit. All other parameter settings
are the same as in the denoising experiments.

\subsection{The Complete DURR}

After training the restoration unit, the policy unit is trained
using the Deep Q-learning algorithm stated above until
full convergence. Then the two units are combined to form
the complete DURR model.

\subsubsection{Image Denoising}

We select DnCNN-B\citep{zhang2017beyond} and UNLNet$_5$
\citep{lefkimmiatis2017universal} for comparisons
since these models are designed for blind image denoising. Moreover,
we also compare our model with non-learning-based algorithms
BM3D \citep{dabov2007image} and WNMM \citep{gu2014weighted}.
The noise levels are assumed known for BM3D and WNMM due to their requirements.
Comparison results are shown in Table \ref{sample-table}.

\begin{figure}[htp!]
	\centering
	\begin{subfigure}[t]{0.3\textwidth}
		\centering
		\includegraphics[width=1\textwidth]{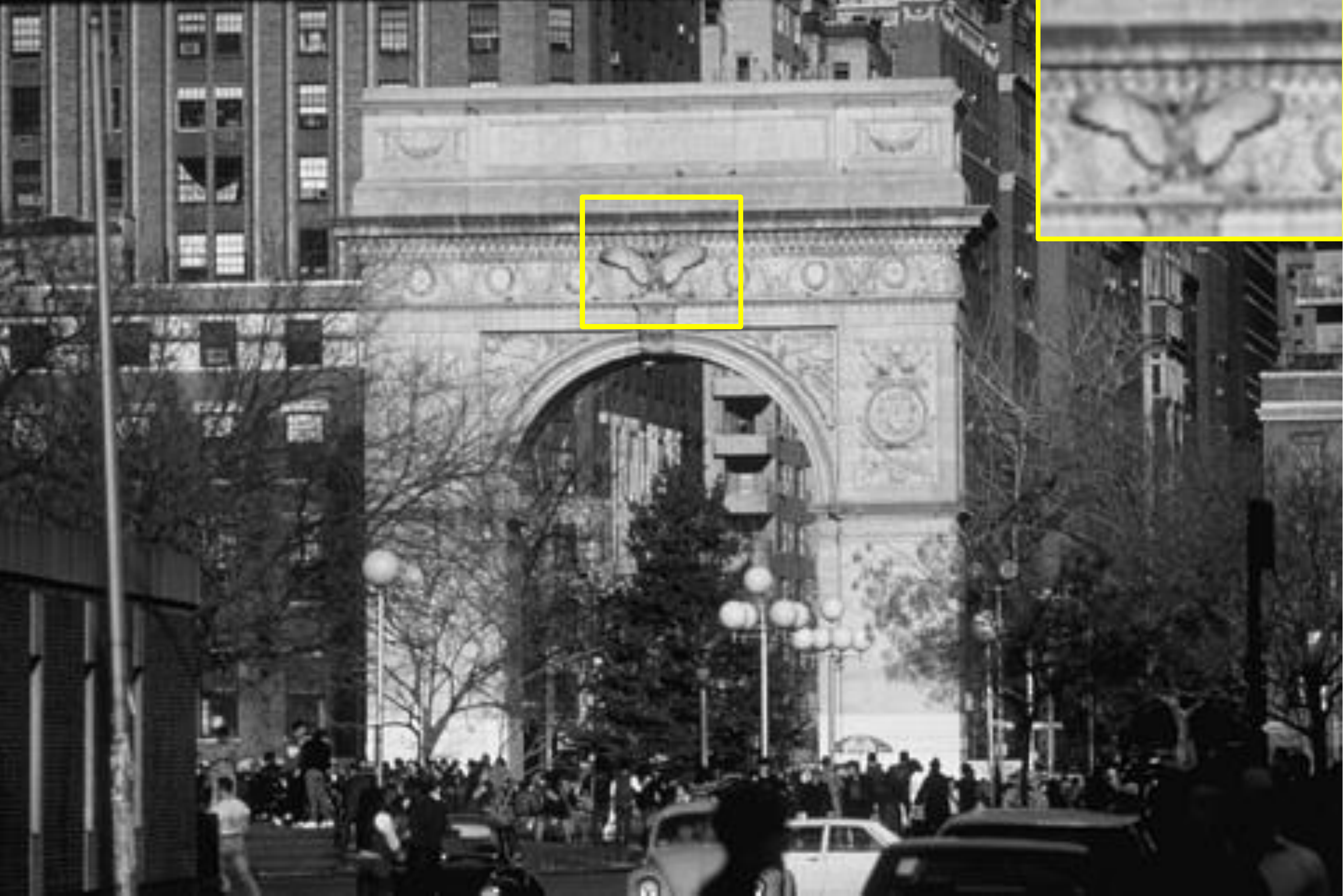}
		\subcaption*{Ground Truth}
	\end{subfigure}
	\quad
	\begin{subfigure}[t]{0.3\textwidth}
		\centering
		\includegraphics[width=1\textwidth]{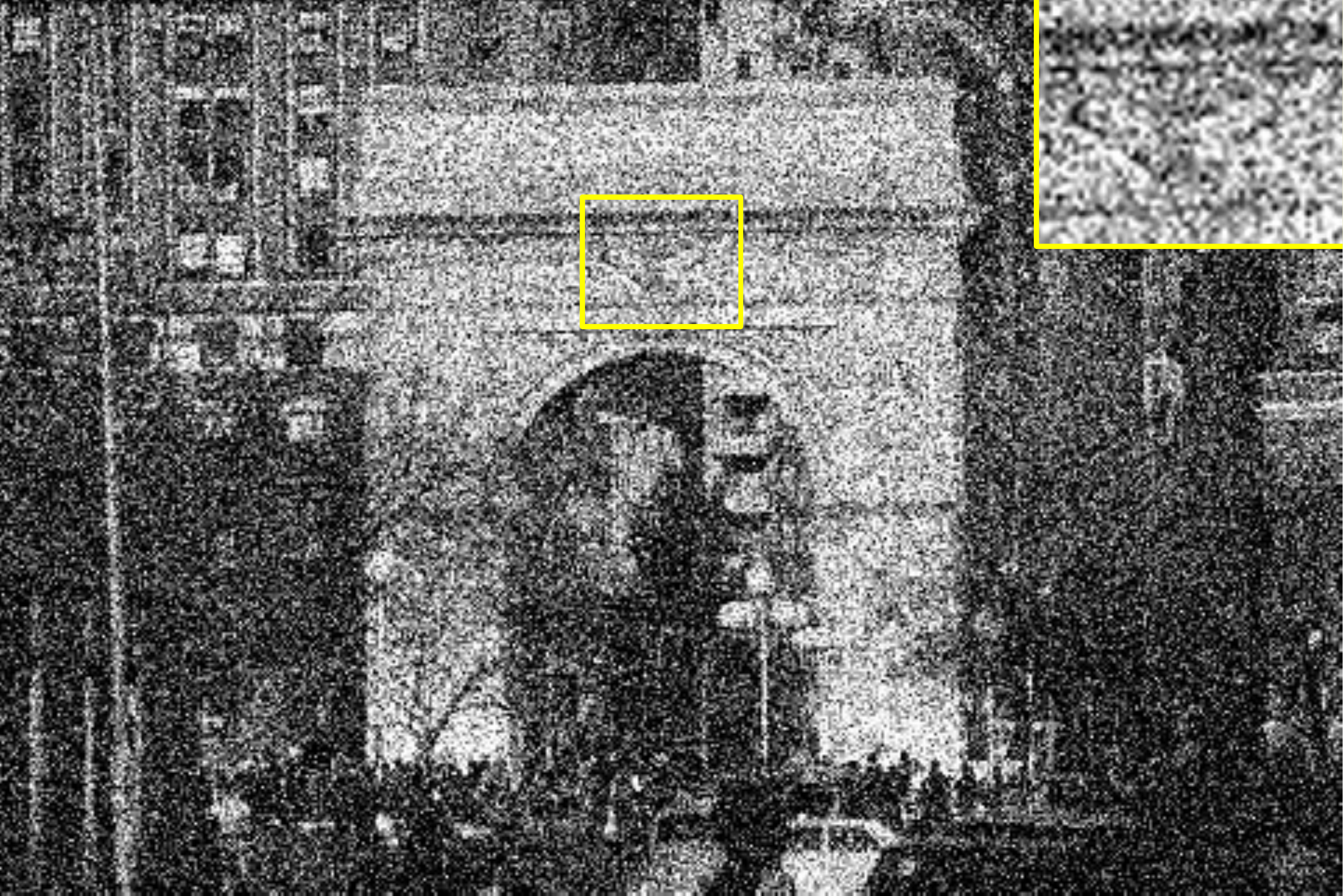}
		\subcaption*{Noisy Input, 13.22dB}
	\end{subfigure}
	\quad
	\begin{subfigure}[t]{0.3\textwidth}
		\centering
		\includegraphics[width=1\textwidth]{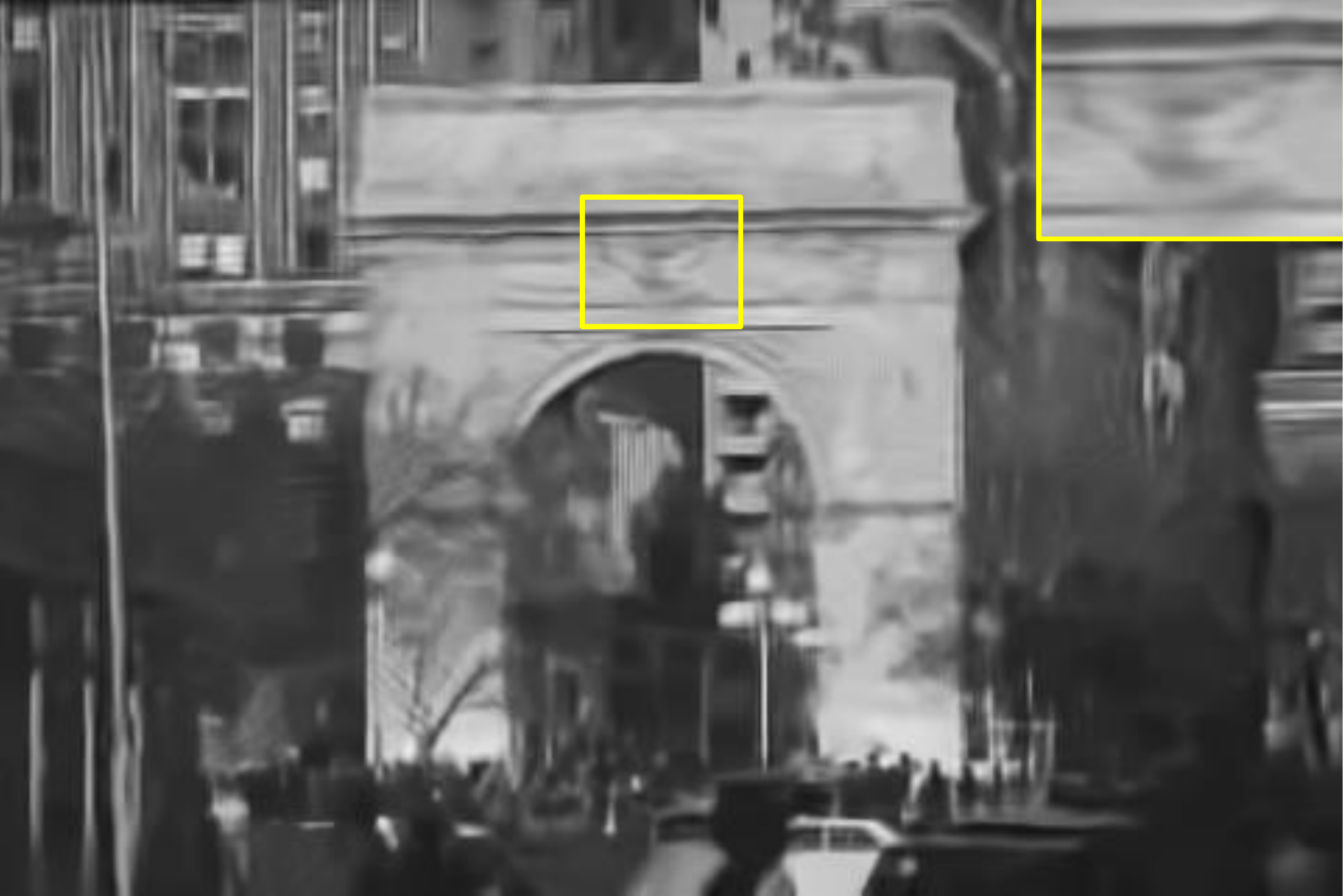}
		\subcaption*{(a) BM3D, 21.35dB}
	\end{subfigure}\\
	\begin{subfigure}[t]{0.3\textwidth}
		\centering
		\includegraphics[width=1\textwidth]{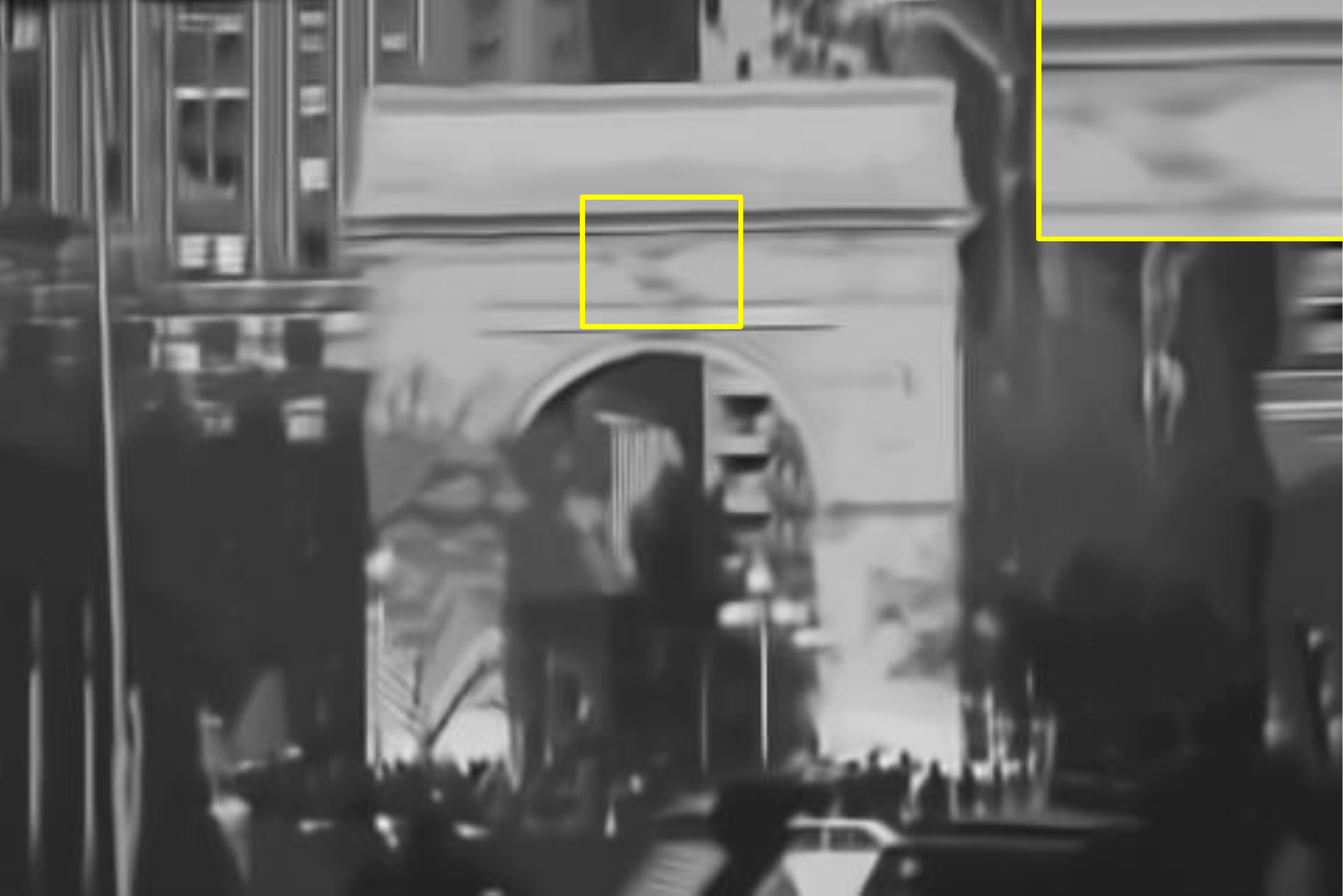}
		\subcaption*{(b) WNMM, 21.02dB}
	\end{subfigure}
	\quad
	\begin{subfigure}[t]{0.3\textwidth}
		\centering
		\includegraphics[width=1\textwidth]{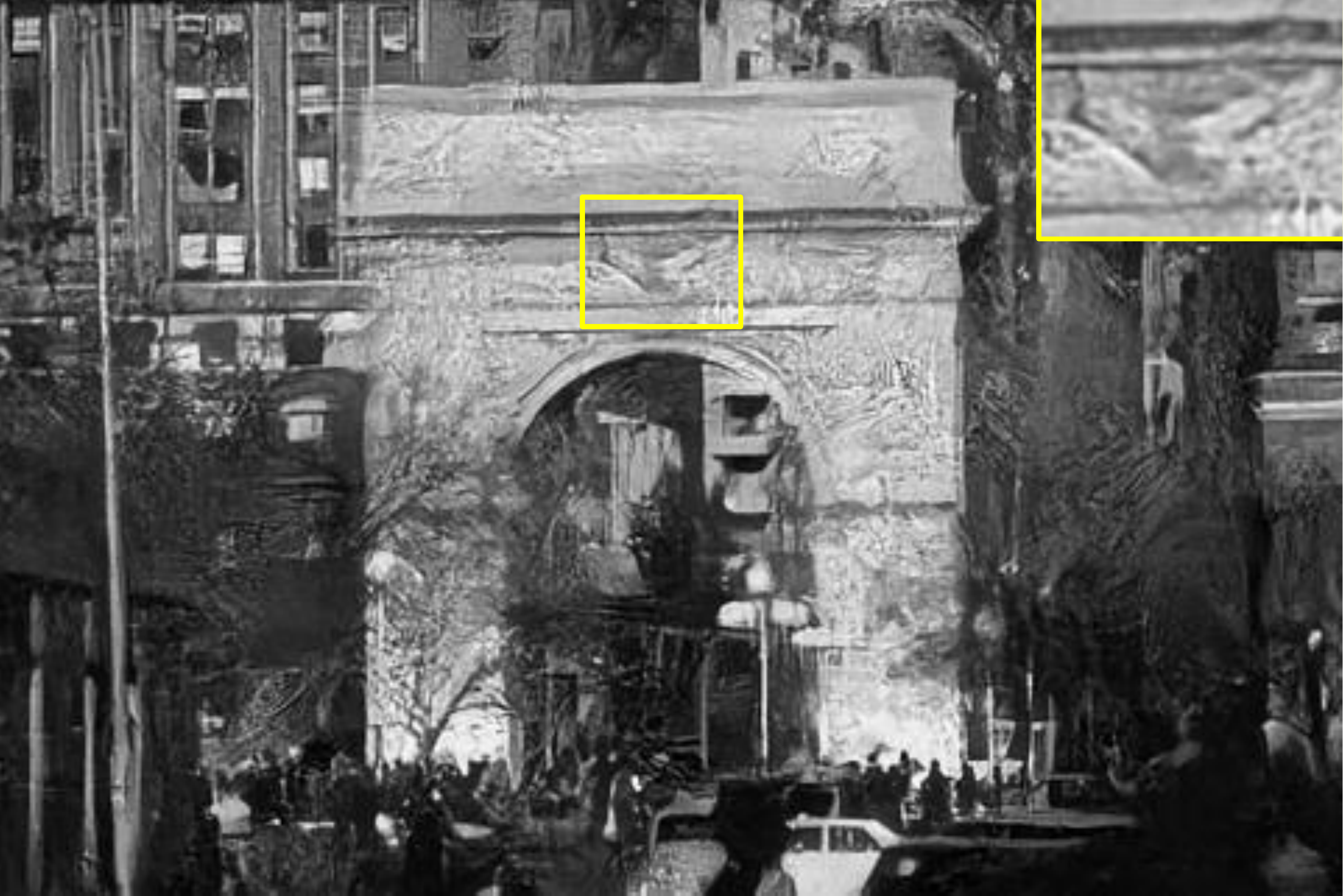}
		\subcaption*{(c) DnCNN, 21.86dB}
	\end{subfigure}
	\quad
	\begin{subfigure}[t]{0.3\textwidth}
		\centering
		\includegraphics[width=1\textwidth]{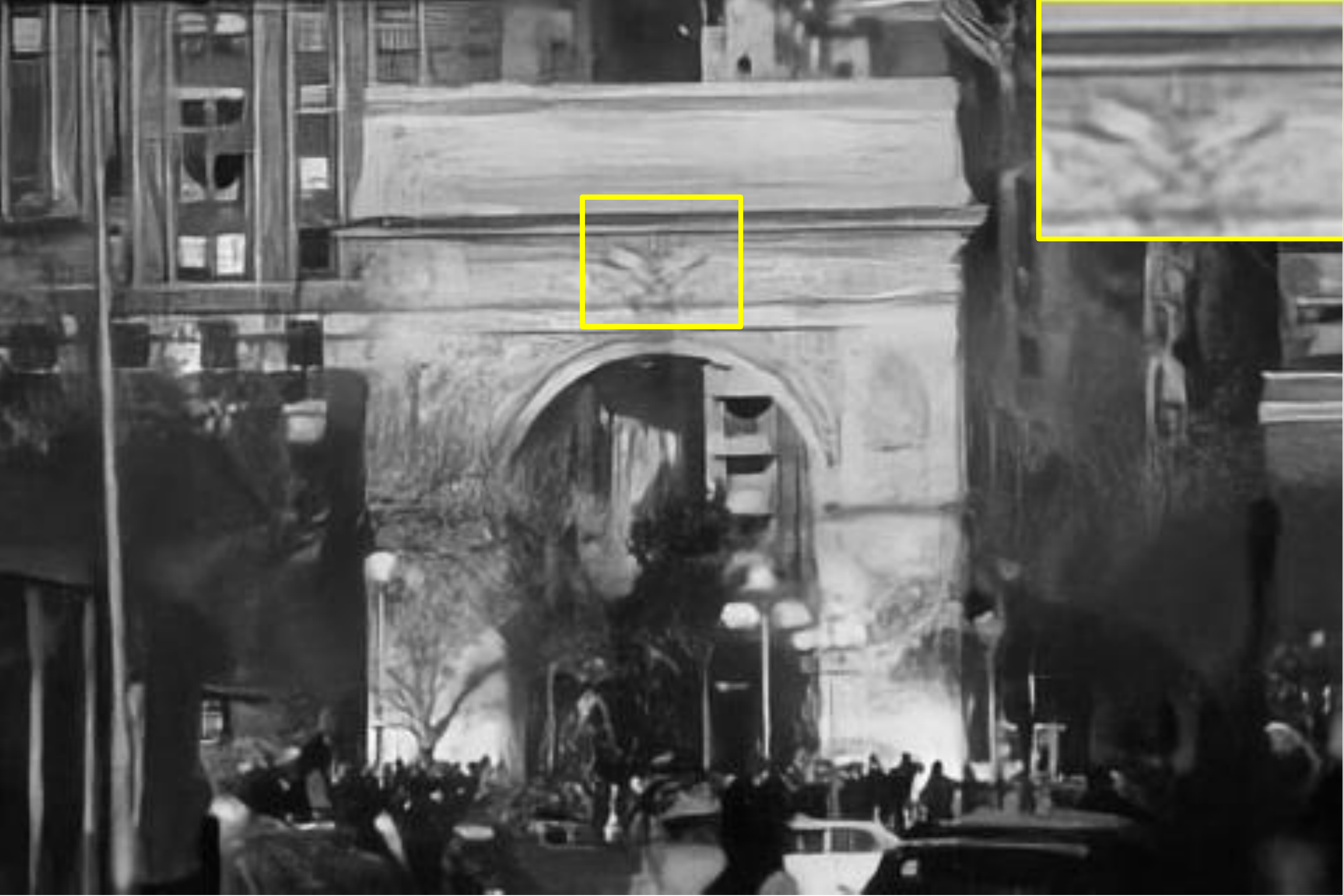}
		\subcaption*{(d) DURR, 22.84dB}
	\end{subfigure}
	\caption{Denoising results of an image from BSD68 with
		noise level 65 (unseen by both DnCNN and DURR in their
		training sets).}
	\label{test2}
\end{figure}

Despite the fact that the parameters of our model ($1.8\times10^5$ for
the restoration unit and $1.0\times10^5$ for the policy unit) is less than
the DnCNN (approximately $7.0\times10^5$), one can see that DURR outperforms
DnCNN on most of the noise-levels.
More interestingly, DURR does not degrade too much when the the noise level
goes beyond the level we used during training. The noise level $\sigma=65, 75$
is not included in the training set of both DnCNN and DURR. DnCNN reports
notable drops of PSNR when evaluated on the images with such noise levels,
while DURR only reports small drops of PSNR (see the last row of
Table \ref{sample-table} and Fig. \ref{test2}). Note that
the reason we do not provide the results of UNLNet$_5$ in
Table \ref{sample-table} is because the authors of
\cite{lefkimmiatis2017universal} has not released their codes yet,
and they only reported the noise levels from 15 to 55 in their paper.
We also want to emphasize that they trained two networks, one for
the low noise level ($5\le\sigma\le29$) and one for
higher noise level ($30\le\sigma\le55$). The reason is that
due to the use of the constraint $||y-x||_2\le \epsilon$ by
\cite{lefkimmiatis2017universal}, we should not expect the model
generalizes well to the noise levels surpasses the noise level of
the training set.

\begin{figure}[ht]
	\centering
	\begin{subfigure}[t]{0.3\textwidth}
		\centering
		\includegraphics[width=1\textwidth]{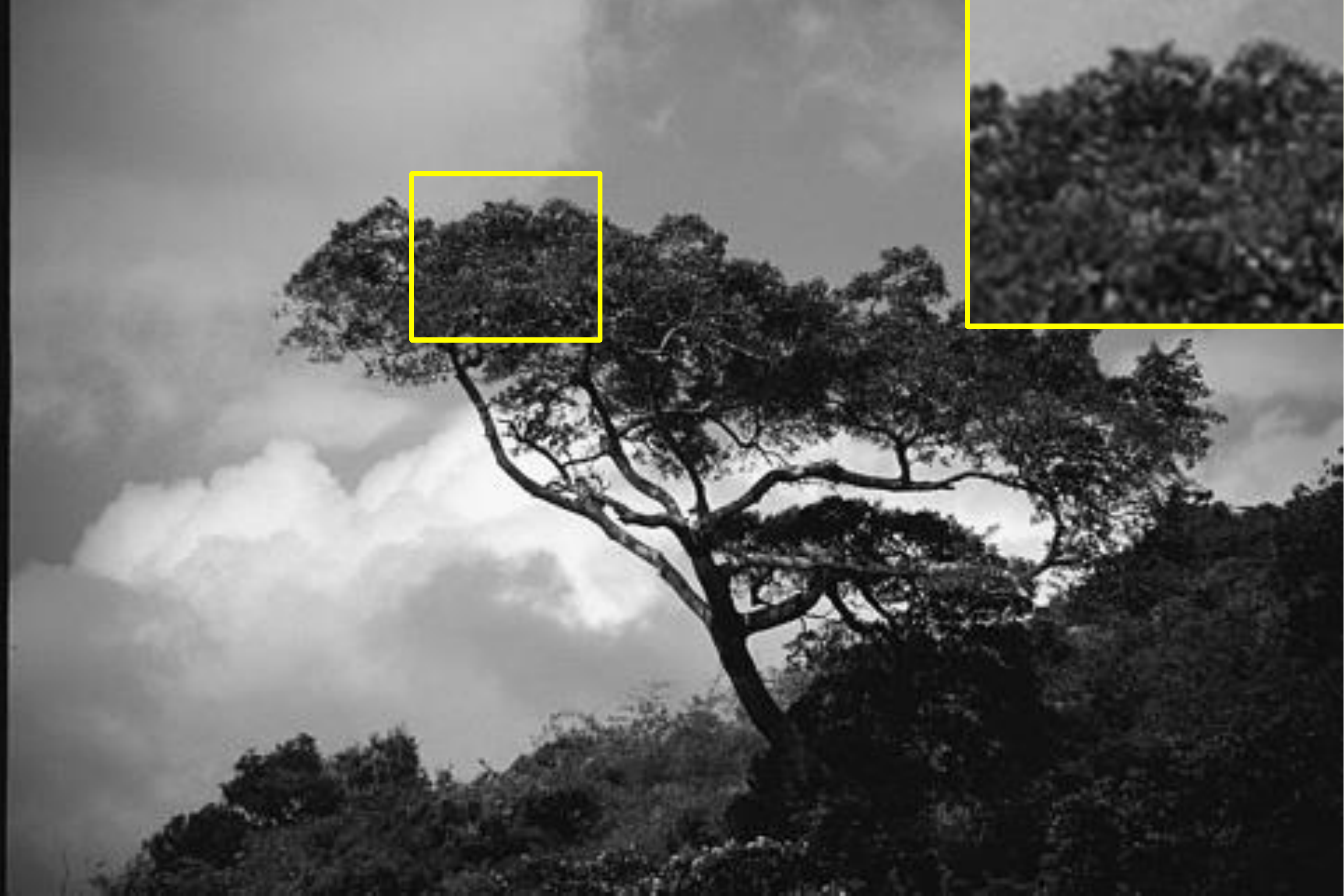}
		\subcaption*{Ground Truth}
	\end{subfigure}
	\quad
	\begin{subfigure}[t]{0.3\textwidth}
		\centering
		\includegraphics[width=1\textwidth]{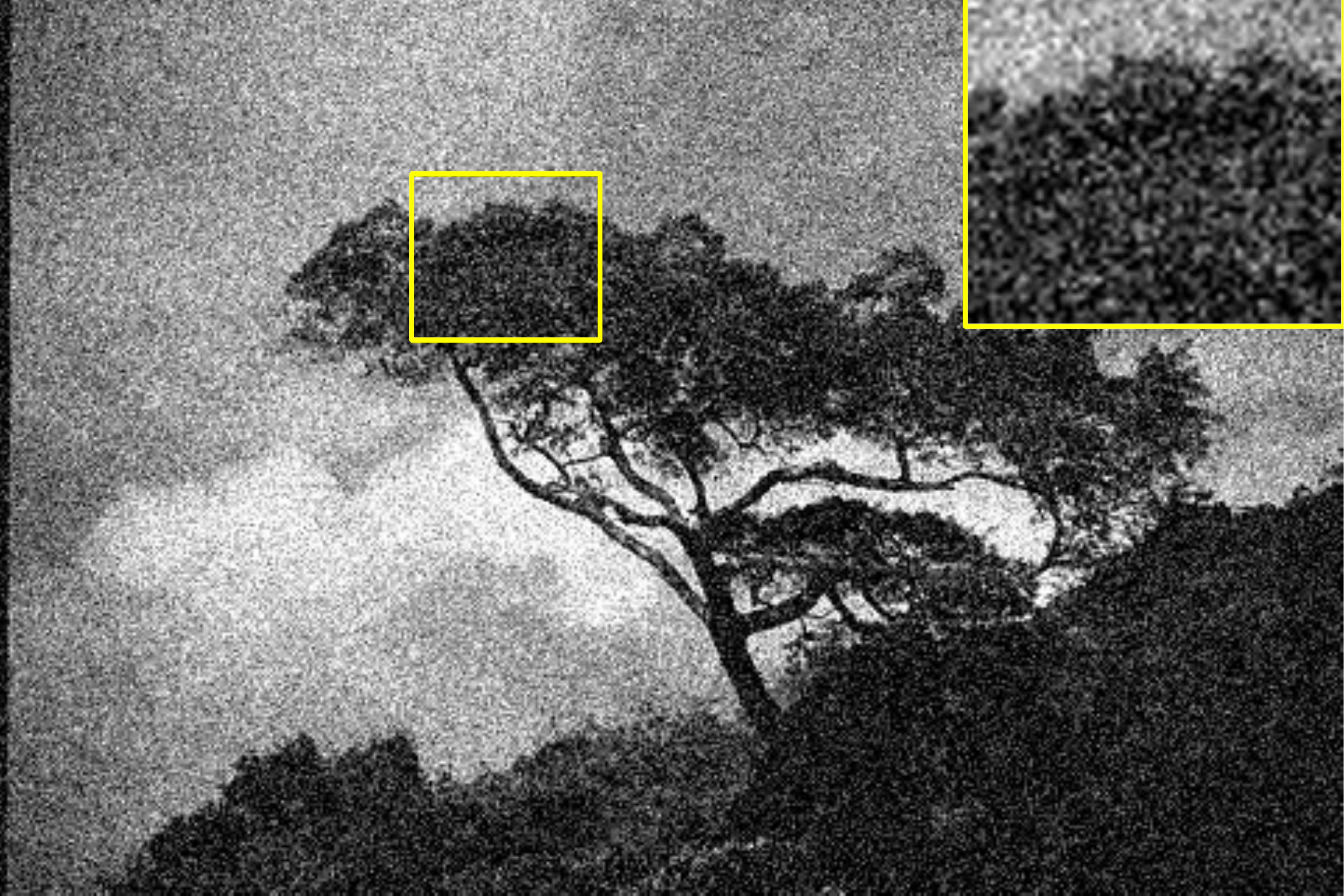}
		\subcaption*{Noisy Input, 17.84dB}
	\end{subfigure}
	\quad
	\begin{subfigure}[t]{0.3\textwidth}
		\centering
		\includegraphics[width=1\textwidth]{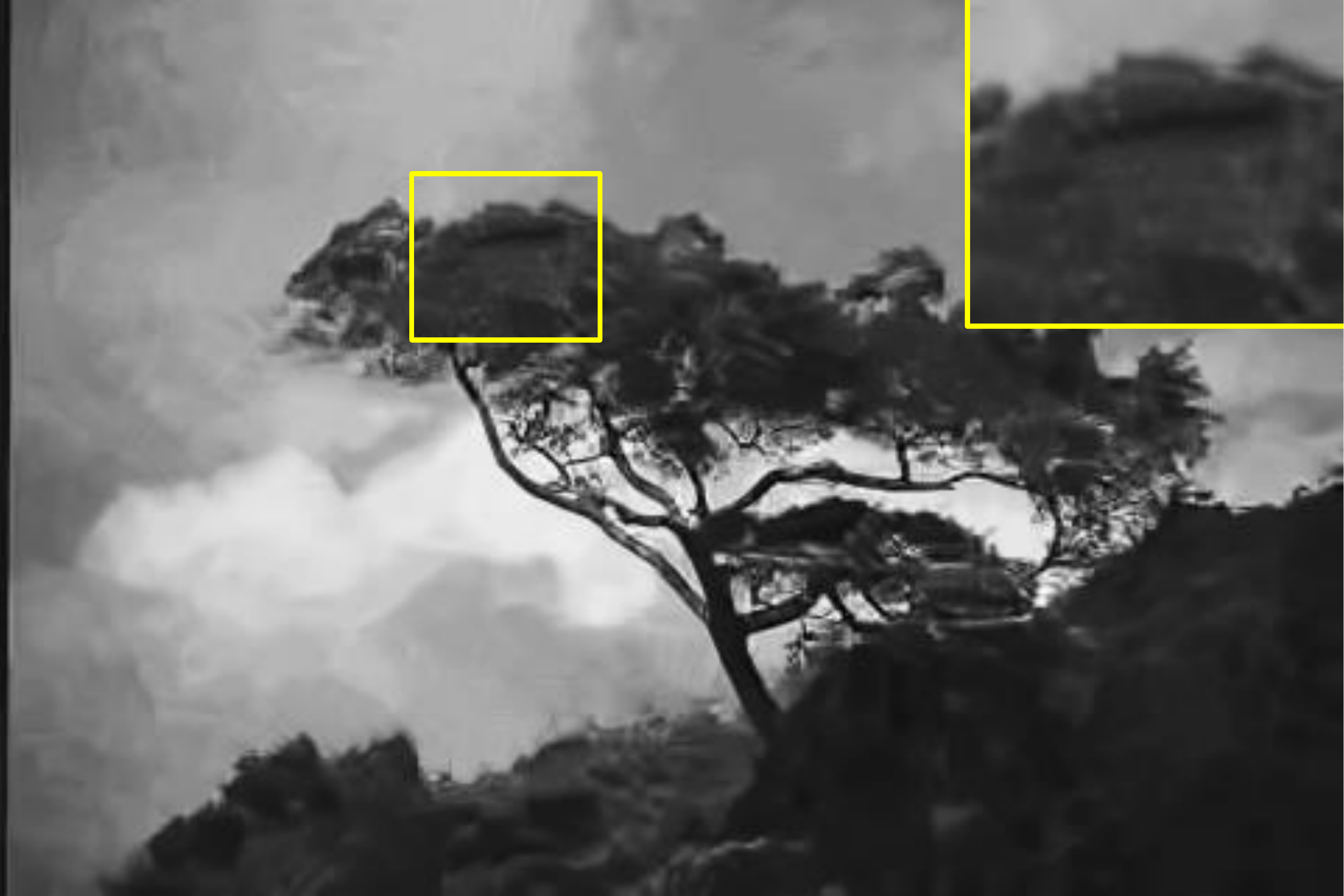}
		\subcaption*{(a) BM3D, 26.23dB}
	\end{subfigure}\\
	\begin{subfigure}[t]{0.3\textwidth}
		\centering
		\includegraphics[width=1\textwidth]{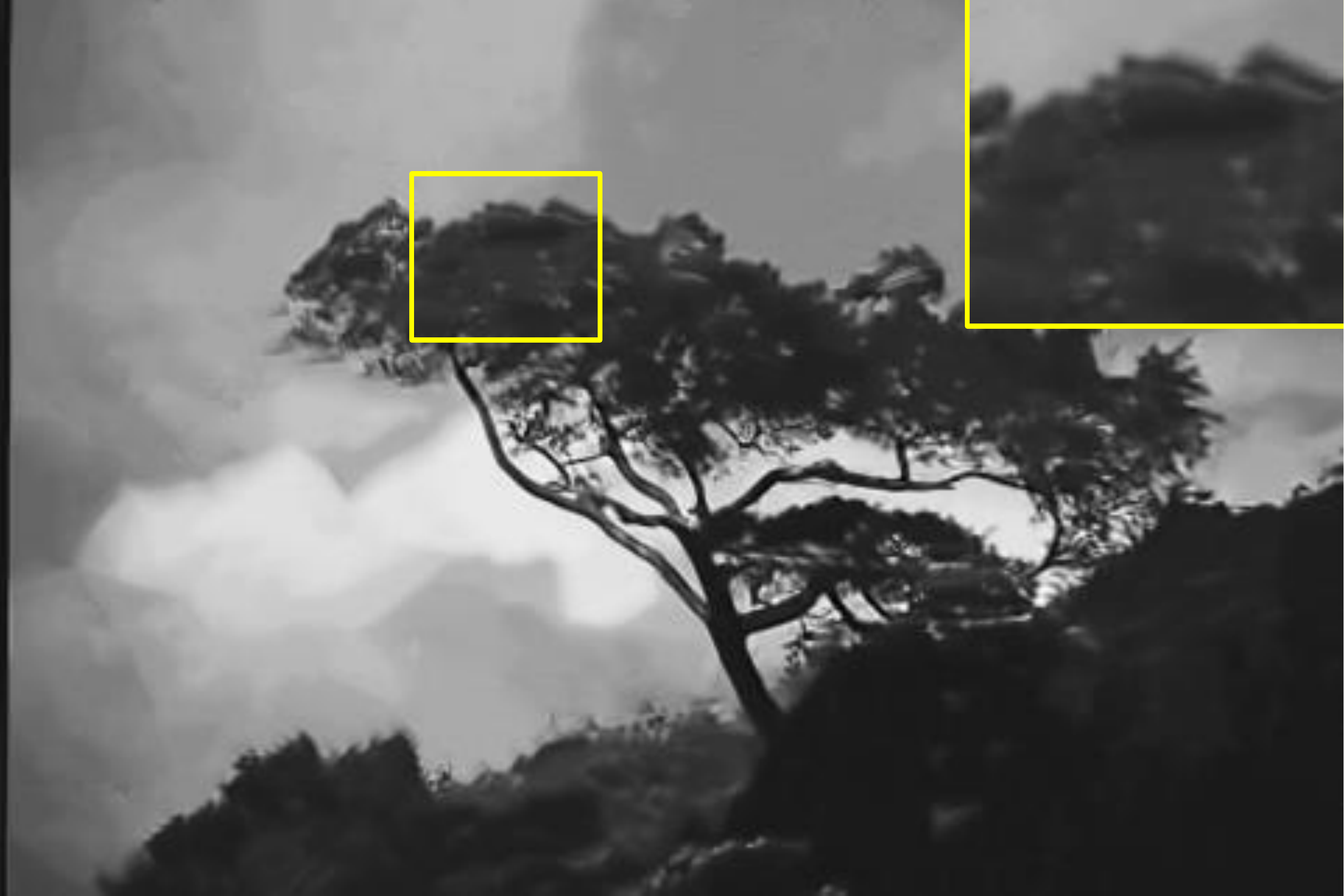}
		\subcaption*{(b) WNMM, 26.35dB}
	\end{subfigure}
	\quad
	\begin{subfigure}[t]{0.3\textwidth}
		\centering
		\includegraphics[width=1\textwidth]{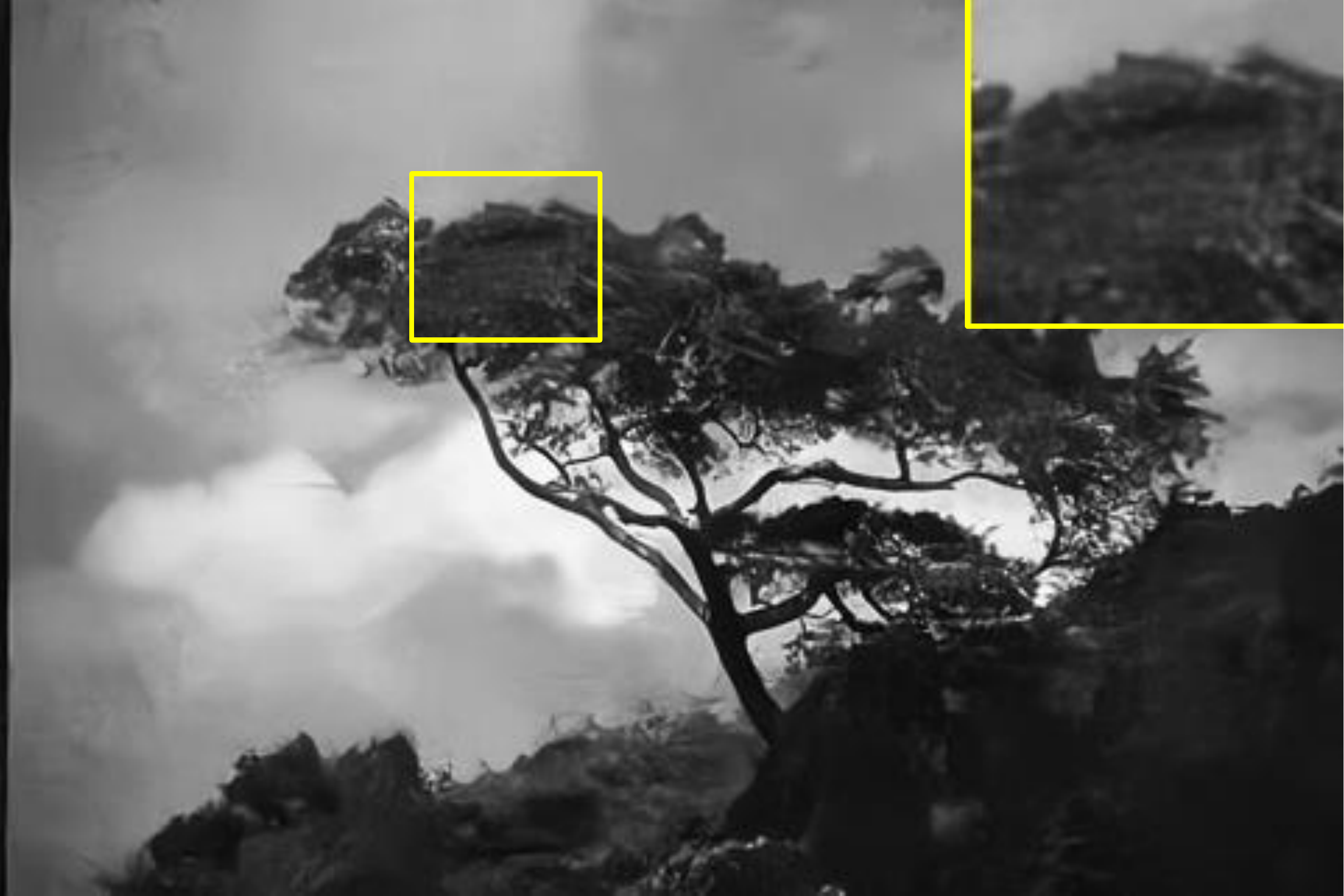}
		\subcaption*{(c) DnCNN, 27.31dB}
	\end{subfigure}
	\quad
	\begin{subfigure}[t]{0.3\textwidth}
		\centering
		\includegraphics[width=1\textwidth]{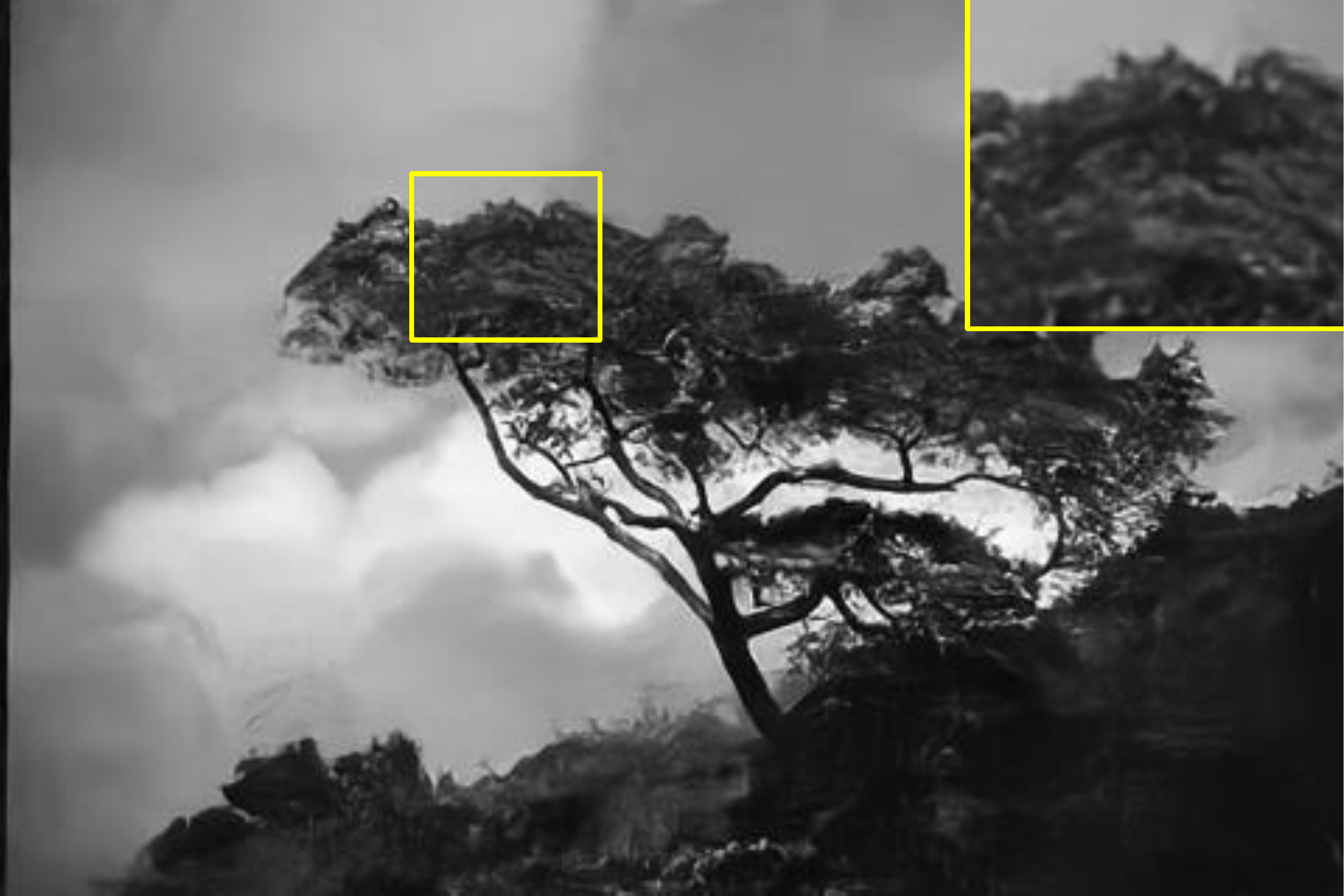}
		\subcaption*{(d) DURR, 27.42dB}
	\end{subfigure}
	\caption{Denoising results of an image from BSD68 with
		noise level 35.}
	\label{test1}
\end{figure}

For qualitative comparisons, some restored images of different models
on the BSD68 dataset are presented in Fig. \ref{test1} and Fig. \ref{test2}.
As can be seen, more details are preserved in DURR than other models.
It is worth noting that the noise level of the input image in Fig. \ref{test2}
is 65, which is unseen by both DnCNN and DURR during training. Nonetheless, DURR achieves
a significant gain of nearly 1 dB than DnCNN. Moreover, the texture on
the cameo is very well restored by DURR. These results clearly
indicate the strong generalization ability of our model.

More interestingly, due to the generalization ability in denoising, DURR
is able to handle the problem of real image denoising without
additional training. For testing, we test the images obtained
from \cite{lebrun2015noise}. We present the representative results
in Fig. \ref{real} and more results are listed in the appendix.

\begin{table}[ht]
	\centering
	\caption{Average PSNR (dB) results on the BSD68 dataset.
		Values with $^*$ means the corresponding noise level is not present in
		the training data of the model. The best results are indicated in red and the second best results are indicated in blue.}
	\label{sample-table}
	
	\begin{tabular}{l|cccccccccc}
		\toprule
		&BM3D
		&WNMM
		&DnCNN-B
		&UNLNet$_5$
		&DURR \\
		\midrule\midrule
		$\sigma = 25$ & 28.55        & 28.73        & \blue{29.15} & 28.96 & \red{29.16}     \\
		$\sigma = 35$ & 27.07        & 27.28        & \blue{27.66} & 27.50 & \red{27.72}     \\
		$\sigma = 45$ & 25.99        & 26.26        & \blue{26.62} & 26.48 & \red{26.71}     \\
		$\sigma = 55$ & 25.26        & 25.49        & \blue{25.80} & 25.64 & \red{25.91}     \\
		$\sigma = 65$ & \blue{24.69} & 24.51        & 23.40$^*$    & -     & \red{25.26$^*$} \\
		$\sigma = 75$ & 22.63        & \blue{22.71} & 18.73$^*$    & -     & \red{24.71$^*$} \\
		\bottomrule
	\end{tabular}
\end{table}

\normalsize

\subsubsection{JPEG Image Deblocking}
For deep learning based models, we select DnCNN-3 \citep{zhang2017beyond}
for comparisons since it is the only known deep model for multiple QFs deblocking.
As the AR-CNN \citep{dong2015compression} is a commonly used baseline, we re-train
the AR-CNN on a training set with mixed QFs and denote this model as AR-CNN-B.
Original AR-CNN as well as a non-learning-based method SA-DCT
\citep{foi2007pointwise} are also tested. The quality factors are assumed known
for these models.

\begin{figure}[htp!]
	\centering
	\begin{subfigure}[t]{0.195\textwidth}
		\centering
		\includegraphics[width=1\textwidth]{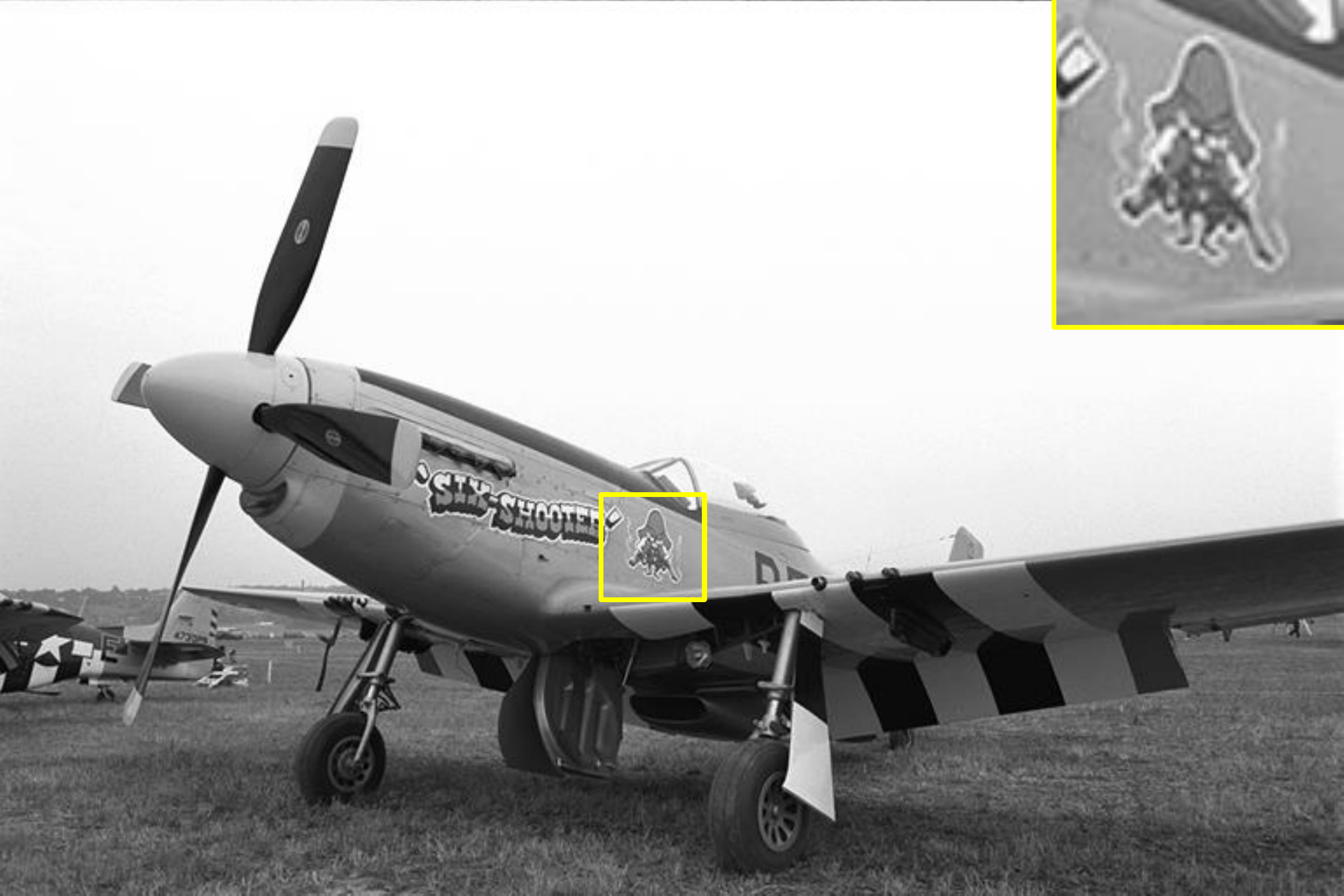}
		\subcaption*{Ground Truth}
	\end{subfigure}
	\begin{subfigure}[t]{0.195\textwidth}
		\centering
		\includegraphics[width=1\textwidth]{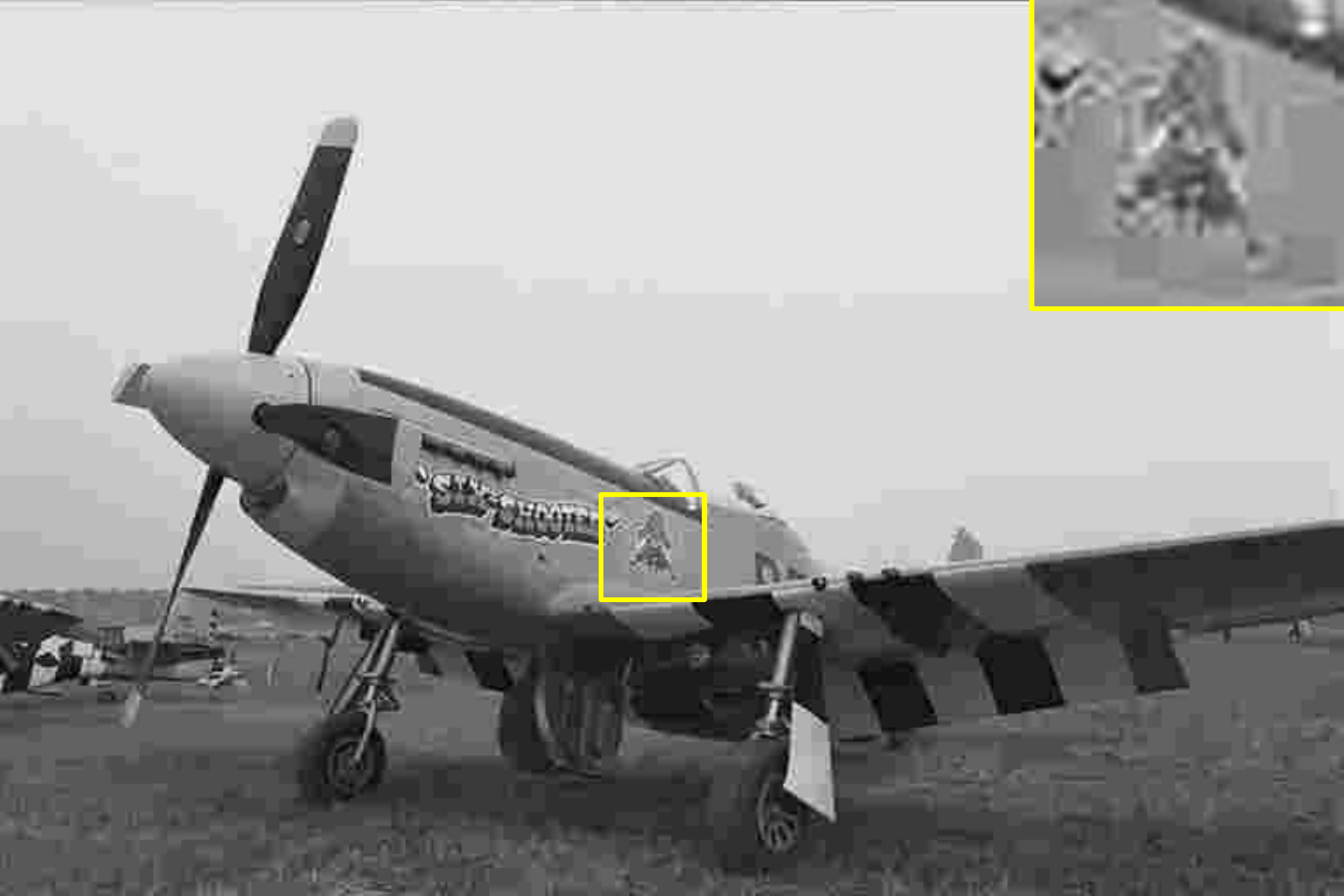}
		\subcaption*{JPEG}
	\end{subfigure}
	\begin{subfigure}[t]{0.195\textwidth}
		\centering
		\includegraphics[width=1\textwidth]{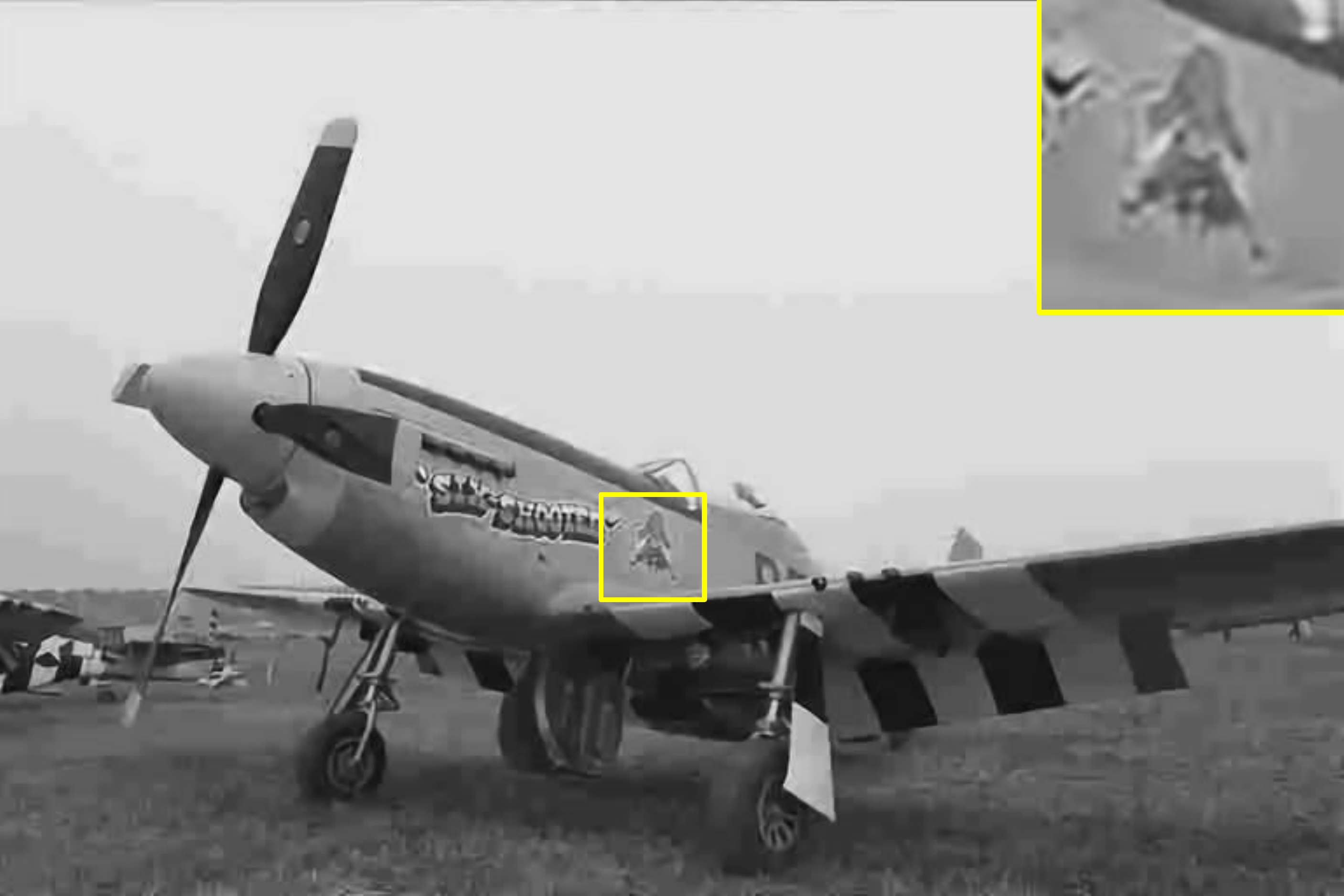}
		\subcaption*{(a) AR-CNN}
	\end{subfigure}
	\begin{subfigure}[t]{0.195\textwidth}
		\centering
		\includegraphics[width=1\textwidth]{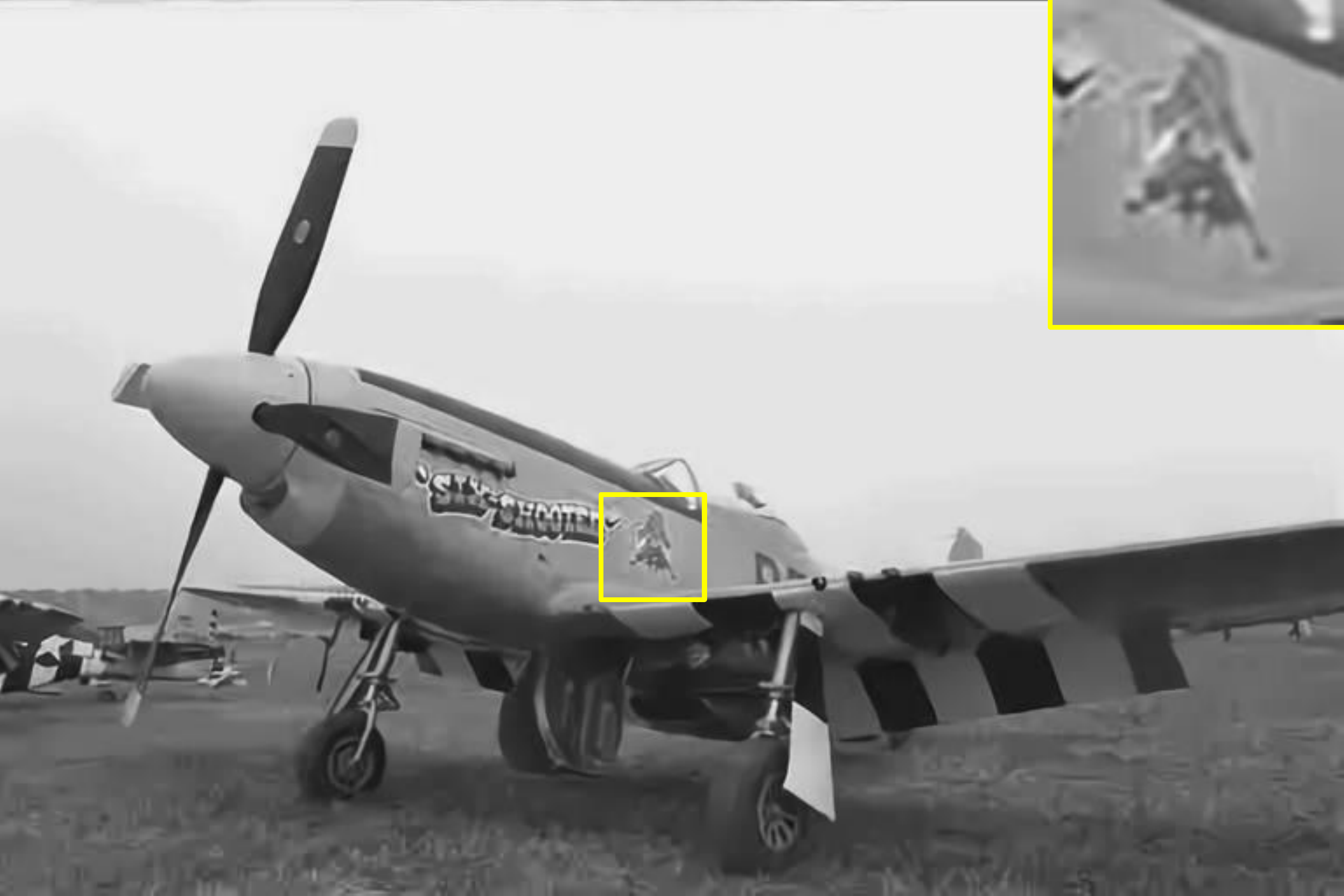}
		\subcaption*{(b) DnCNN}
	\end{subfigure}
	\begin{subfigure}[t]{0.195\textwidth}
		\centering
		\includegraphics[width=1\textwidth]{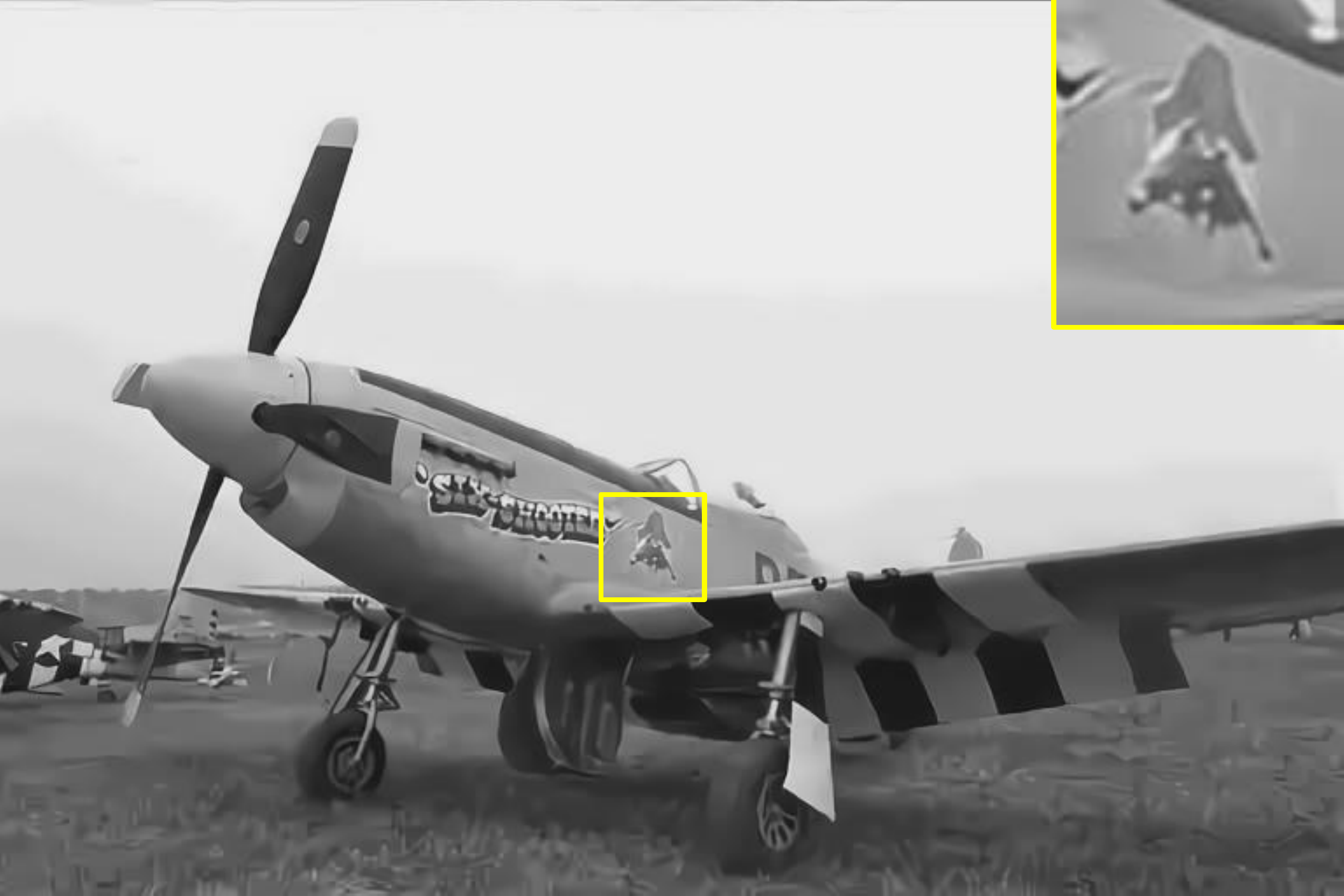}
		\subcaption*{(c) DURR}
	\end{subfigure}
	\caption{JPEG deblocking results of an image from the LIVE1 dataset,
		compressed using QF 10.}
	\label{jpegres2}
\end{figure}

Quantitative results are shown in Table \ref{jpegtab}.
Though the number of parameters of DURR is significantly less than
the DnCNN-3, the proposed DURR outperforms DnCNN-3 in most cases.
Specifically, considerable gains can be observed for our model on seen QFs, and
the performances are comparable on unseen QFs. A representative
result on the LIVE1 dataset is presented in Fig. \ref{jpegres2}.
Our model generates the most clean and accurate details.
More experiment details are given in the appendix.

\begin{table}[htp!]
	\caption{The average PSNR(dB) on the LIVE1 dataset. Values with $^*$ means
		the corresponding QF is not present in the training data of the model.
		The best results are indicated in red and the second best results are indicated in blue.}
	\label{jpegtab}
	\centering
	\begin{tabular}{c|cccccccc}
		\toprule
		QF
		&JPEG
		&SA-DCT
		&AR-CNN
		&AR-CNN-B
		&DnCNN-3
		&DURR\\
		\midrule\midrule
		10   &27.77 &28.65 &28.98 &28.53 &{\color{red}29.40} &{\color{blue}29.23$^*$} \\
		20   &30.07 &30.81 &31.29 &30.88 &{\color{blue}31.59} &{\color{red} 31.68}     \\
		30   &31.41 &32.08 &32.69 &32.31 &{\color{blue}32.98} &{\color{red}33.05}     \\
		40   &32.45 &32.99 &33.63 &33.39 &{\color{blue}33.96} &{\color{red}34.01$^*$} \\
		\bottomrule
	\end{tabular}
	
\end{table}

\begin{figure}[htp!]
	\centering
	\begin{subfigure}[t]{0.19\textwidth}
		\centering
		\includegraphics[width=1\textwidth]{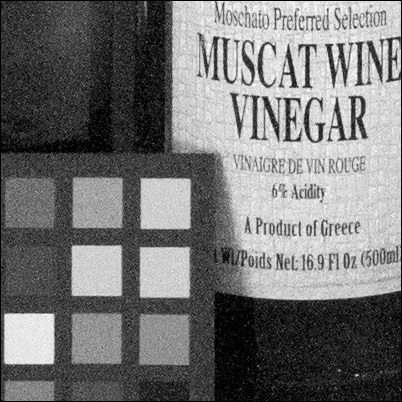}
		\subcaption*{Noisy Image}
	\end{subfigure}
	\begin{subfigure}[t]{0.19\textwidth}
		\centering
		\includegraphics[width=1\textwidth]{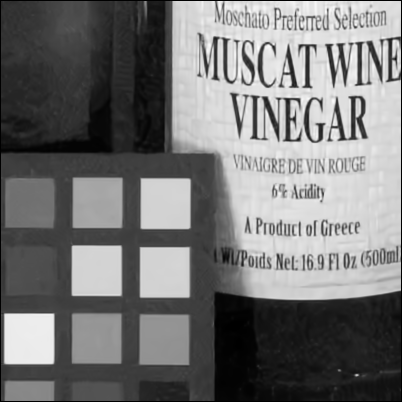}
		\subcaption*{BM3D}
	\end{subfigure}
	\begin{subfigure}[t]{0.19\textwidth}
		\centering
		\includegraphics[width=1\textwidth]{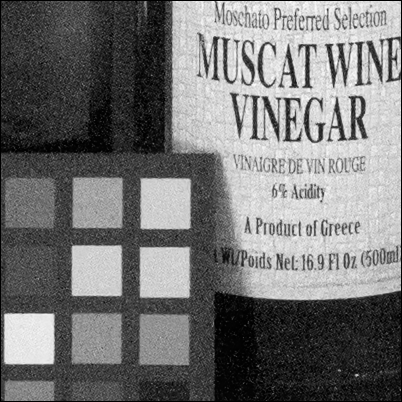}
		\subcaption*{DnCNN}
	\end{subfigure}
	\begin{subfigure}[t]{0.19\textwidth}
		\centering
		\includegraphics[width=1\textwidth]{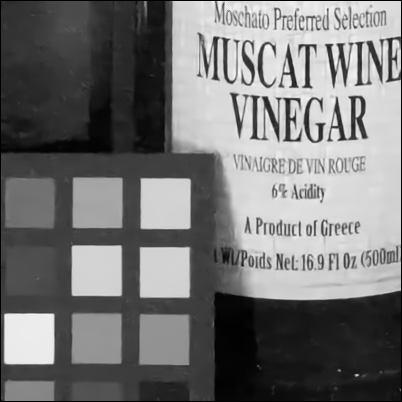}
		\subcaption*{UNet$_5$}
	\end{subfigure}
	\begin{subfigure}[t]{0.19\textwidth}
		\centering
		\includegraphics[width=1\textwidth]{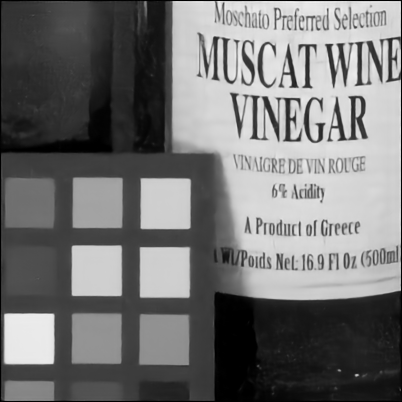}
		\subcaption*{DURR}
	\end{subfigure}
	\caption{Denoising results on a real image from \cite{lebrun2015noise}.}
	\label{real}
\end{figure}

\subsection{Other Applications}

Our model can be easily extended to other applications such as
deraining, dehazing and deblurring. In all these applications,
there are images corrupted at different levels. Rainfall intensity,
haze density and different blur kernels will all effect
the image quality.

\section{Conclusions}

In this paper, we proposed a novel image restoration model based on
the moving endpoint control in order to handle varied noise levels using a single model. The problem was solved by
jointly optimizing two units: restoration unit and policy unit.
The restoration unit used an RNN to realize the dynamics
in the control problem. A policy unit was proposed for
the policy unit to determine the loop times of the restoration unit
for optimal results. Our model achieved the state-of-the-art
results in blind image denoising and JPEG deblocking. Moreover,
thanks to the flexibility of the given policy, DURR has shown
strong abilities of generalization in our experiments.

%
%

\bibliographystyle{iclr2019_conference}
\small
\bibliography{durr_ref}
\normalsize

\newpage

\section*{Appendix}

\subsection{Network Structure}

\subsubsection{The Adopted Structure}

For the restoration unit, we use a minimal
U-Net \citep{ronneberger2015u} style network to predict the residual image.
The input of the restoration unit is the processed image (\textit{i.e.} the
last output) and the original degraded observation. The architecture of
the network is listed in Table. \ref{arch}.
The architecture of the policy unit is listed in Table. \ref{archpoly}.

\begin{table}[htp!]
	\caption{Architecture of the restoration unit. After each convolution layer,
		except the last one, there is a Parametric Rectified Linear Unit (PReLU) layer.
		The output of the second conv. layer is concatenated as a part of the input of
		the second-to-last conv. layer.}
	\label{arch}
	\centering
	\begin{tabular}{ccccc}
		\toprule
		Type & Kernel & Dilation & Stride & Outputs\\
		\hline\hline
		conv.& 5$\times$5& 1 & 1$\times$1 &32\\
		conv.& 3$\times$3& 1 & 1$\times$1 &32\\
		\hline
		conv.& 3$\times$3& 1 & 2$\times$2 &64\\
		
		conv.& 3$\times$3&1&1$\times$1&64\\
		dilated conv.&3$\times$3&2&1$\times$1&64\\
		dilated conv.&3$\times$3&4&1$\times$1&64\\
		\hline
		deconv.&$4\times 4$&1&$1/2 \times1/2$&64\\
		conv.& 3$\times$3&1&1$\times$1&32\\
		conv.& 5$\times$5&1&1$\times$1&1\\
		\hline		
	\end{tabular}
\end{table}

\begin{table}[htp!]
	\caption{Architecture of the policy unit. After each convolution layer, there is a Rectified Linear Unit (ReLU) layer.}
	\label{archpoly}
	
	\centering
	\begin{tabular}{cccccc}
		\toprule
		Type & Kernel & Dilation & Stride & Outputs & Remark\\
		\hline\hline
		conv.& 5$\times$5& 1 & 1$\times$1 &16\\
		conv.& 3$\times$3& 1 & 1$\times$1 &16& Link 1\\
		conv.& 3$\times$3& 1 & 1$\times$1 &16\\
		conv.& 3$\times$3& 1 & 1$\times$1 &16& Add Link 1\\
		\hline
		conv.& 3$\times$3& 1 & 2$\times$2 &32&Link 2\\
		
		conv.& 3$\times$3& 1 & 1$\times$1 &32\\
		conv.& 3$\times$3& 1 & 1$\times$1 &32&Add Link 2\\
		\hline
		conv.& 3$\times$3& 1 & 2$\times$2 &64&Link 3\\
		
		conv.& 3$\times$3& 1 & 1$\times$1 &64\\
		conv.& 3$\times$3& 1 & 1$\times$1 &64&Add Link 3\\
		\hline
		\multicolumn{6}{c}{Global Average Pooling}\\
		\hline	
		\multicolumn{6}{c}{LSTM with 32 hidden units}\\
		\hline	
		fc.  & - &- &- &1&\\
		\hline
	\end{tabular}
\end{table}

\subsubsection{Discussions}

\begin{figure}[htp!]
	\centering
	
	\begin{subfigure}[t]{0.22\textwidth}
		\centering
		\includegraphics[width=1\textwidth]{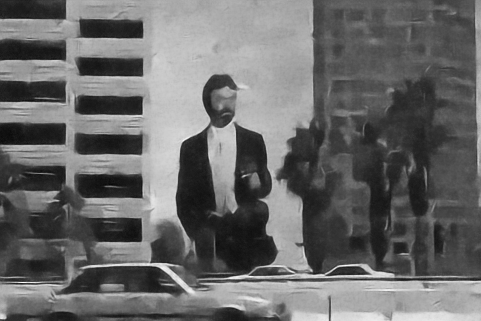}
	\end{subfigure}
	\begin{subfigure}[t]{0.22\textwidth}
		\centering
		\includegraphics[width=1\textwidth]{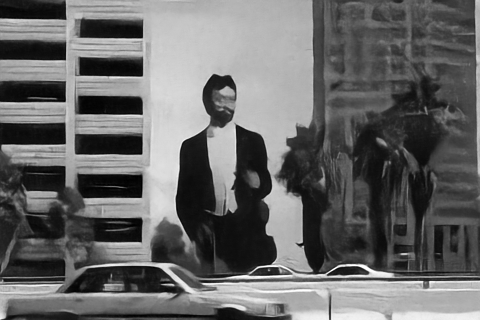}
	\end{subfigure}
	\hspace{2pt}
	\begin{subfigure}[t]{0.22\textwidth}
		\centering
		\includegraphics[width=1\textwidth]{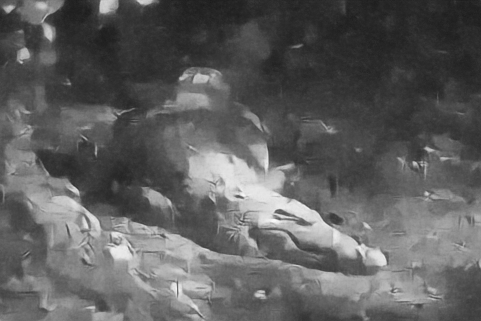}
	\end{subfigure}
	\begin{subfigure}[t]{0.22\textwidth}
		\centering
		\includegraphics[width=1\textwidth]{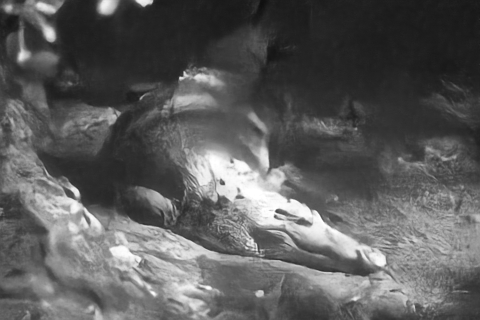}
	\end{subfigure}\\
	\begin{subfigure}[t]{0.22\textwidth}
		\centering
		\includegraphics[width=1\textwidth]{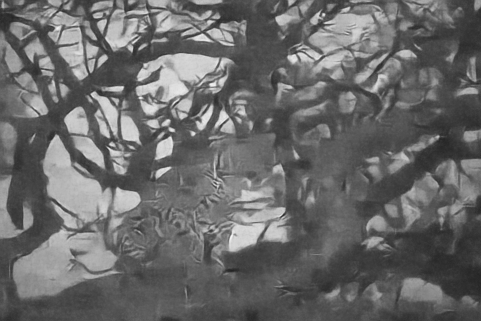}
	\end{subfigure}
	\begin{subfigure}[t]{0.22\textwidth}
		\centering
		\includegraphics[width=1\textwidth]{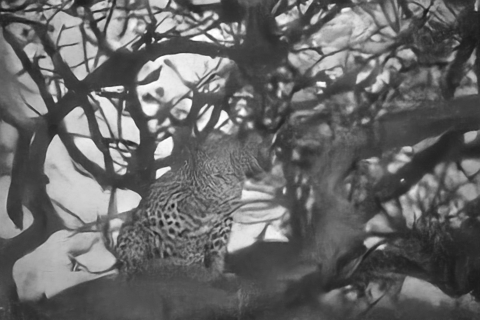}
	\end{subfigure}
	\hspace{2pt}
	\begin{subfigure}[t]{0.22\textwidth}
		\centering
		\includegraphics[width=1\textwidth]{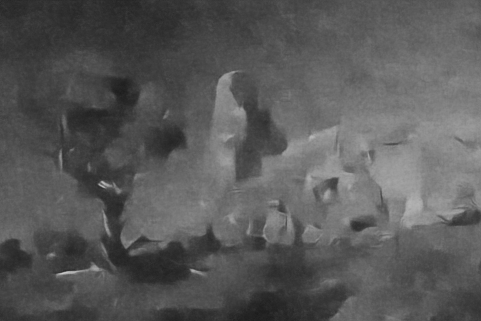}
	\end{subfigure}
	\begin{subfigure}[t]{0.22\textwidth}
		\centering
		\includegraphics[width=1\textwidth]{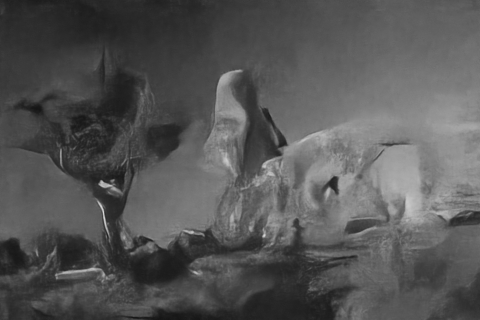}
	\end{subfigure}\\
	
	\caption{Denoising results of different restoration unit structure designs.
		Left images are produced by the AR-CNN-like unit. Right images are produced
		by our proposed minimal U-Net style network.}
	\label{str}
\end{figure}

Fig. \ref{str} justify the rationality of our restoration unit design.
An AR-CNN-like structure is tested in our experiments, the
number of parameters is comparable with the restoration unit adopted in DURR.

As the Fig. \ref{str} illustrates, the U-Net style network
(the one we adopted) generated images tend to have significantly
less artifacts as well as more pleasing qualities, though the
PSNR results of these images are close.

\subsection{Analysis on Generalization Power and Efficiencies}
In this part, we try to analysis the generalization power and
time efficiency of our models. We carry out a new experiment
on denoising under fair settings, where all models
are trained using BSD400 images with $\sigma \in [25, 45]$.
Furthermore, our inference time can be greatly reduced
while performance being kept, if we meticulously shrink the width
of the two units, and modify the policy to apply the enhance unit for
two times on each restoration stage. This model is called
\textbf{D}oppio-\textbf{DURR} in the table.

Experiment results are in Tab.\ref{exptable} and Tab.\ref{exptable1}.
It can be seen that the performances of our models
surpass DnCNN-B on all noise levels and two metrics
when testing under this fair setting, especially on unseen noise levels.
These results further proves the generalization power of our models.

As for the inference time, our model D-DURR is the fastest among all
noise levels. The DURR is faster than the DnCNN-B on low noise levels.
Due to the dynamically unfolding process, the DURR could be slower
than the DnCNN-B when the noise level goes higher.

\begin{table}[htp!]
	\caption{Average PSNR / SSIM on BSD68.
	Red for the best. Blue for the second best.}
	\vspace{-0.08in}
	\label{exptable}
	\centering
	\begin{tabular}{lccccccc}
		\toprule
		 &$\sigma=15$ (unseen)
		 &$\sigma=25$
		 &$\sigma=35$
		 &$\sigma=45$
		 &$\sigma=55$ (unseen)\\
		\midrule
		\midrule
		DnCNN-B
		&30.55/0.849
		&29.16/0.824
		&27.69/0.770
		&26.66/0.742
		&22.84/0.506 \\
		DURR
		&\red{31.32}/\red{0.883}
		&\red{29.28}/\red{0.838}
		&\red{27.84}/\red{0.795}
		&\red{26.83}/\red{0.757}
		&\red{25.80}/\red{0.704} \\
		D-DURR
		&\blue{31.19}/\blue{0.878}
		&\blue{29.19}/\blue{0.829}
		&\blue{27.72}/\blue{0.786}
		&\blue{26.72}/\blue{0.749}
		&\blue{25.71}/\blue{0.700} \\
		\bottomrule
	\end{tabular}
\end{table}

\begin{table}[htp!]
	\caption{Average Inference Time (ms) on BSD68.
	Red for the best. Blue for the second best.}
	\vspace{-0.08in}
	\label{exptable1}
	\centering
	\begin{tabular}{lccccccc}
		\toprule
		 &$\sigma=15$ (unseen)
		 &$\sigma=25$
		 &$\sigma=35$
		 &$\sigma=45$
		 &$\sigma=55$ (unseen)\\
		\midrule
		\midrule
		DnCNN-B
		&4.71
		&4.71
		&\blue{4.71}
		&\blue{4.71}
		&\blue{4.71} \\
		DURR
		&\blue{2.66}
		&\blue{4.69}
		&6.75
		&9.78
		&13.09 \\
		D-DURR
		&\red{1.28}
		&\red{2.31}
		&\red{3.01}
		&\red{3.79}
		&\red{4.65} \\
		\bottomrule
	\end{tabular}
\end{table}

\subsection{Further Results}

\subsubsection{Image Denoising}

The performaces of DnCNN and DURR under extreme noise conditions ($\sigma=95$)
is tested. Though the noise level is unseen for both models, it can be easily
observed from Fig. \ref{appeal} that the proposed DURR outperforms DnCNN
on both quantitative measurements and visual qualities.

\begin{figure}[htp!]
	\centering
	\begin{subfigure}[t]{0.28\textwidth}
		\centering
		\includegraphics[width=1\textwidth]{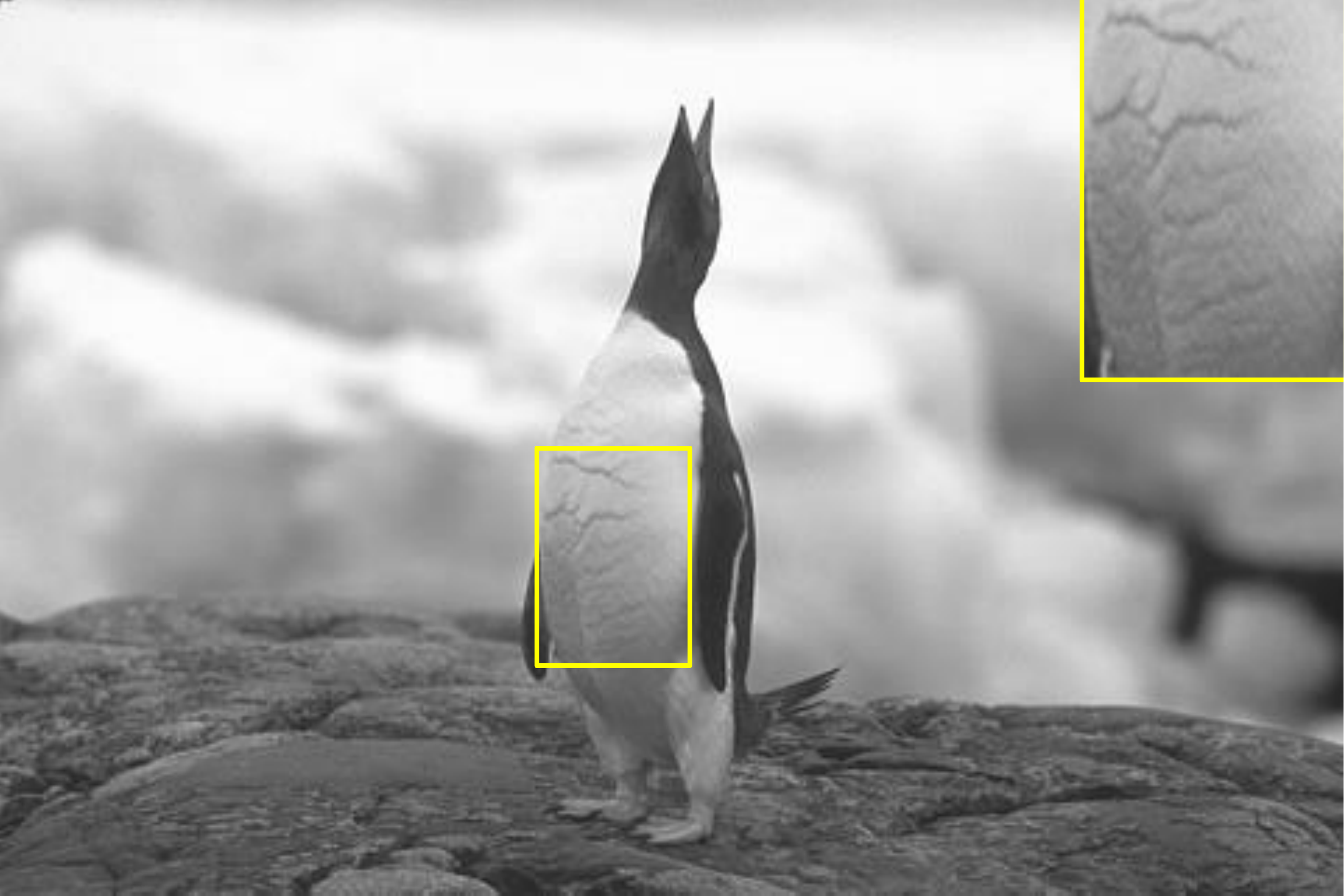}
		\subcaption*{Ground Truth}
	\end{subfigure}
	\quad
	\begin{subfigure}[t]{0.28\textwidth}
		\centering
		\includegraphics[width=1\textwidth]{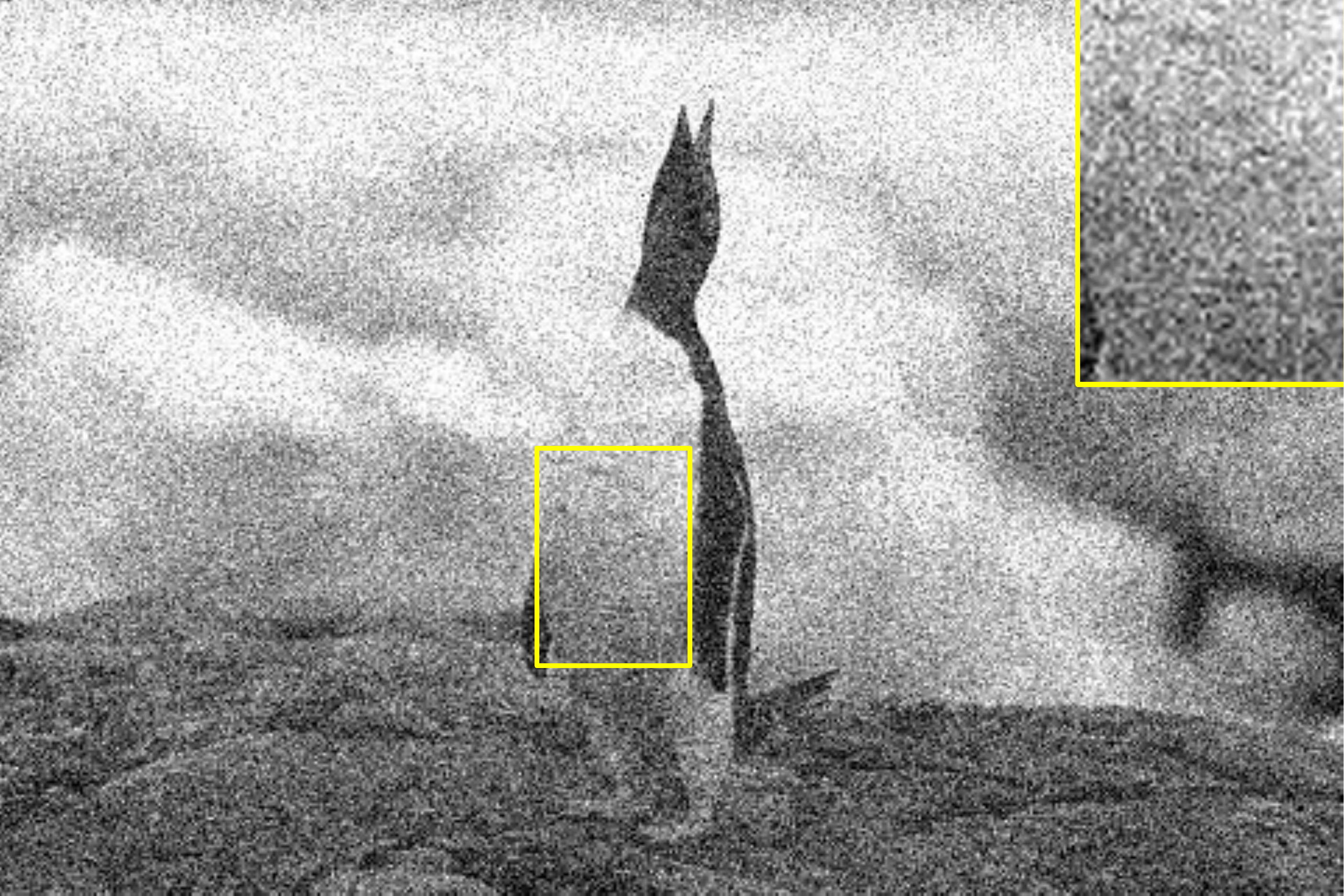}
		\subcaption*{Noisy Input, 17.94dB}
	\end{subfigure}
	\quad
	\begin{subfigure}[t]{0.28\textwidth}
		\centering
		\includegraphics[width=1\textwidth]{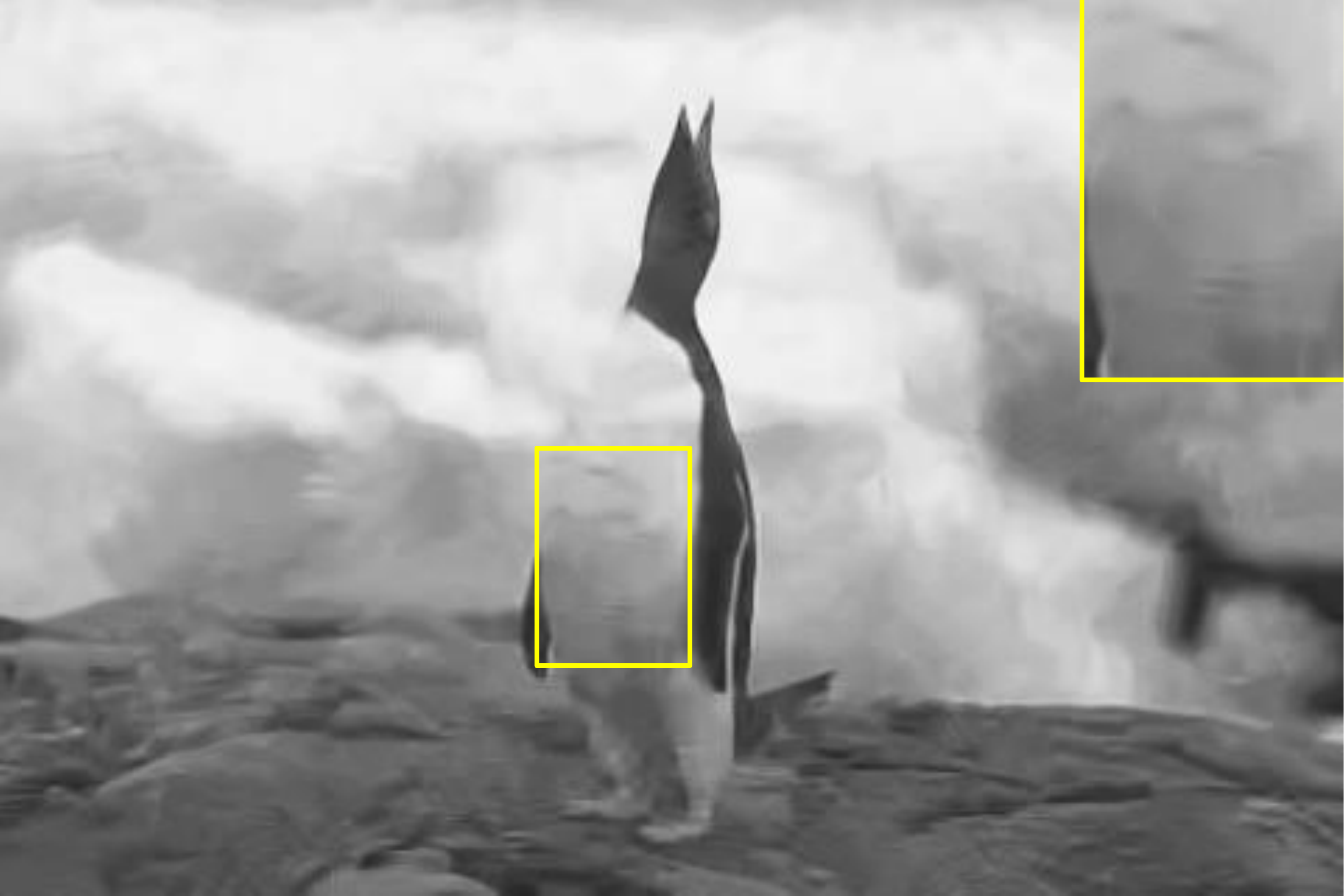}
		\subcaption*{(a) BM3D, 29.60dB}
	\end{subfigure}\\
	\begin{subfigure}[t]{0.28\textwidth}
		\centering
		\includegraphics[width=1\textwidth]{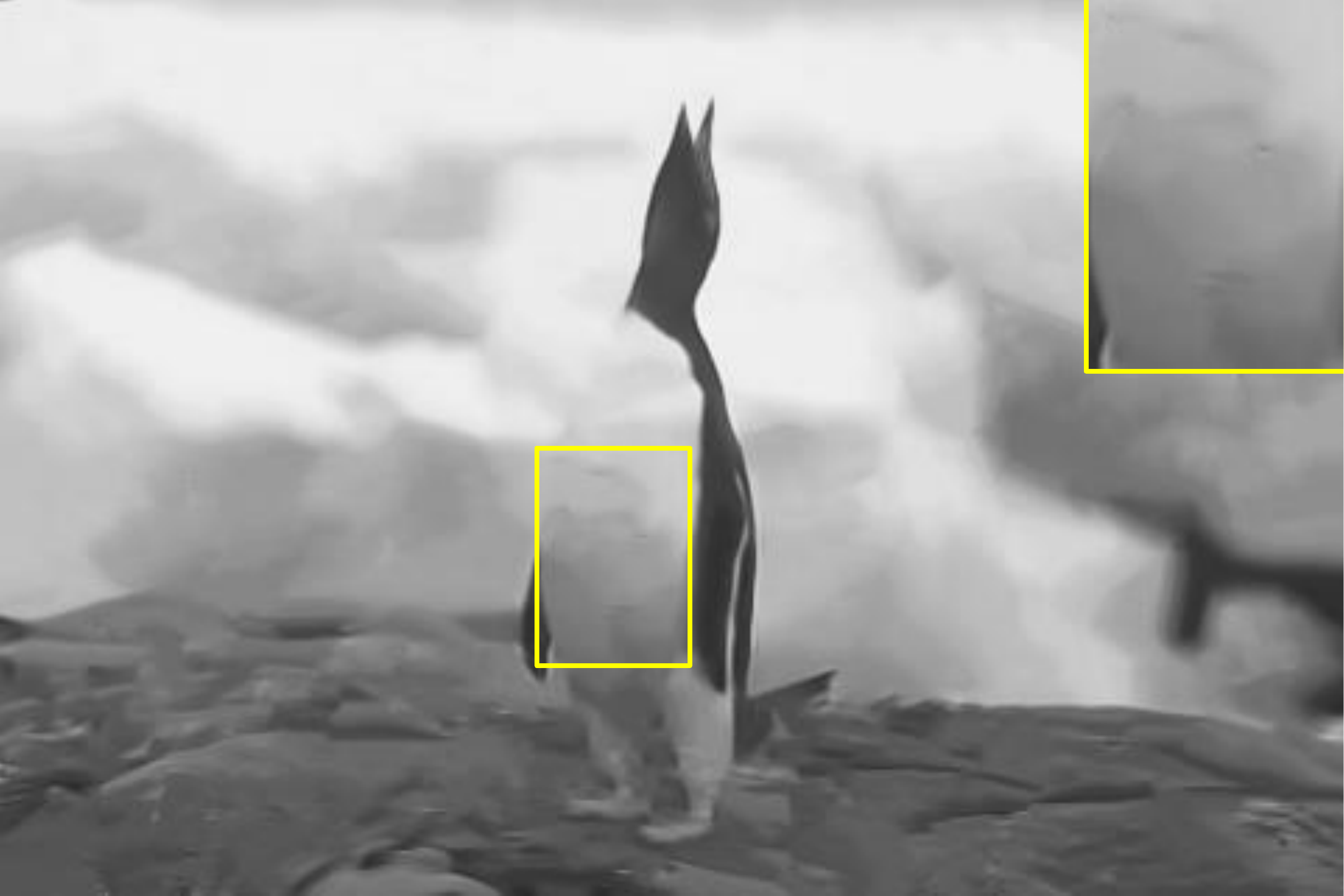}
		\subcaption*{(b) WNMM, 29.73dB}
	\end{subfigure}
	\quad
	\begin{subfigure}[t]{0.28\textwidth}
		\centering
		\includegraphics[width=1\textwidth]{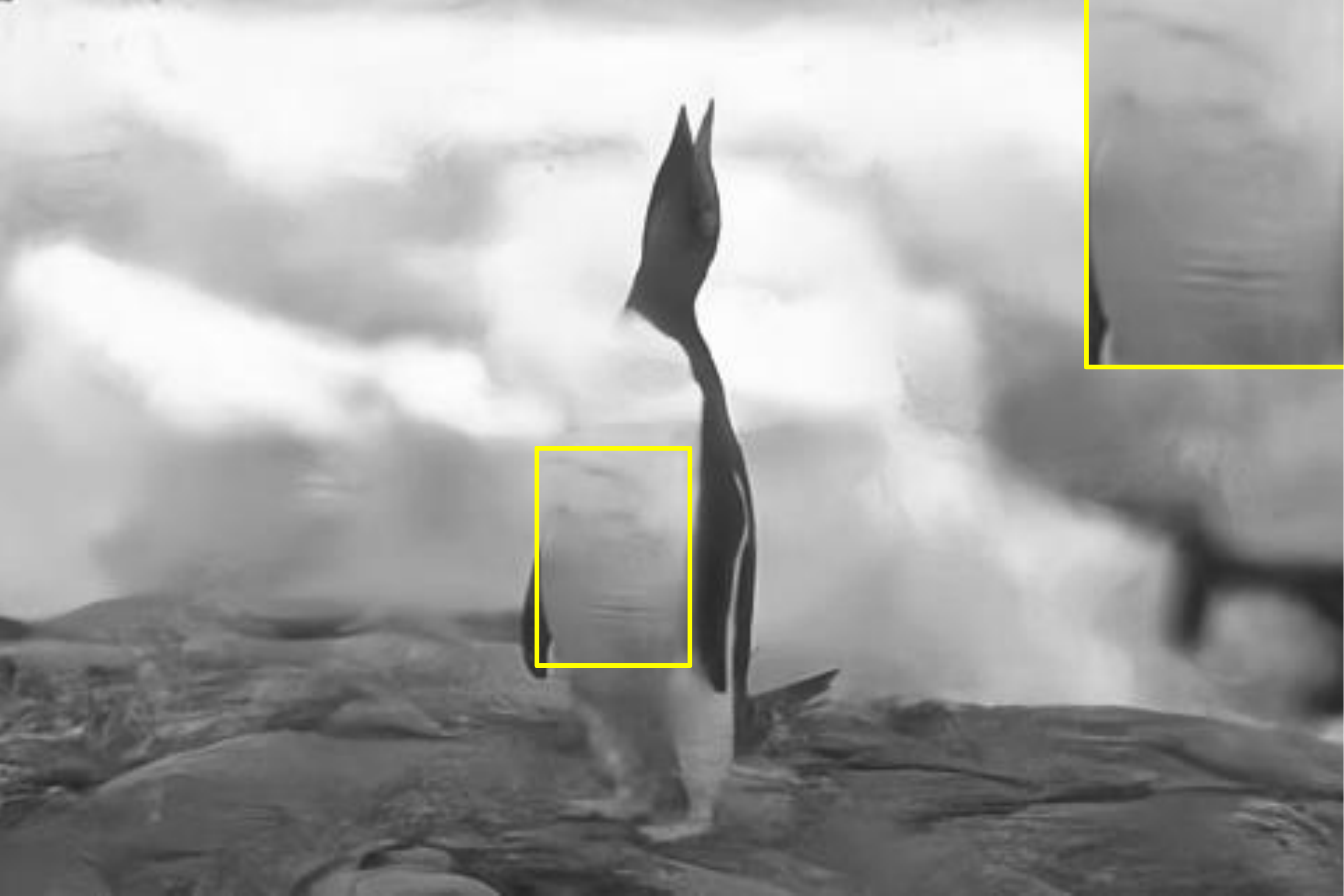}
		\subcaption*{(c) DnCNN, 31.67dB}
	\end{subfigure}
	\quad
	\begin{subfigure}[t]{0.28\textwidth}
		\centering
		\includegraphics[width=1\textwidth]{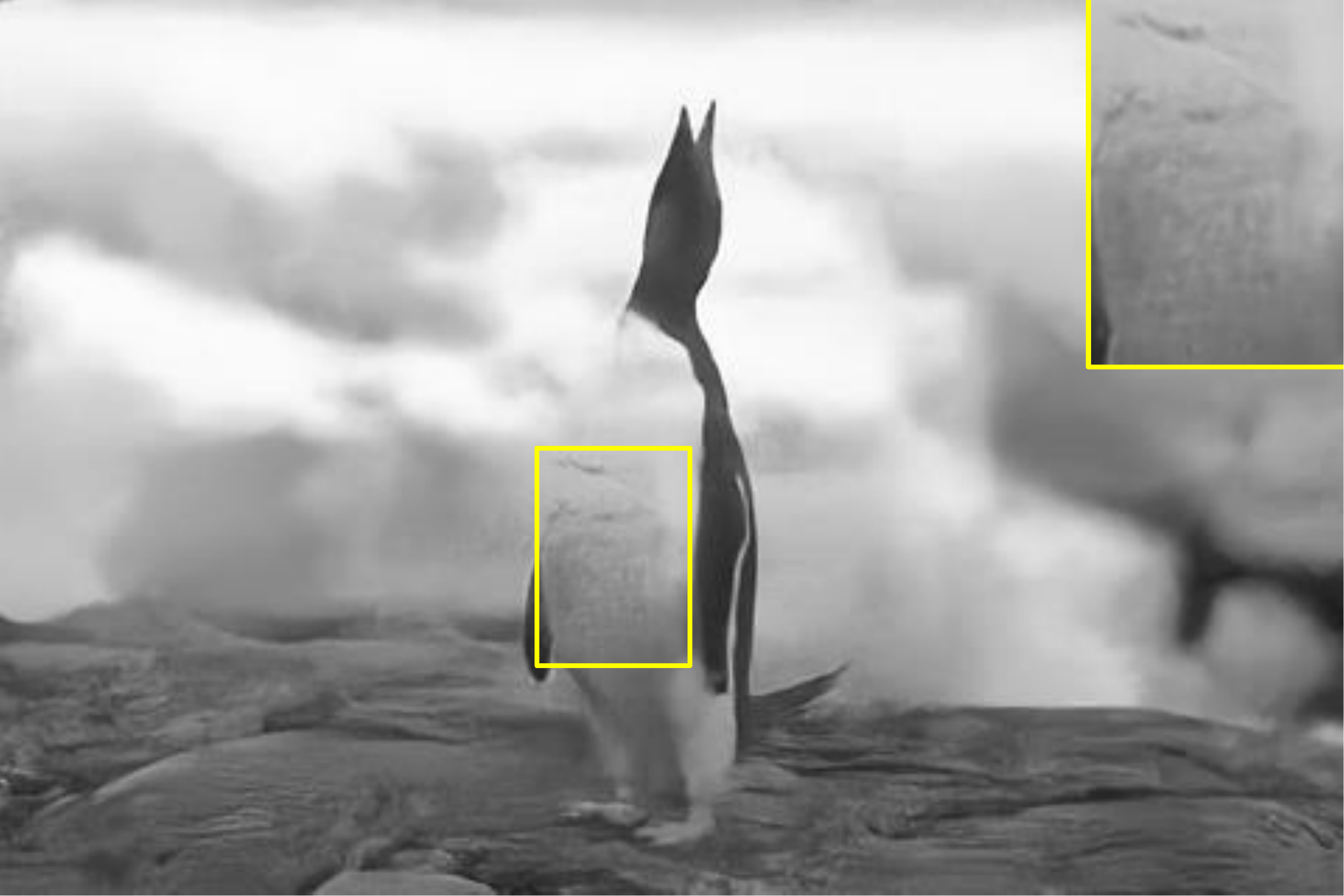}
		\subcaption*{(d) DURR, 31.72dB}
	\end{subfigure}
\caption{Denoising results of an image from BSD68 with
	noise level 35.}
\label{dn2}

\end{figure}

\begin{figure}[hp!]
	\centering
	\begin{subfigure}[t]{0.48\textwidth}
		\centering
		\includegraphics[width=1\textwidth]{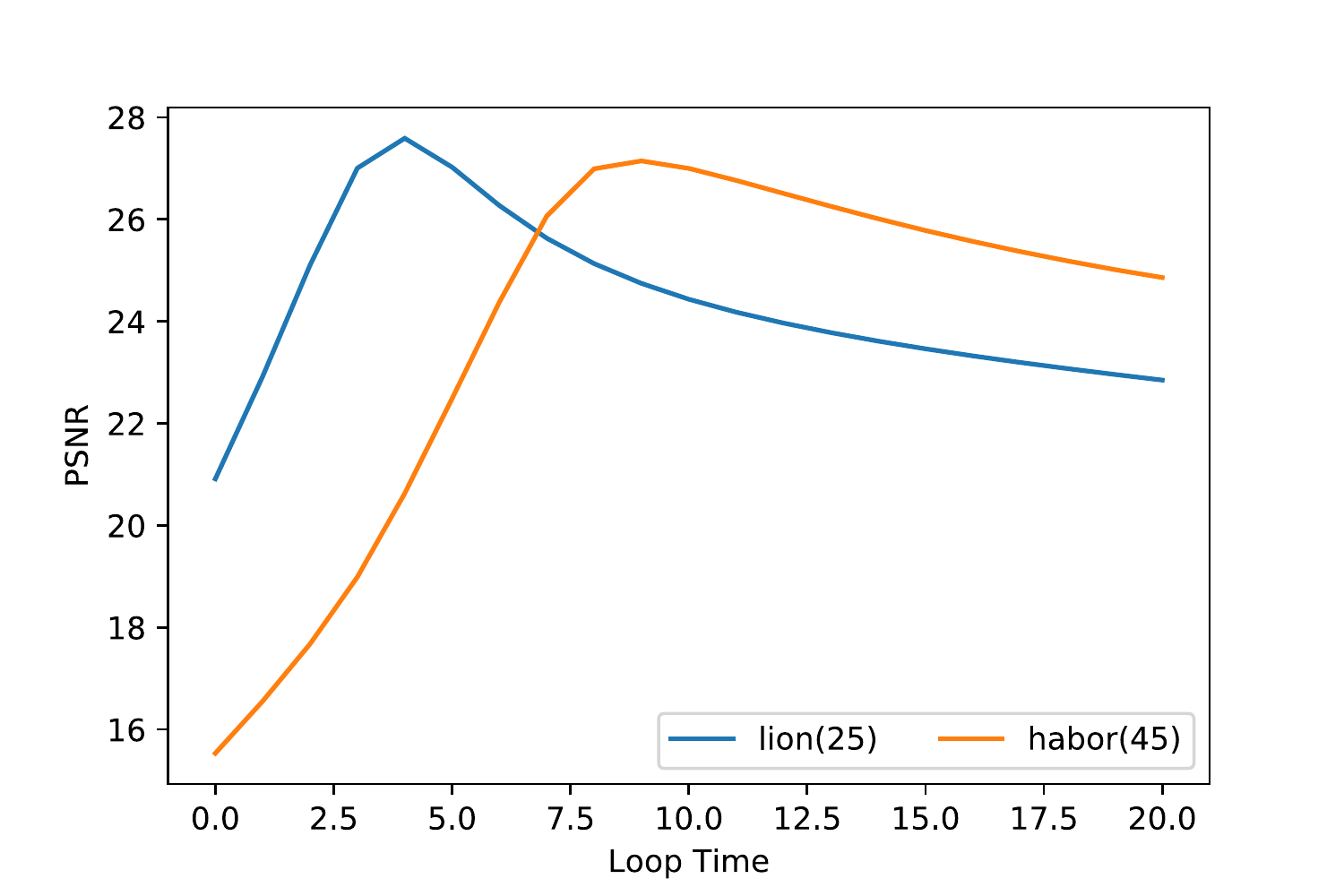}
		\subcaption*{PSNR tendency for lion and harbor images in Fig. \ref{loop}.}
	\end{subfigure}
	\quad
	\begin{subfigure}[t]{0.48\textwidth}
		\centering
		\includegraphics[width=1\textwidth]{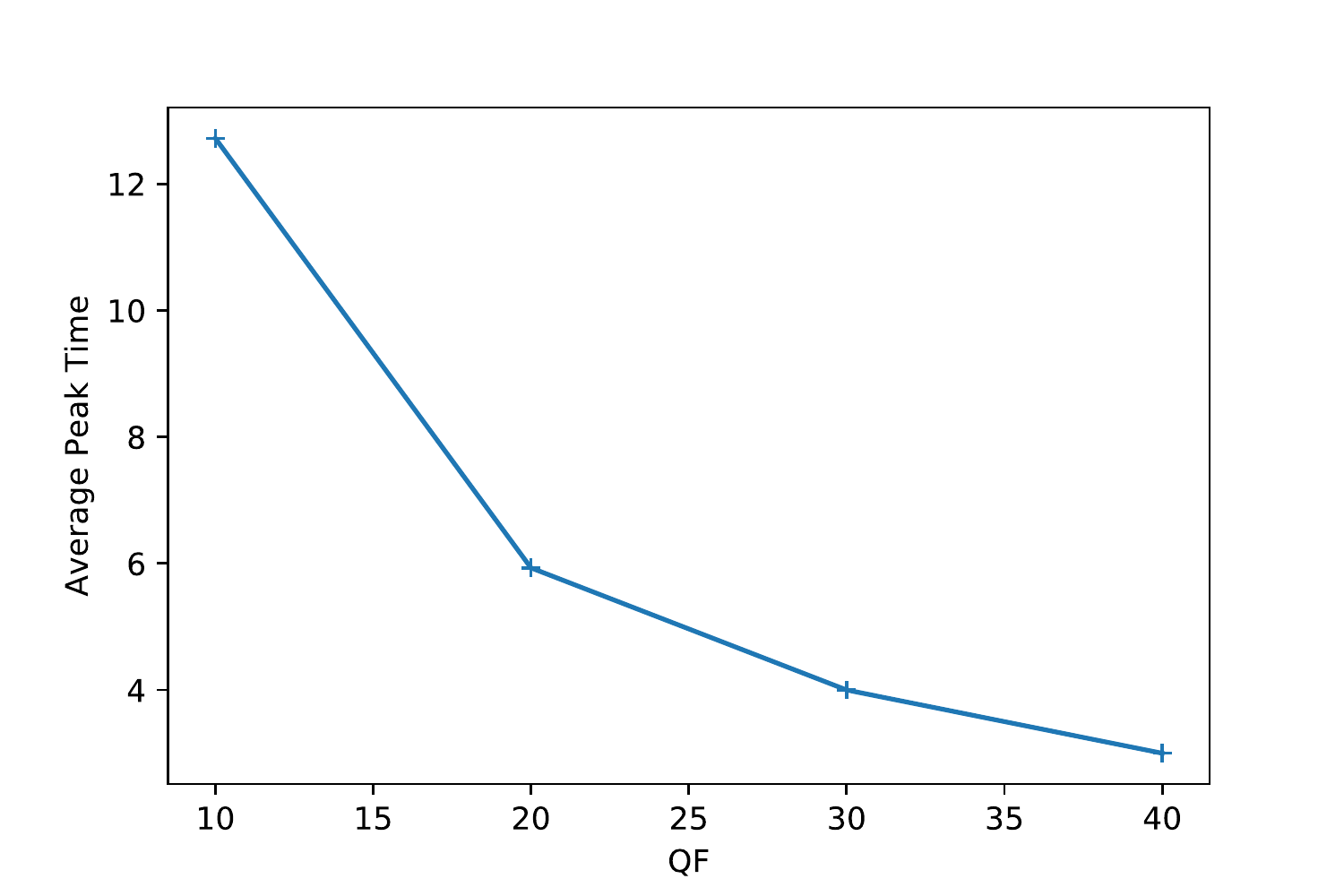}
		\subcaption*{Averge loop time relates to QF.}
	\end{subfigure}
	\caption{Image quality's relation to loop times.}
	\label{psnr}
\end{figure}

\begin{figure}[htp!]
	\centering
	\begin{subfigure}[t]{0.25\textwidth}
		\centering
		\includegraphics[width=1\textwidth]{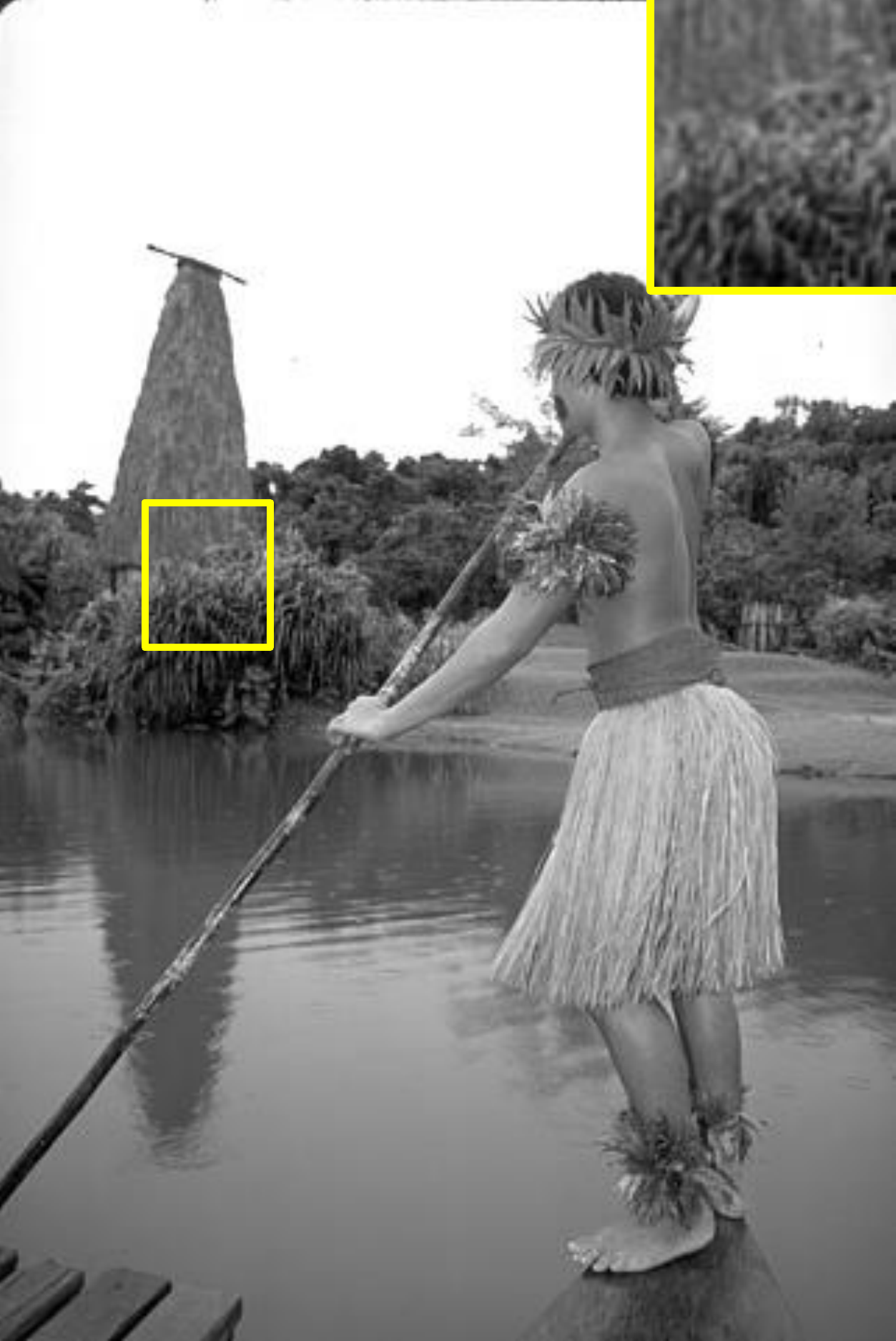}
		\subcaption*{Ground Truth}
	\end{subfigure}
	\quad
	\begin{subfigure}[t]{0.25\textwidth}
		\centering
		\includegraphics[width=1\textwidth]{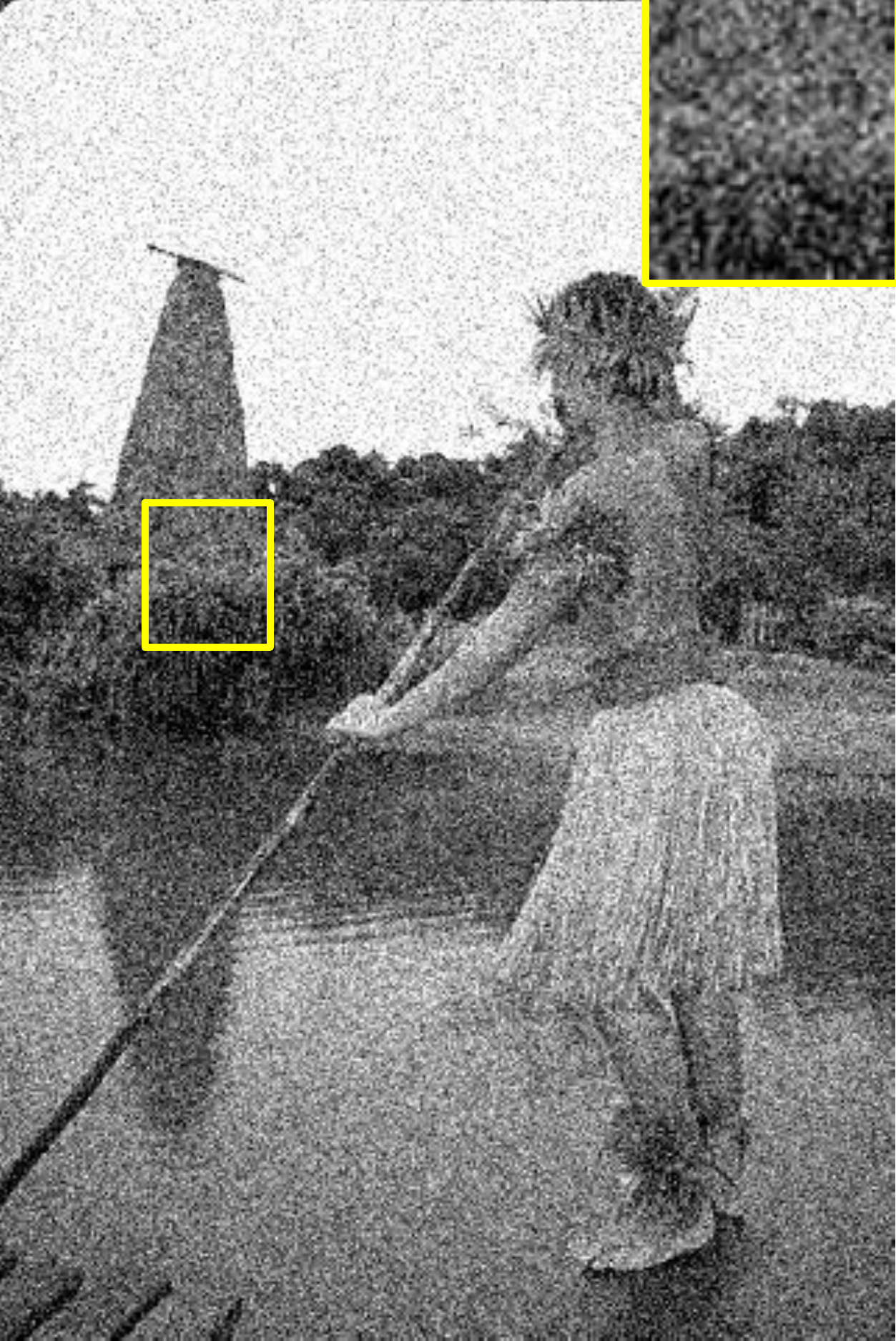}
		\subcaption*{Noisy Input, 18.05dB}
	\end{subfigure}
	\quad
	\begin{subfigure}[t]{0.25\textwidth}
		\centering
		\includegraphics[width=1\textwidth]{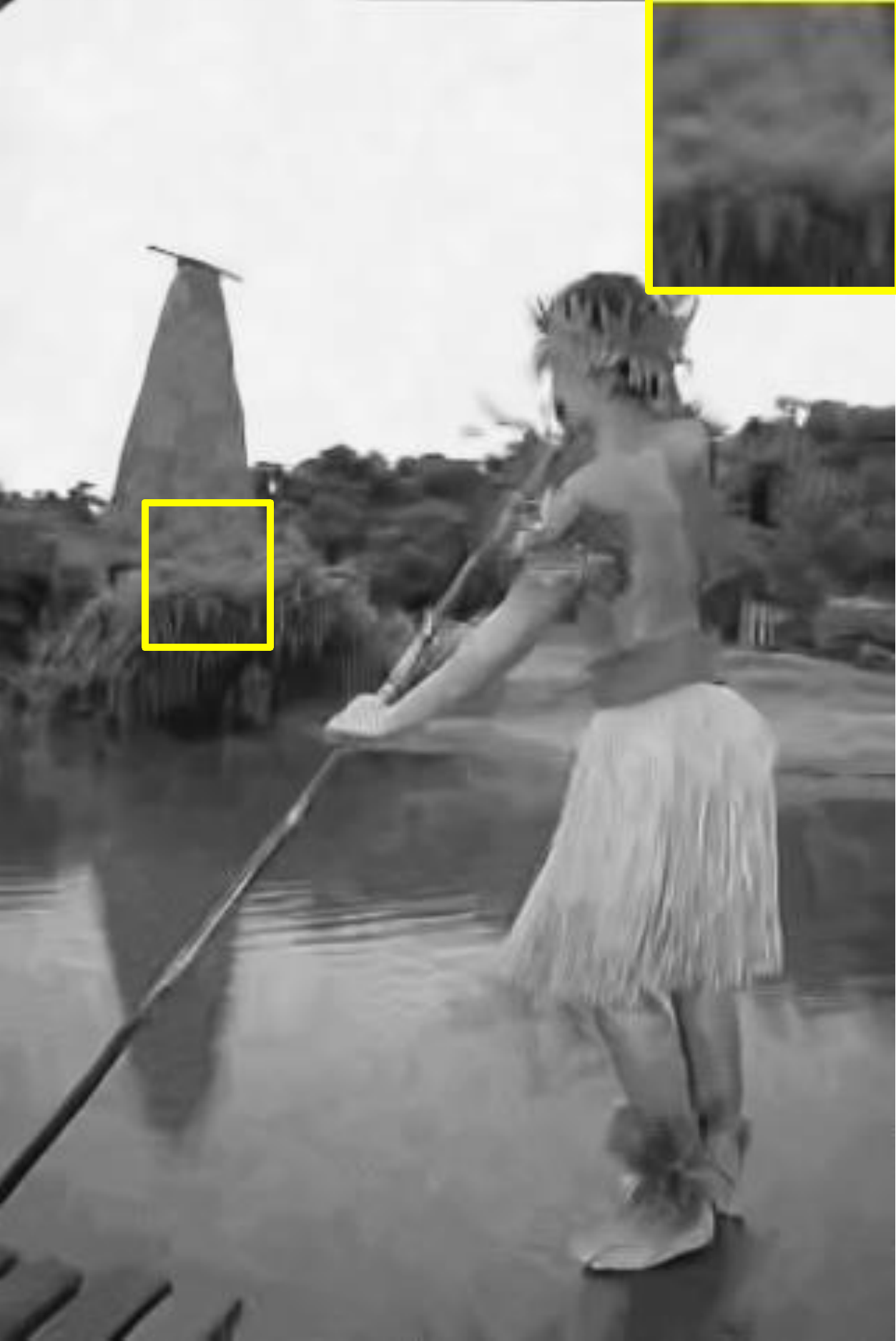}
		\subcaption*{BM3D, 26.22dB}
	\end{subfigure}\\
	\begin{subfigure}[t]{0.25\textwidth}
		\centering
		\includegraphics[width=1\textwidth]{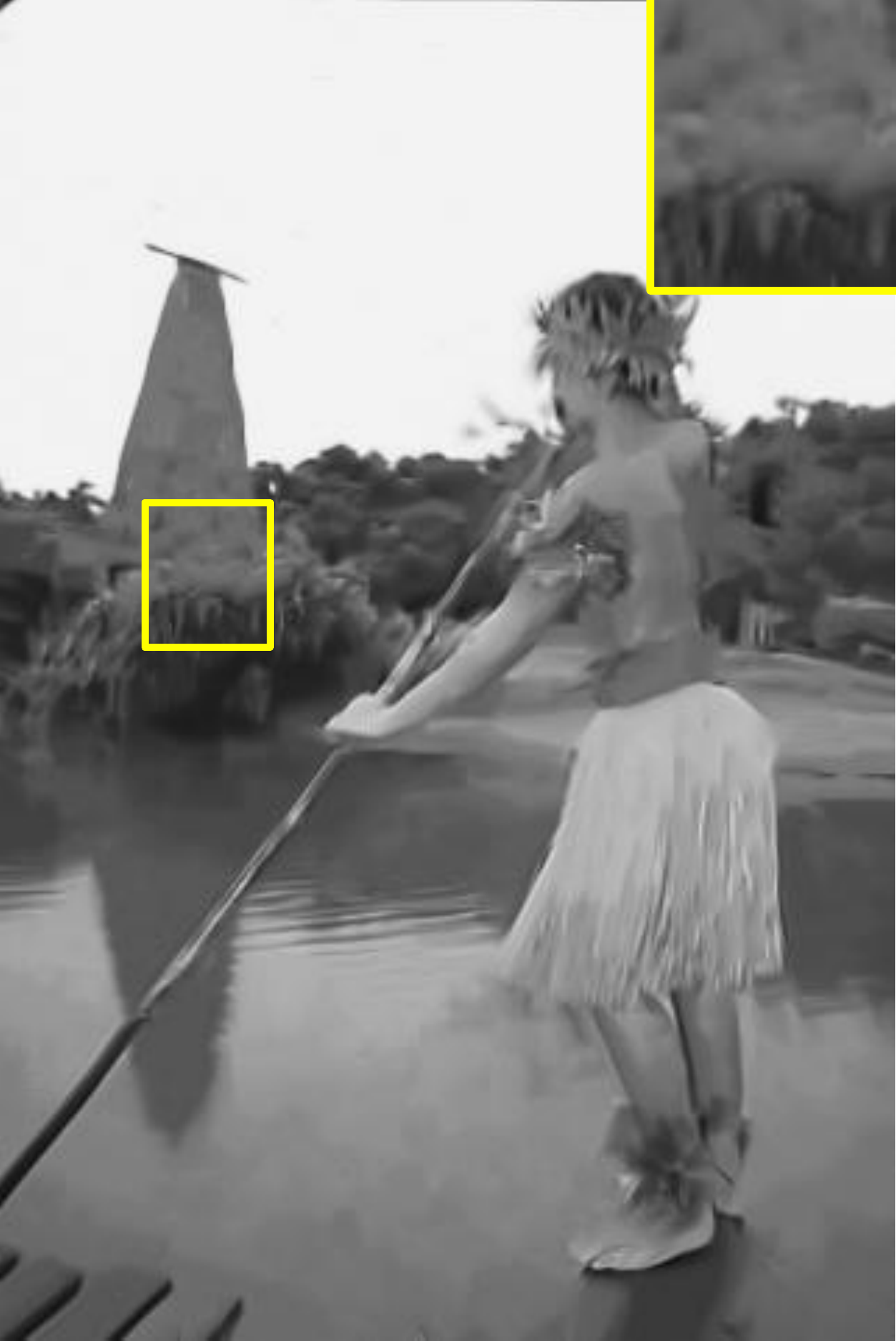}
		\subcaption*{WNMM, 26.54dB}
	\end{subfigure}
	\quad
	\begin{subfigure}[t]{0.25\textwidth}
		\centering
		\includegraphics[width=1\textwidth]{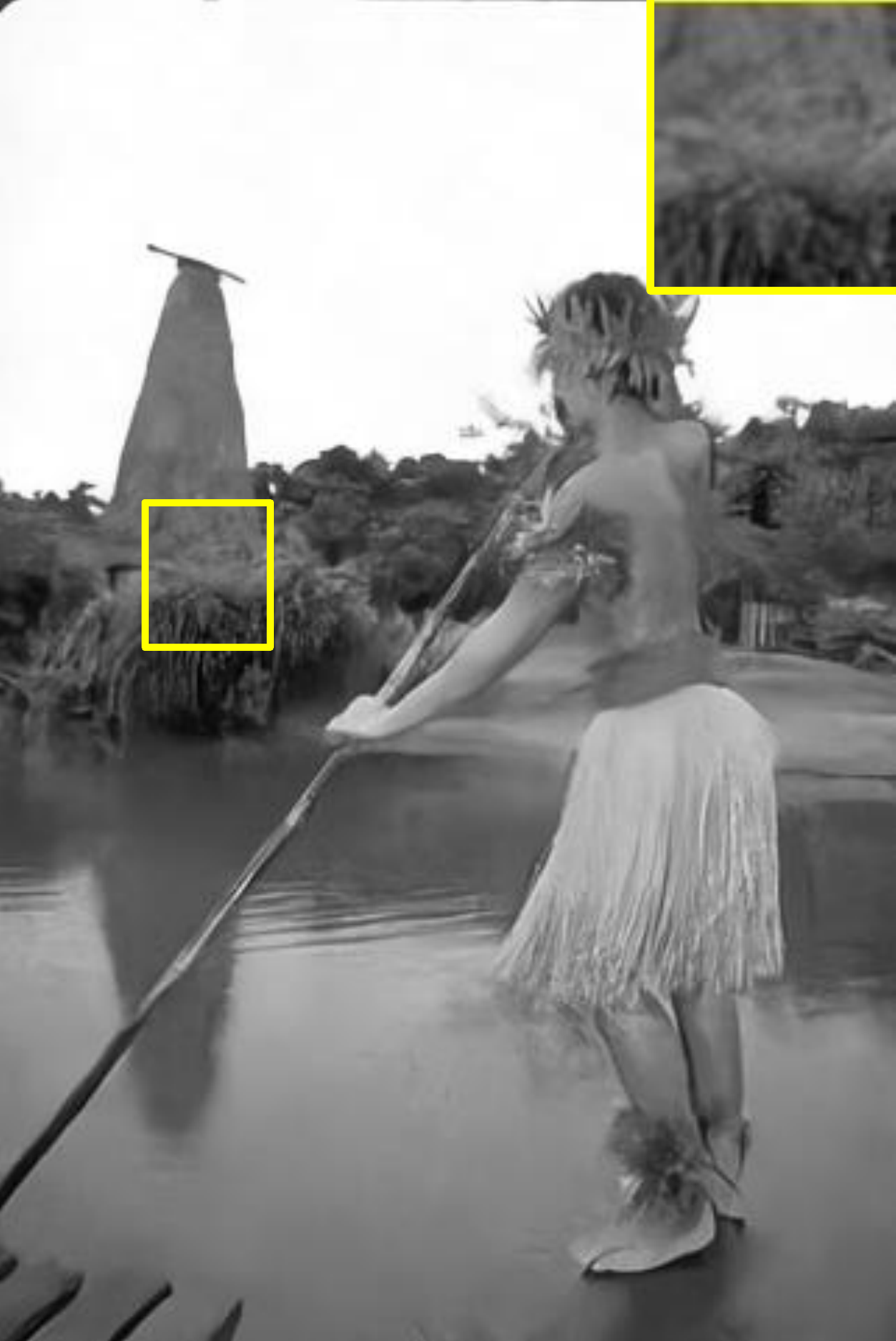}
		\subcaption*{DnCNN, 28.64dB}
	\end{subfigure}
	\quad
	\begin{subfigure}[t]{0.25\textwidth}
		\centering
		\includegraphics[width=1\textwidth]{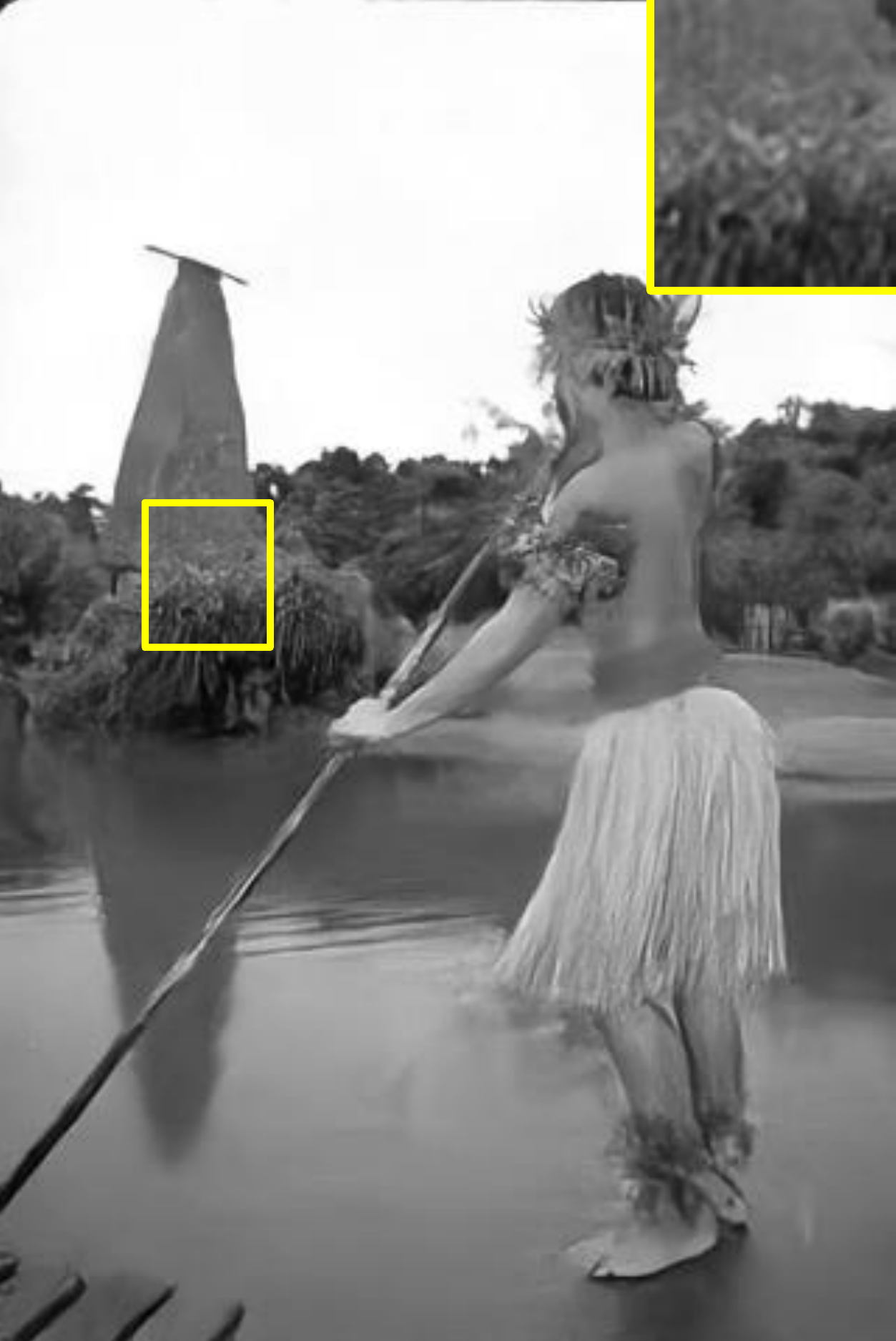}
		\subcaption*{DURR, 28.72dB}
	\end{subfigure}
	\caption{Denoising results of an image from BSD68 with
		noise level 35.}
	\label{dn1}
\end{figure}

\begin{figure}[htp!]
	\centering
	\begin{subfigure}[t]{0.22\textwidth}
		\centering
		\includegraphics[width=1\textwidth]{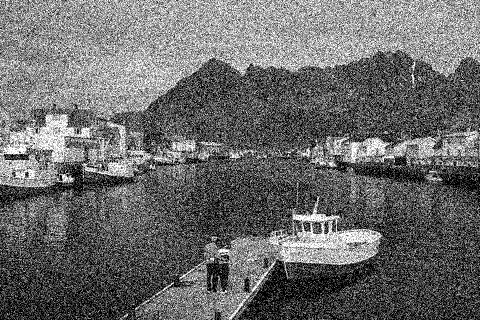}
	\end{subfigure}
	\begin{subfigure}[t]{0.22\textwidth}
		\centering
		\includegraphics[width=1\textwidth]{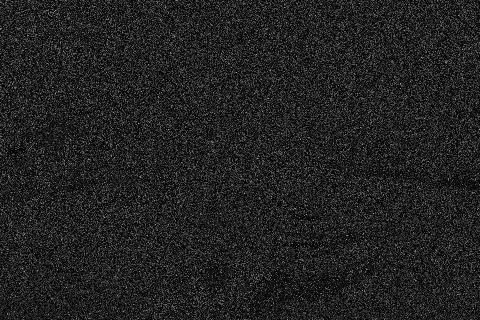}
	\end{subfigure}
	\hspace{2pt}
	\begin{subfigure}[t]{0.22\textwidth}
		\centering
		\includegraphics[width=1\textwidth]{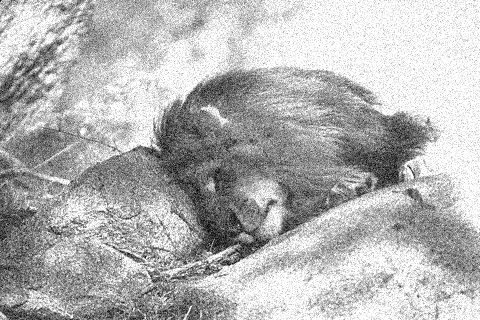}
	\end{subfigure}
	\begin{subfigure}[t]{0.22\textwidth}
		\centering
		\includegraphics[width=1\textwidth]{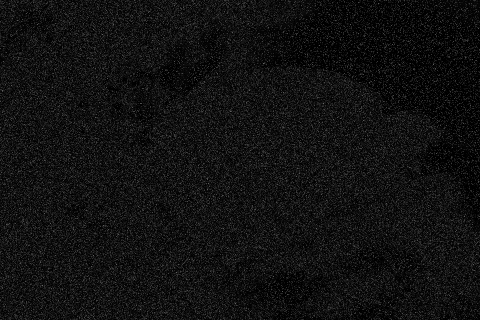}
	\end{subfigure}\\
	\begin{subfigure}[t]{0.22\textwidth}
		\centering
		\includegraphics[width=1\textwidth]{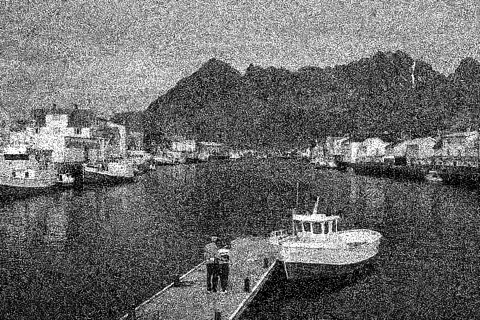}
	\end{subfigure}
	\begin{subfigure}[t]{0.22\textwidth}
		\centering
		\includegraphics[width=1\textwidth]{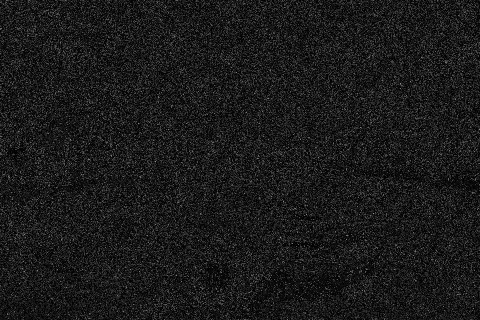}
	\end{subfigure}
	\hspace{2pt}
	\begin{subfigure}[t]{0.22\textwidth}
		\centering
		\includegraphics[width=1\textwidth]{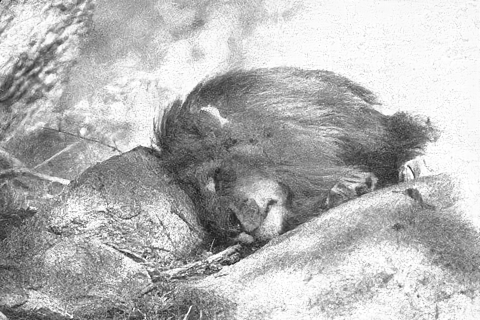}
	\end{subfigure}
	\begin{subfigure}[t]{0.22\textwidth}
		\centering
		\includegraphics[width=1\textwidth]{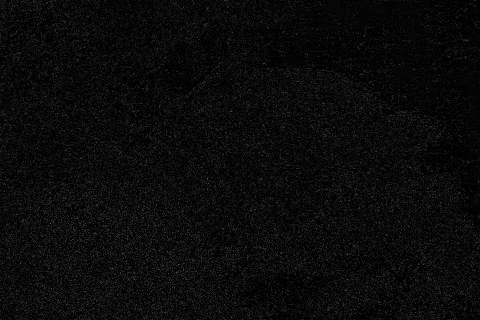}
	\end{subfigure}\\
	\begin{subfigure}[t]{0.22\textwidth}
		\centering
		\includegraphics[width=1\textwidth]{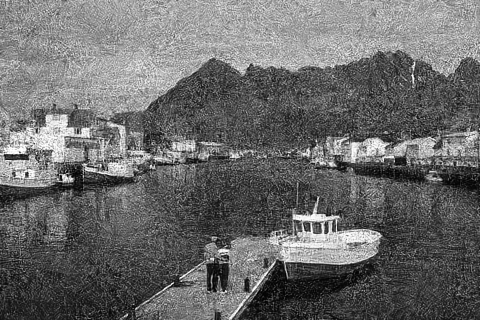}
	\end{subfigure}
	\begin{subfigure}[t]{0.22\textwidth}
		\centering
		\includegraphics[width=1\textwidth]{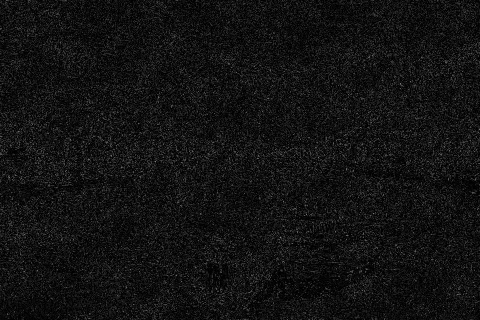}
	\end{subfigure}
	\hspace{2pt}
	\begin{subfigure}[t]{0.22\textwidth}
		\centering
		\includegraphics[width=1\textwidth]{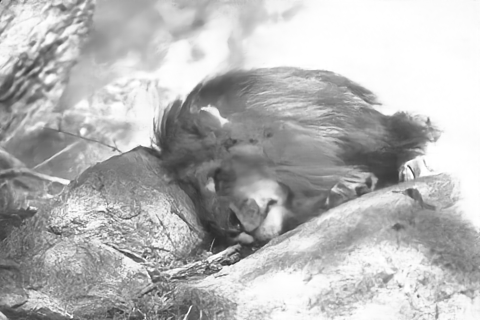}
	\end{subfigure}
	\begin{subfigure}[t]{0.22\textwidth}
		\centering
		\includegraphics[width=1\textwidth]{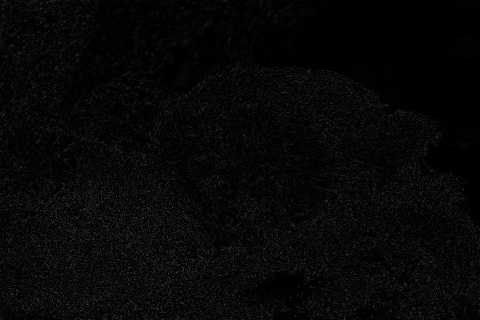}
	\end{subfigure}\\
	\begin{subfigure}[t]{0.22\textwidth}
		\centering
		\includegraphics[width=1\textwidth]{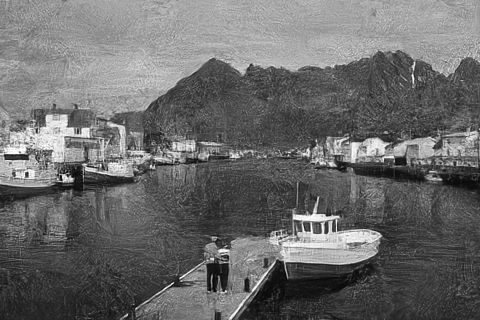}
	\end{subfigure}
	\begin{subfigure}[t]{0.22\textwidth}
		\centering
		\includegraphics[width=1\textwidth]{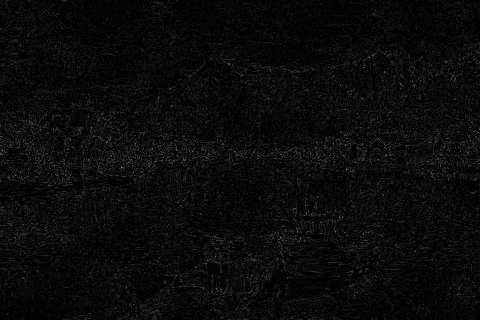}
	\end{subfigure}
	\hspace{2pt}
	\begin{subfigure}[t]{0.22\textwidth}
		\centering
		\includegraphics[width=1\textwidth]{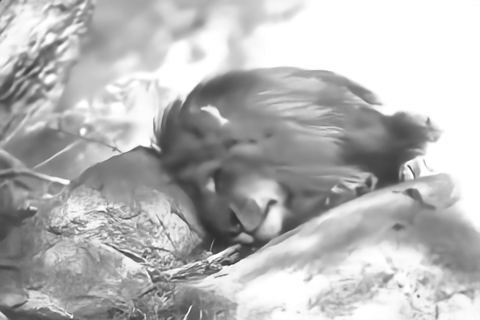}
	\end{subfigure}
	\begin{subfigure}[t]{0.22\textwidth}
		\centering
		\includegraphics[width=1\textwidth]{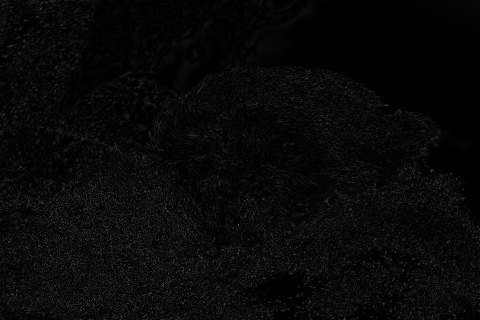}
	\end{subfigure}\\
	\begin{subfigure}[t]{0.22\textwidth}
		\centering
		\includegraphics[width=1\textwidth]{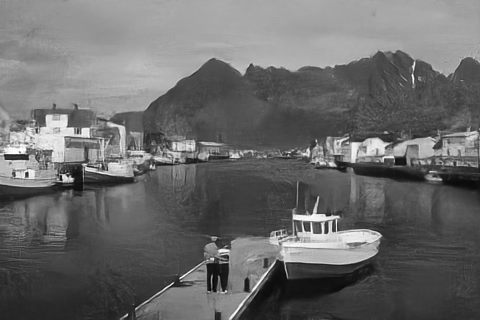}
	\end{subfigure}
	\begin{subfigure}[t]{0.22\textwidth}
		\centering
		\includegraphics[width=1\textwidth]{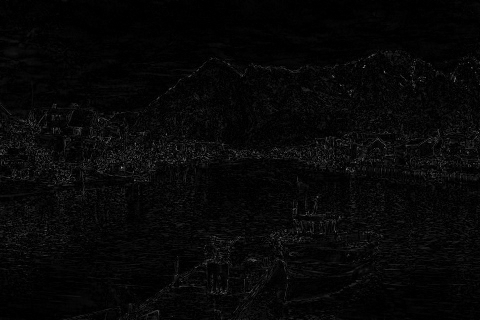}
	\end{subfigure}
	\hspace{2pt}
	\begin{subfigure}[t]{0.22\textwidth}
		\centering
		\includegraphics[width=1\textwidth]{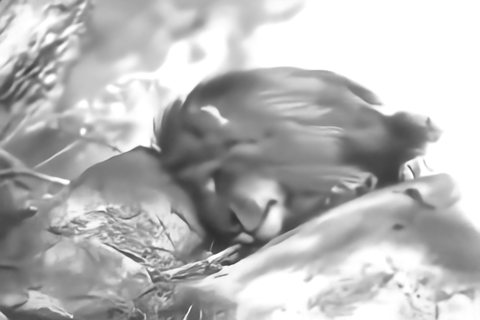}
	\end{subfigure}
	\begin{subfigure}[t]{0.22\textwidth}
		\centering
		\includegraphics[width=1\textwidth]{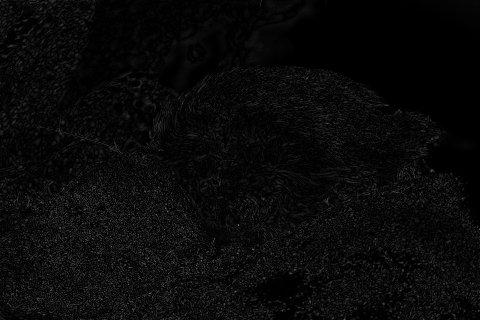}
	\end{subfigure}\\
	\begin{subfigure}[t]{0.22\textwidth}
		\centering
		\includegraphics[width=1\textwidth]{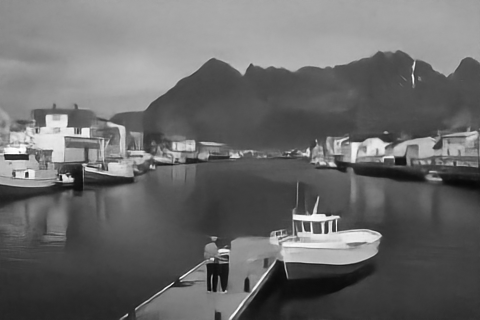}
	\end{subfigure}
	\begin{subfigure}[t]{0.22\textwidth}
		\centering
		\includegraphics[width=1\textwidth]{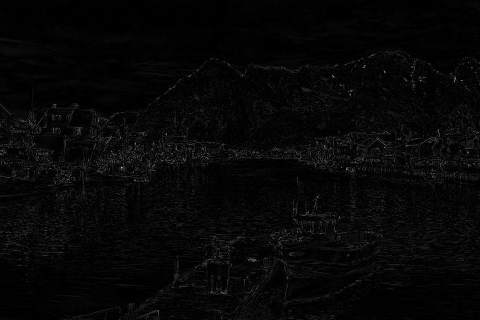}
	\end{subfigure}
	\hspace{2pt}
	\begin{subfigure}[t]{0.22\textwidth}
		\centering
		\includegraphics[width=1\textwidth]{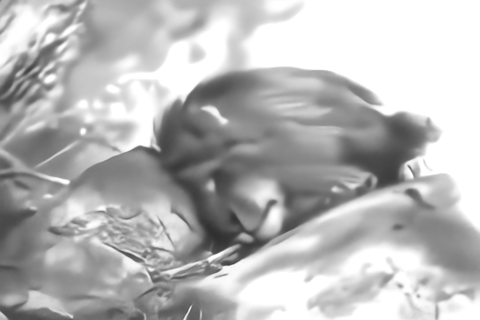}
	\end{subfigure}
	\begin{subfigure}[t]{0.22\textwidth}
		\centering
		\includegraphics[width=1\textwidth]{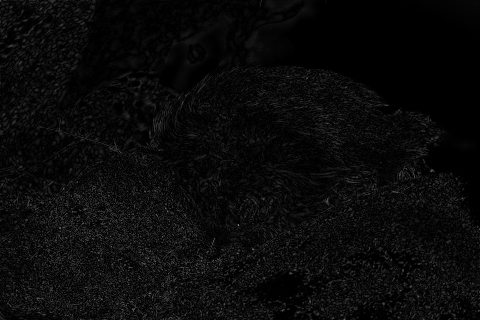}
	\end{subfigure}\\
	\caption{Denoising results on images from the BSD68 dataset.
	The input harbor image's noise level is set to 45 and the lion image's
	noise level is set to 25.}
	\label{loop}
\end{figure}

We further report the results of our algorithm in Fig. \ref{dn1} and Fig. \ref{dn2}.
We demonstrate the output of every second iteration of the restoration unit in Fig. \ref{loop}.
We also plot the PSNR variety when passing the restoration unit different times,
the tendency is plotted in Fig. \ref{psnr}.
The test performance increases during passing the first few steps,
but the benefit seems to diminish after a peak. To demonstrate this point
more intuitively, the residual image with our output and ground truth
is also demonstrated in Fig. \ref{loop}. This indicates us that adaptively
choose a stopping time is reasonable and necessary.

\subsection{Real Image Denoising}

\begin{figure}[htp!]
	\centering
	\begin{subfigure}[t]{0.15\textwidth}
		\centering
		\includegraphics[width=1\textwidth]{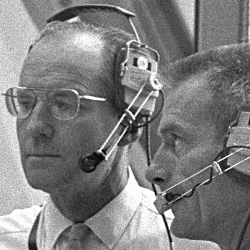}
	\end{subfigure}
	\begin{subfigure}[t]{0.15\textwidth}
		\centering
		\includegraphics[width=1\textwidth]{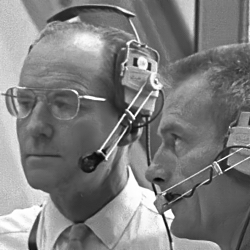}
	\end{subfigure}
	\begin{subfigure}[t]{0.15\textwidth}
		\centering
		\includegraphics[width=1\textwidth]{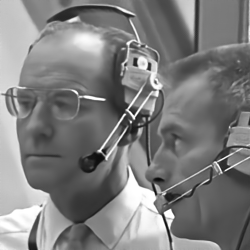}
	\end{subfigure}
	\begin{subfigure}[t]{0.15\textwidth}
		\centering
		\includegraphics[width=1\textwidth]{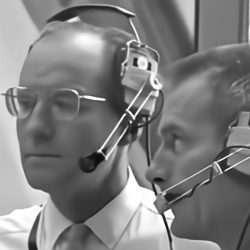}
	\end{subfigure}
	\begin{subfigure}[t]{0.15\textwidth}
		\centering
		\includegraphics[width=1\textwidth]{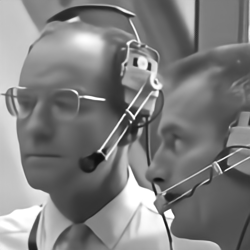}
	\end{subfigure}
	\begin{subfigure}[t]{0.15\textwidth}
		\centering
		\includegraphics[width=1\textwidth]{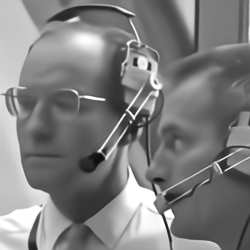}
	\end{subfigure}\\
	\centering
	\begin{subfigure}[t]{0.15\textwidth}
		\centering
		\includegraphics[width=1\textwidth]{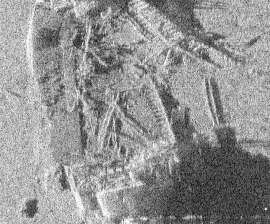}
		\subcaption*{ \scriptsize Noisy Image}
	\end{subfigure}
	\begin{subfigure}[t]{0.15\textwidth}
		\centering
		\includegraphics[width=1\textwidth]{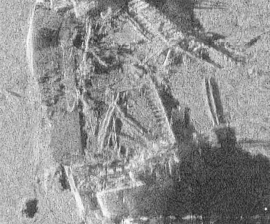}
		\subcaption*{1}
	\end{subfigure}
	\begin{subfigure}[t]{0.15\textwidth}
		\centering
		\includegraphics[width=1\textwidth]{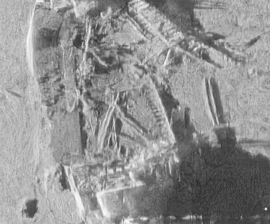}
		\subcaption*{2}
	\end{subfigure}
	\begin{subfigure}[t]{0.15\textwidth}
		\centering
		\includegraphics[width=1\textwidth]{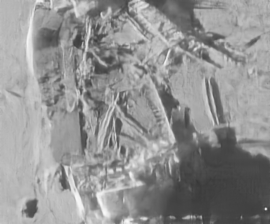}
		\subcaption*{3}
	\end{subfigure}
	\begin{subfigure}[t]{0.15\textwidth}
		\centering
		\includegraphics[width=1\textwidth]{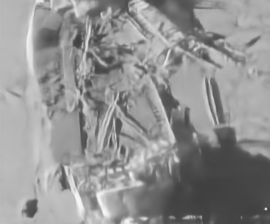}
		\subcaption*{4}
	\end{subfigure}
	\begin{subfigure}[t]{0.15\textwidth}
		\centering
		\includegraphics[width=1\textwidth]{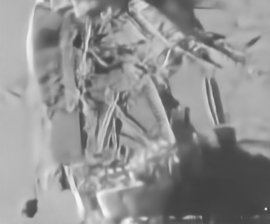}
		\subcaption*{5}
	\end{subfigure}
	\caption{Denoising result of a real image. The subcaption denotes the unfolding time.}
	\label{real1}
\end{figure}

In this section, we demonstrate more results of processing the real images.
In Fig. \ref{real1},
we demonstrate the output of the restoration unit with
different unfolding times (\textit{i.e.} passing the restoration unit
with different times). Results demonstrate that our network has
strong generalization ability and can be used to handle the problem
of real image denoising. Fig. \ref{real1}
show that our restoration unit behaves much like a bilateral filter,
which preserves the edges and reduces the noise.
If we filter the images for too many times, the images
tend to become over-smoothed.

\subsection{JPEG Deblocking}

\begin{figure}[htp!]
	\centering
	\begin{subfigure}[t]{0.25\textwidth}
		\centering
		\includegraphics[width=1\textwidth]{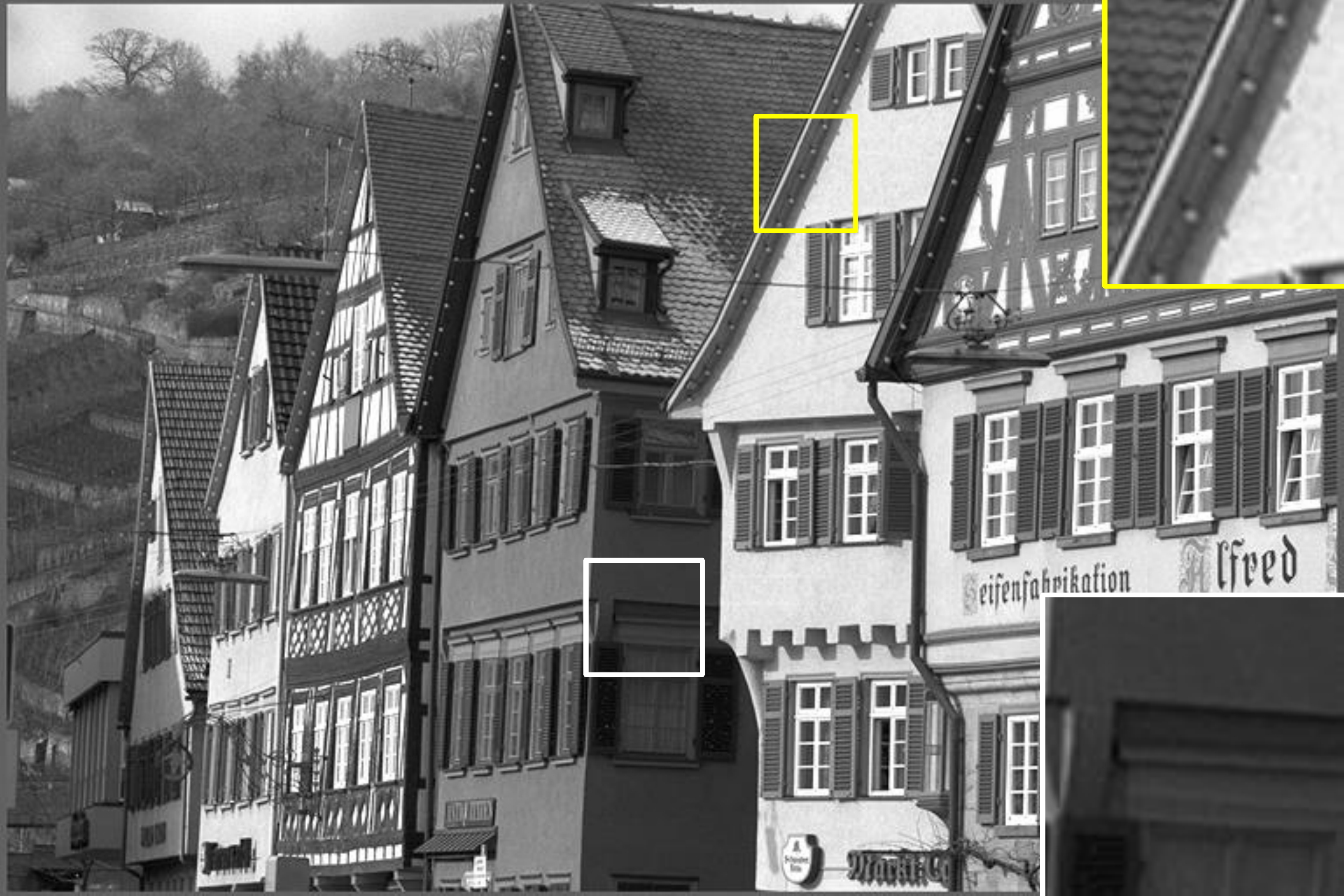}
		\subcaption*{Ground Truth}
	\end{subfigure}
	\begin{subfigure}[t]{0.25\textwidth}
		\centering
		\includegraphics[width=1\textwidth]{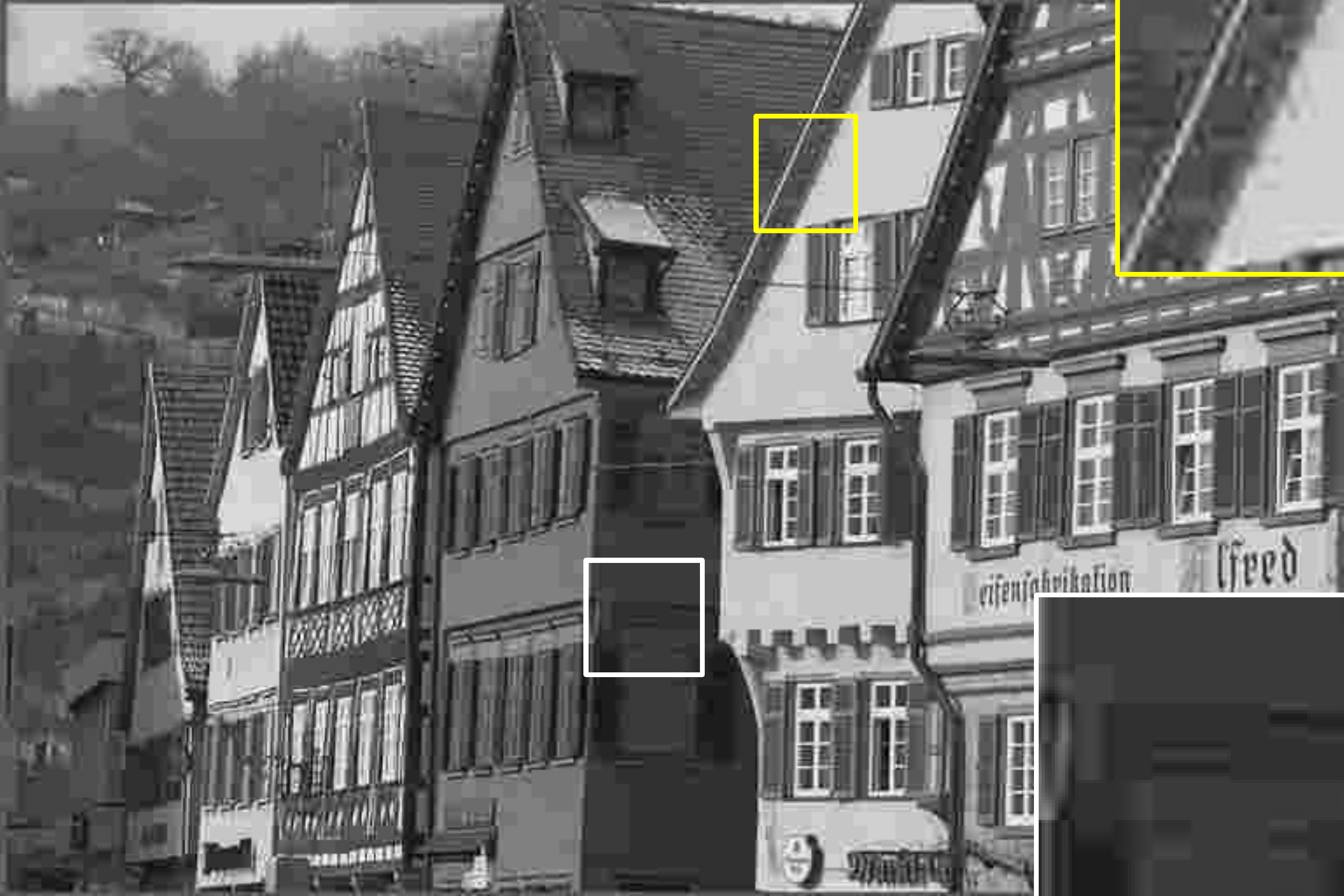}
		\subcaption*{JPEG}
	\end{subfigure}
	\begin{subfigure}[t]{0.25\textwidth}
		\centering
		\includegraphics[width=1\textwidth]{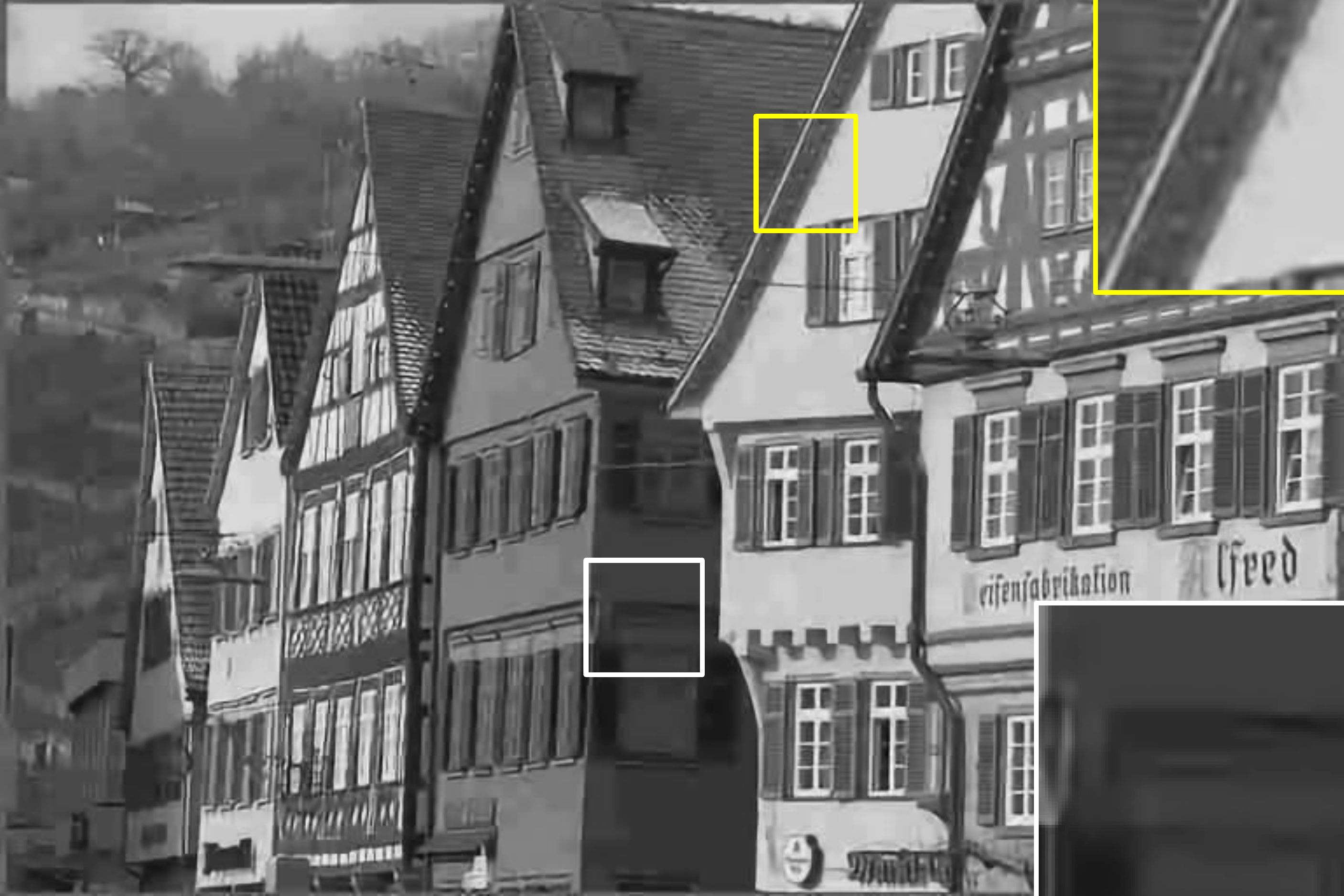}
		\subcaption*{(a) AR-CNN}
	\end{subfigure}

	\begin{subfigure}[t]{0.25\textwidth}
		\centering
		\includegraphics[width=1\textwidth]{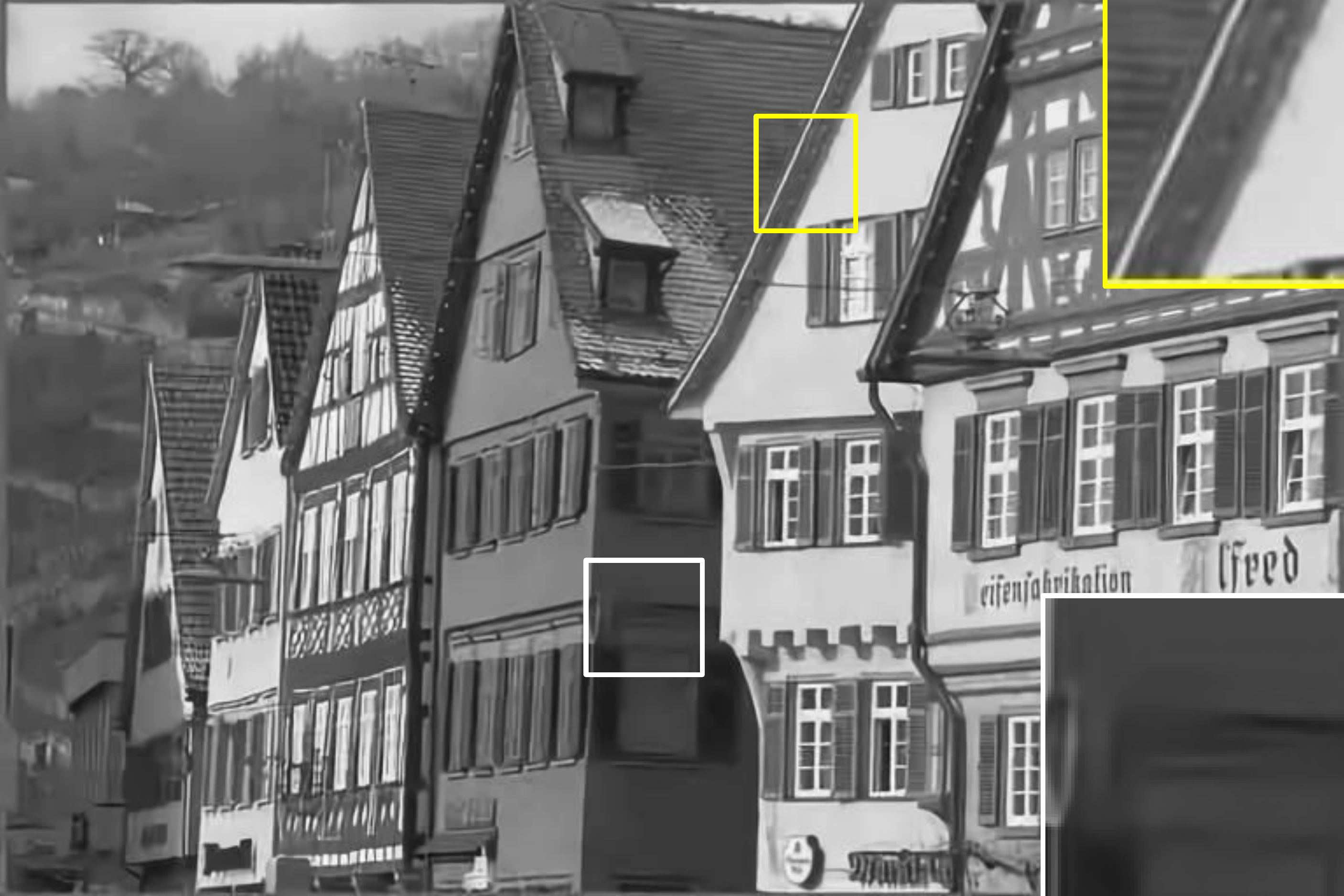}
		\subcaption*{(b) DnCNN}
	\end{subfigure}
	\begin{subfigure}[t]{0.25\textwidth}
		\centering
		\includegraphics[width=1\textwidth]{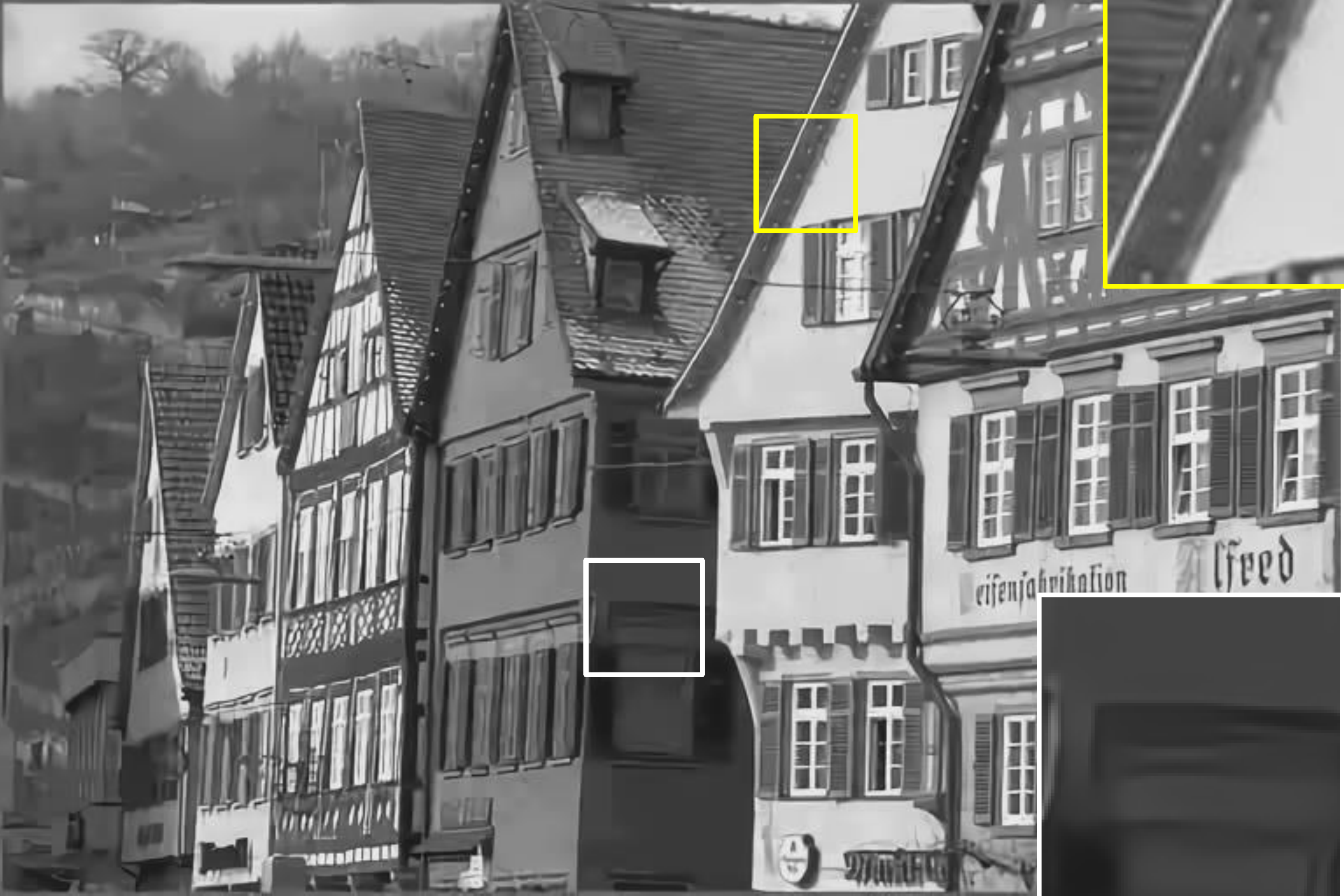}
		\subcaption*{(c) DURR}
	\end{subfigure}
	\caption{JPEG deblocking results of an image from the LIVE1 dataset,
		compressed using QF 10.}
	\label{jpegres}
\end{figure}

Here we demonstrate in Fig. \ref{jpegres} that our model is
able to remove the noise while preserving the structures. It
can be easily seen in the white zoom-in boxes that the edges of
the windows is well-preserved after the processing of DURR. In
the meantime DnCNN fails to keep the structure.

\end{document}